%% file: ms.tex
\documentclass{article}

\pdfoutput=1

\usepackage{arxiv}

\usepackage[utf8]{inputenc} 
\usepackage[T1]{fontenc}    
\usepackage{hyperref}       
\usepackage{url}            
\usepackage{booktabs}       
\usepackage{amsfonts}       
\usepackage{nicefrac}       
\usepackage{microtype}      

\usepackage{amssymb}
\usepackage{latexsym}

\usepackage{xcolor}
\usepackage{algorithmic}
\usepackage{algorithm}
\usepackage{graphicx}
\usepackage{caption}
\usepackage{subcaption}

\usepackage[toc,page]{appendix}

\title{BiNet: Degraded-Manuscript Binarization in Diverse Document Textures and Layouts using Deep Encoder-Decoder Networks}

\author{
   Maruf A. Dhali\thanks{Corresponding author; \newline Address: Nijenborgh 9, 9747 AG Groningen, The Netherlands; Email: m.a.dhali@rug.nl, web: \url{www.rug.nl/staff/m.a.dhali/}}, \hspace{0.5cm} Jan Willem de Wit, \hspace{0.5cm} Lambert Schomaker \vspace{0.2cm} \\ 
  Department of Artificial Intelligence, Bernoulli Institute\\
  University of Groningen \\
  The Netherlands\\
}

\begin{document}
\maketitle

\begin{abstract}
Handwritten document-image binarization is a semantic segmentation process to differentiate ink pixels from background pixels. It is one of the essential steps towards character recognition, writer identification, and script-style evolution analysis. The binarization task itself is challenging due to the vast diversity of writing styles, inks, and paper materials. It is even more difficult for historical manuscripts due to the aging and degradation of the documents over time. One of such manuscripts is the Dead Sea Scrolls (DSS) image collection, which poses extreme challenges for the existing binarization techniques. This article proposes a new binarization technique for the DSS images using the deep encoder-decoder networks. Although the artificial neural network proposed here is primarily designed to binarize the DSS images, it can be trained on different manuscript collections as well. Additionally, the use of transfer learning makes the network already utilizable for a wide range of handwritten documents, making it a unique multi-purpose tool for binarization.  Qualitative results and several quantitative comparisons using both historical manuscripts and datasets from handwritten document image binarization competition (H-DIBCO and DIBCO) exhibit the robustness and the effectiveness of the system. The best performing network architecture proposed here is a variant of the U-Net encoder-decoders.
\end{abstract}


\input{tex/1-introduction.tex}
\input{tex/2-background.tex}
\input{tex/3-methods.tex}
\input{tex/4-experiments.tex}

\input{tex/5-results.tex}
\input{tex/6-conclusions.tex}
\input{tex/acknowledgements.tex}

\bibliographystyle{unsrt}

\newpage
\input{tex/7-appendix.tex}

\end{document}

%% file: tex/1-introduction.tex
\section{Introduction}
In a digitized image of a handwritten document, the ink-based pixels are the result of a physical ink deposition process, where a surface material absorbs the pigment creating the foreground. In contrast, the original unaffected material texture appears as the background.  In a typical handwritten document,\\
\\
$N_{ink} << N_{background}$\\
where, $N_{ink}$ is the number of ink pixels and $N_{background}$ is the number of background pixels.\\ 
\\
The binarization process of handwritten document-image allocates a binary value to each pixel of the image \cite{calvo2019selectional}; $0$ for ink and $1$ for not-ink. Thus, this process separates the foreground (meaningful information, in general, the ink) from the background (the surface material). The binarized images are compressed and pose significant importance towards analyzing the document. It facilitates character recognition, segmentation, and transcription pipeline \cite{almeida2018new, louloudis2008text, chaki2014exploring}. Further processing of the handwriting towards writer identification and script-style development also depends on the success of the binarization process itself \cite{dhali2017digital, he2015junction}. Over the years, many techniques have been proposed to perform binarization tasks. However, it is always challenging to obtain good results due to the diversity of the handwritten documents. One technique may perform well for some specific type of documents but fail for other types. The problem becomes challenging when it comes to historical manuscript binarization \cite{sulaiman2019degraded}.%

\begin{figure}[h!]
	\centering
	\begin{subfigure}[]{0.33\textwidth}
		\centering
		\fbox{\includegraphics[height=4.3cm]{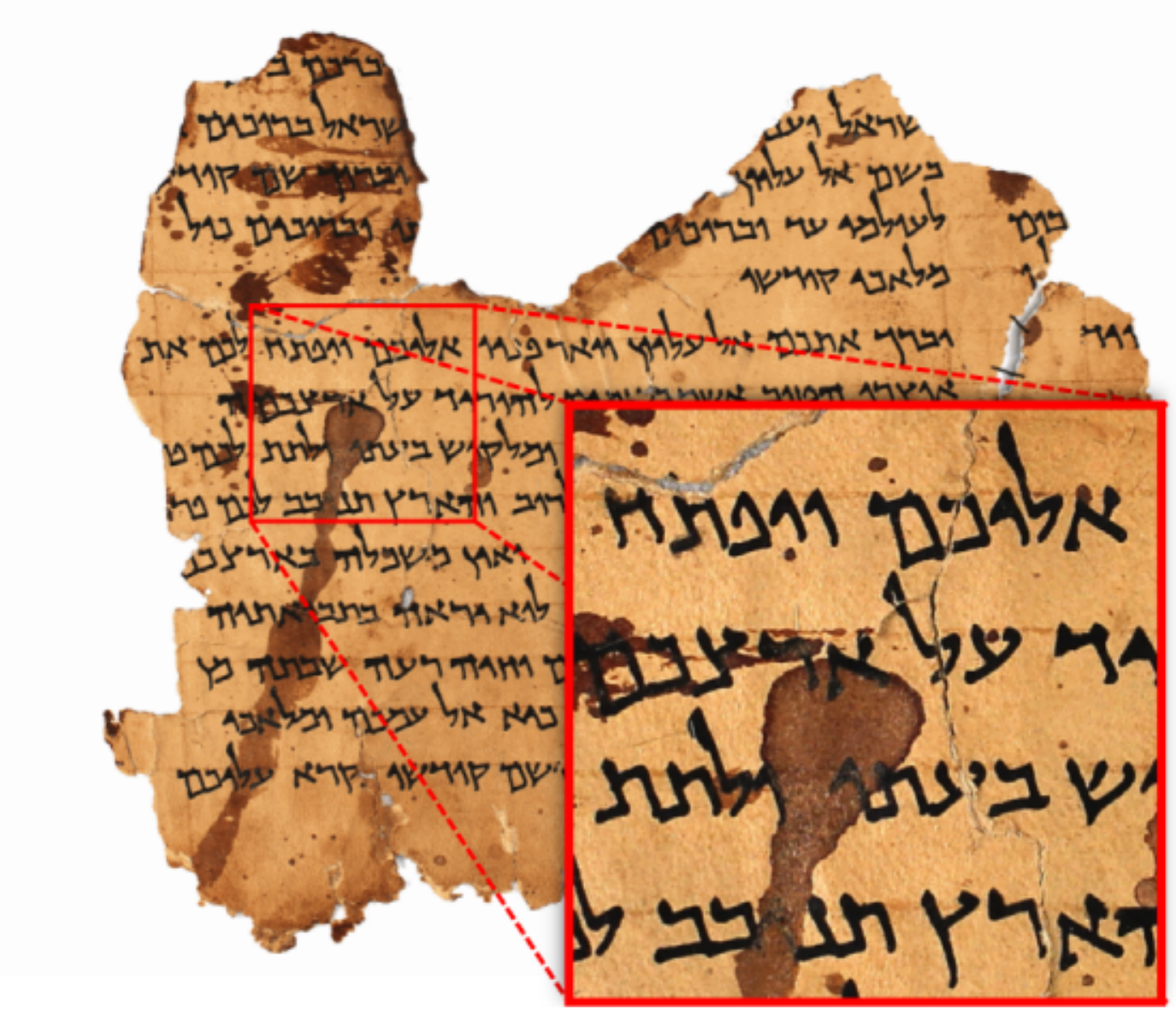}}%
		\caption{Plate \textit{671-1}}
	\end{subfigure}\hfill
	\begin{subfigure}[]{0.33\textwidth}
		\centering
		\fbox{\includegraphics[height=4.3cm]{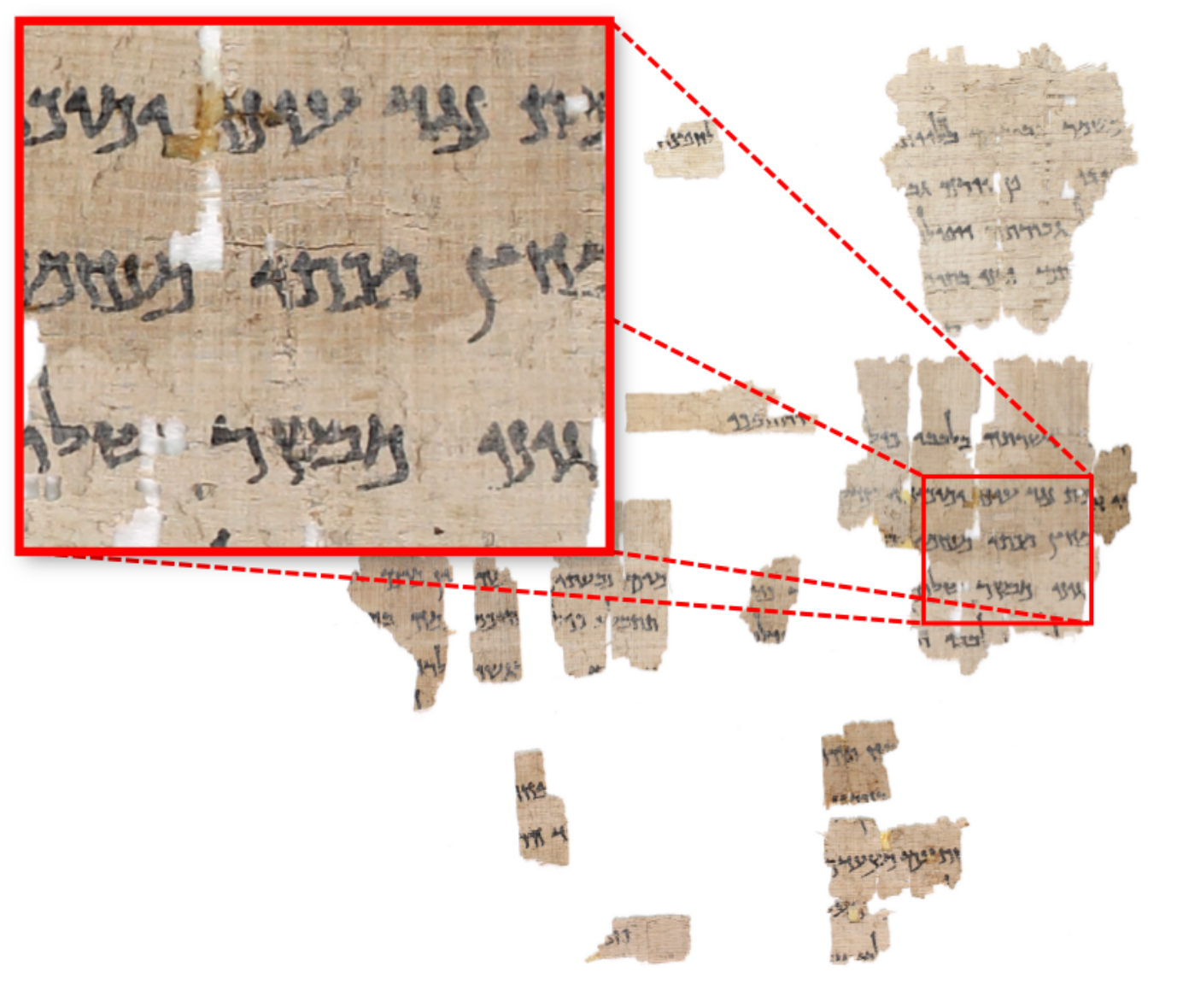}}%
		\caption{Plate \textit{117}}
	\end{subfigure}\hfill
	\begin{subfigure}[]{0.33\textwidth}
		\centering
		\fbox{\includegraphics[height=4.3cm]{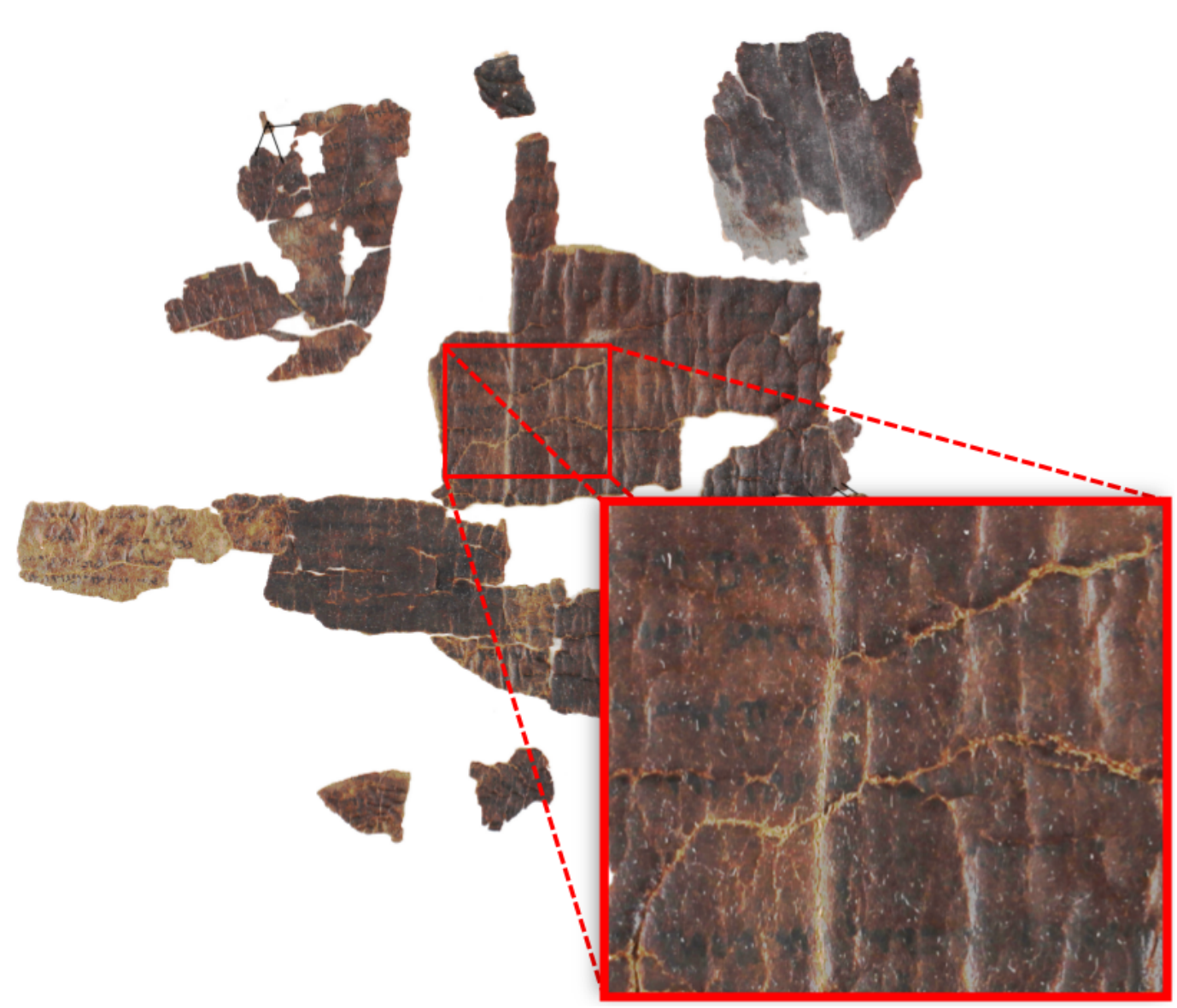}}%
		\caption{Plate \textit{215}}
	\end{subfigure}
	\caption{Three images from the Dead Sea Scrolls collection show the diversity of the materials and their current state with difficult readability due to various degradation. They are all RGB-colored images of the plates (physical arrangements of the fragmented materials on a plane surface). Plate \textit{607-1} contains only one physical fragment, whereas the next two plates (\textit{117} and \textit{215}) contain multiple physical fragments. Fragments from plate \textit{117} were produced from papyrus, and the repetitive patterns of the fibers are visible in the zoomed-in section of the image. The other two plates contain fragments made from parchments.}
	\label{fig:intro3img}
\end{figure}%

Numerous historical manuscripts collections are residing all over the world \cite{marinai2004general}. Most of them are significant, both culturally and scientifically \cite{antonacopoulos2004document}. The Dead Sea Scrolls (DSS) are one such collection. They are ancient manuscripts discovered in the mid-20th century in the Judaean Desert, in between Jerusalem and the Dead Sea. Most were written over a period of almost four centuries (ca. 250 BC to ca. 135 AD) and hold tremendous historical, religious, and linguistic significance \cite{popovic2016ancient}. The recent digitization of this collection has opened the door for pattern recognition techniques to be applied to revise existing hypotheses on the writers and dates of these scrolls \cite{dhali2017digital}. However, these documents have diverse document textures and are heavily degraded (see Figure \ref{fig:intro3img}), mostly due to the materials, natural aging, the preservation processes, and the places they were kept in. In order to perform pattern recognition techniques on the original content (texts) only, the images need to be preprocessed. One of the critical steps in this preprocessing is a binarization technique with the ability to keep the original content of these documents as intact as possible.

\begin{figure}[h!]
	\centering
	\begin{subfigure}[]{0.48\textwidth}
		\centering
		\includegraphics[height=4.3cm]{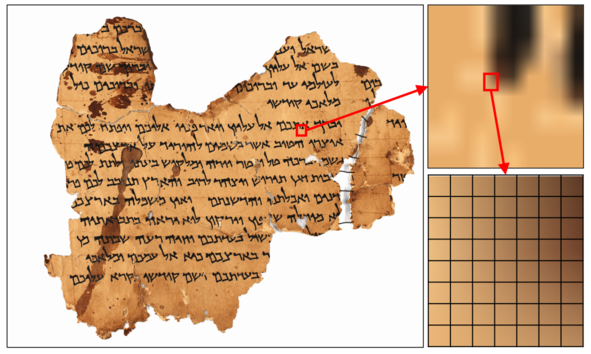}%
		\caption{Plate \textit{671-1}; RGB-colored image}
	\end{subfigure}
	\begin{subfigure}[]{0.48\textwidth}
		\centering
		\includegraphics[height=4.3cm]{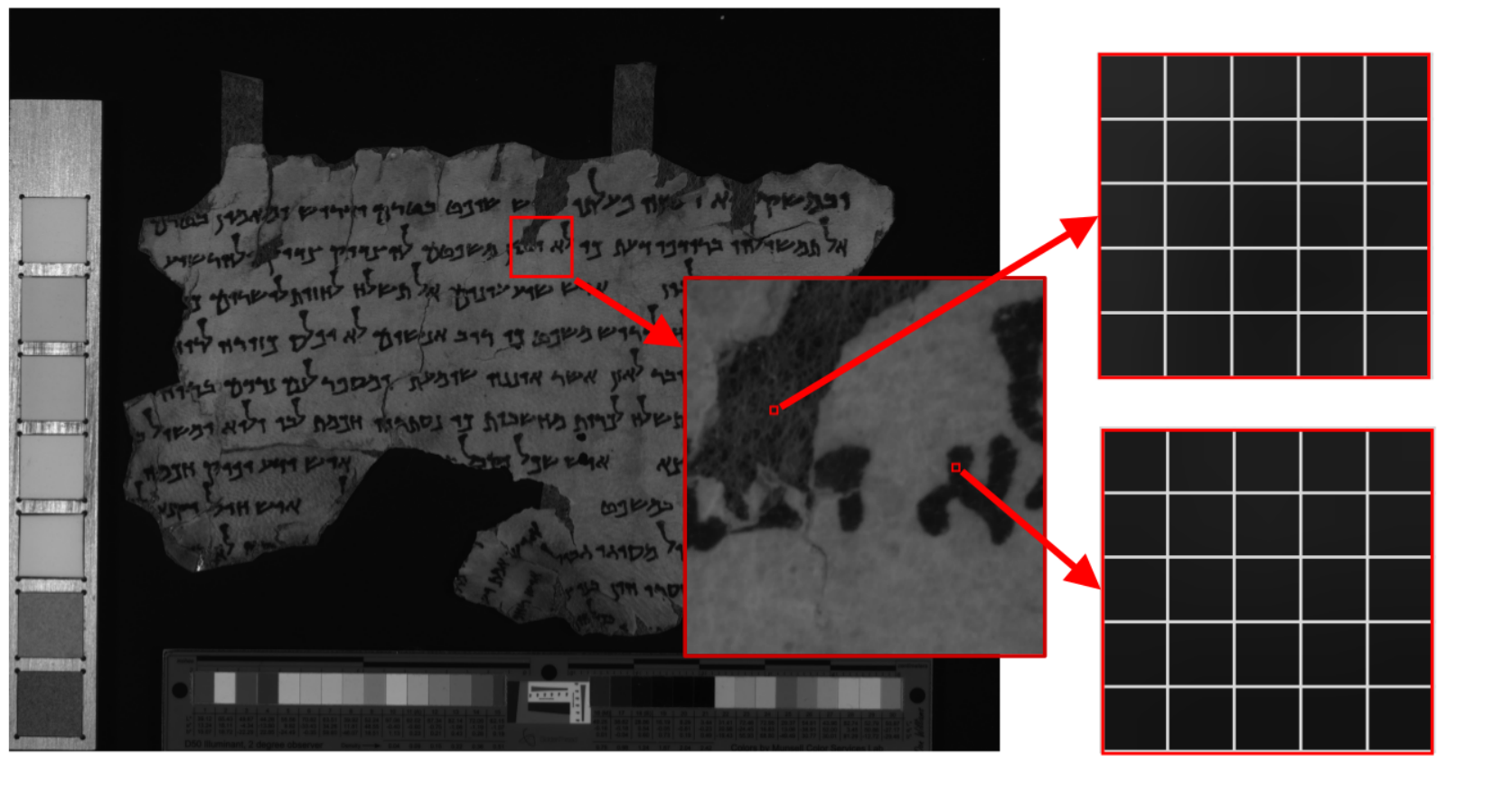}%
		\caption{Plate \textit{123}, Fragment \textit{2}; infrared intensity image}
	\end{subfigure}
	\caption{The degraded DSS documents show the variation of contrast. The \textit{left} image-set shows an RGB-colored plate with zoomed in to pixel level at the edge of an ink-stroke. The \textit{right} image-set shows an IR-image (captured in $924nm$ wavelength of light) of a fragment in grayscale. Here, the two zoomed-in portions from the degraded background (top) and the ink (bottom) show no visual differences in color intensity.}
	\label{fig:introZoomed}
\end{figure}

There are several challenges in binarizing the DSS images. Similar to many other historical manuscripts, the DSS collection profoundly suffers from document degradation problems. Due to aging and natural causes, the individual glyphs (characters) of the DSS images often show fading effects. Some of the images also show thinning of the characters along with broken (missing or completely faded) parts. Some images often suffer from uneven illumination problems due to the surface material \cite{sulaiman2017database}. On top of all these, the most severe issue is the low contrast between ink and background (see Figure \ref{fig:introZoomed}).%
 
\subsection{Why binarization is still important}
Although many modern deep-learning methods in document analysis can be trained end-to-end, directly with a grayscale or color image, this is not desirable in an e-science approach for the humanities studies. For instance, a direct end-to-end solution on clustering the DSS collection can be achieved for writer identification and script-style evolution, but there is always a risk of getting the solution for the wrong cause as in the `Russian (hidden) tank problem' \cite{dreyfus1992artificial}.  The decision of a neural network may be based on spurious correlations with the texture of the background of different materials, such as papyrus and parchment. The fiber statistics of papyrus manufacturing batches may add to the wrong cause, as well. Additionally, irrelevant materials like the background, rice papers, number tags, scale bars, color calibrators, and other patterns must not be allowed to contribute to the process. So binarization is necessary, and it needs to be precise. 

\begin{figure}[h!]
	\centering
	\begin{subfigure}[]{0.2\textwidth}
		\centering
		\fbox{\includegraphics[height=2.2cm]{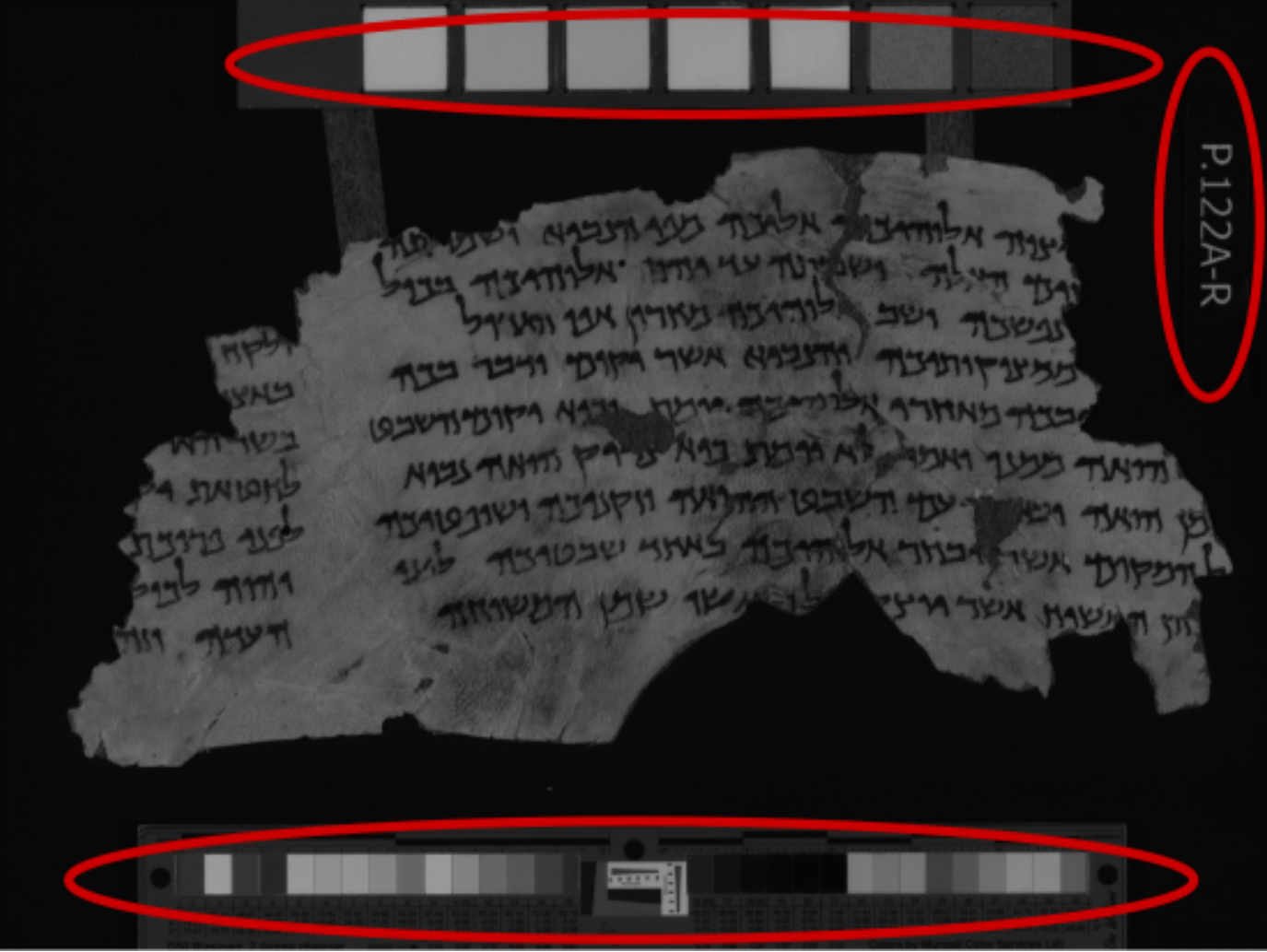}}%
		\caption{Plate \textit{122A}, Frag. \textit{1}}
	\end{subfigure}\hfill
	\begin{subfigure}[]{0.2\textwidth}
		\centering
		\fbox{\includegraphics[height=2.2cm]{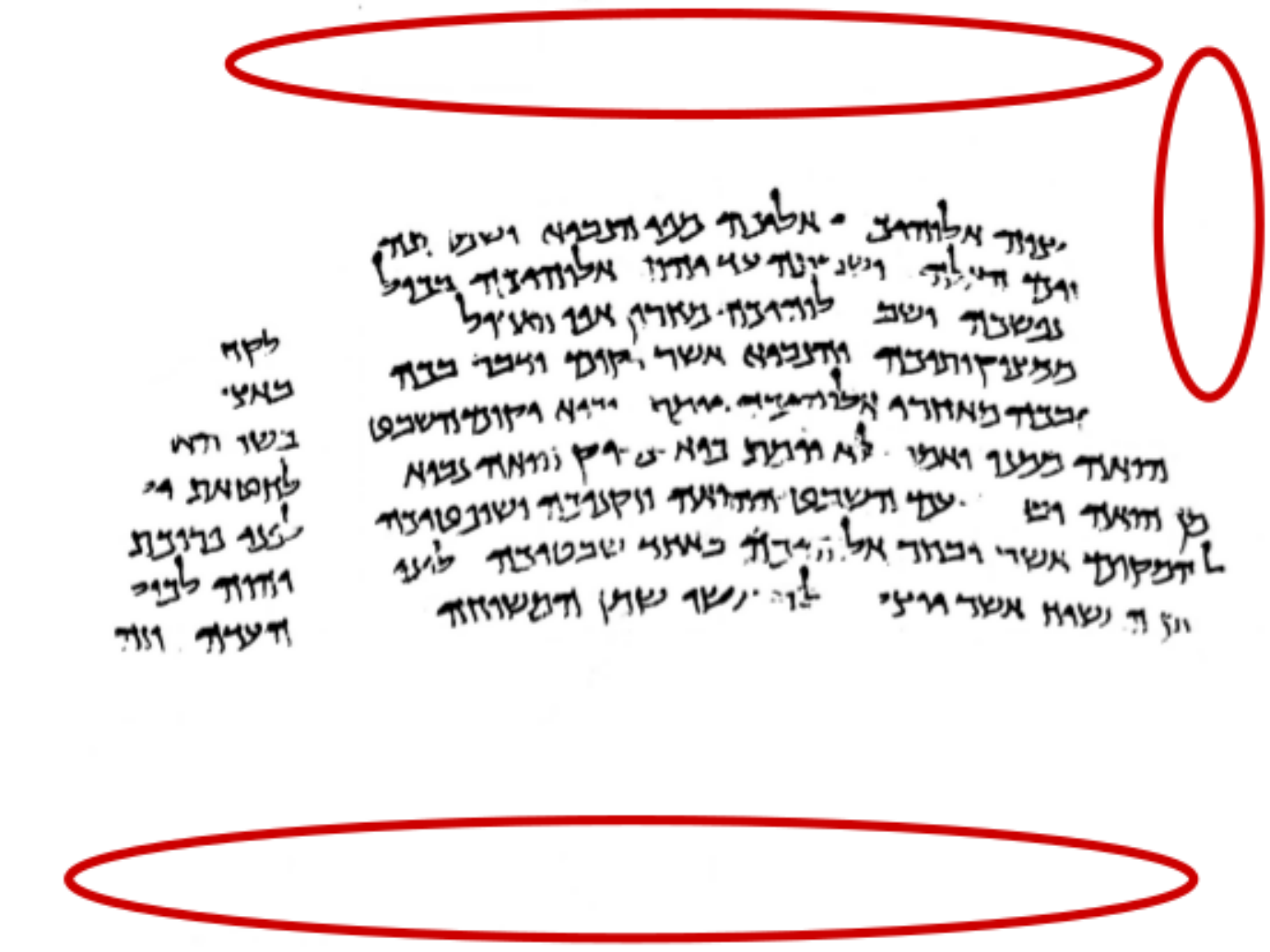}}%
		\caption{Manually labeled}
		\label{subfig:manual1}
	\end{subfigure}\hfill
	\begin{subfigure}[]{0.2\textwidth}
		\centering
		\fbox{\includegraphics[height=2.2cm]{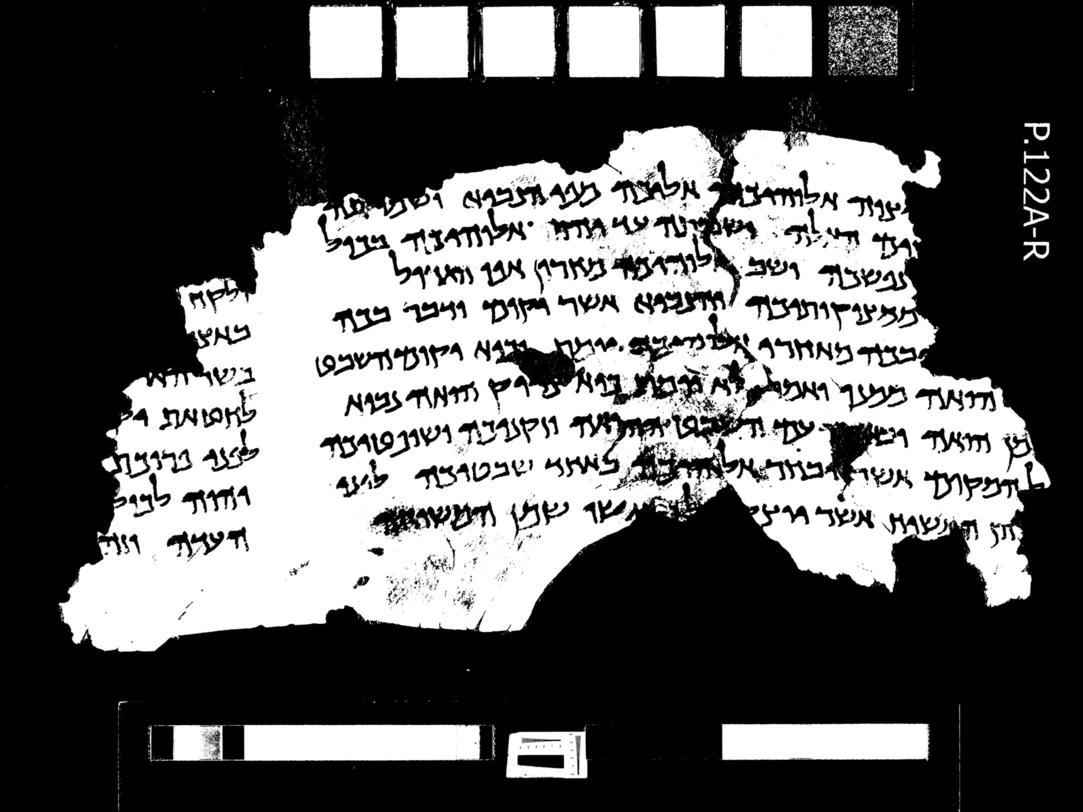}}%
		\caption{Otsu}
	\end{subfigure}\hfill
	\begin{subfigure}[]{0.2\textwidth}
		\centering
		\fbox{\includegraphics[height=2.2cm]{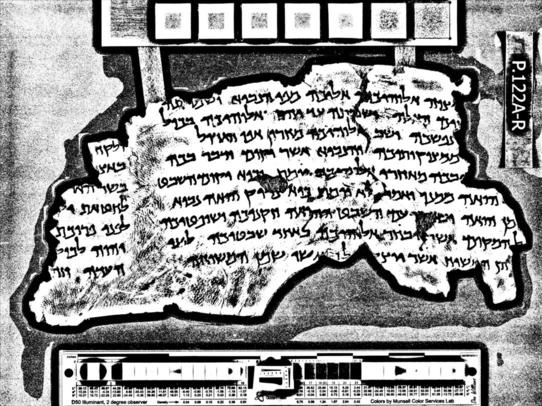}}%
		\caption{Niblack}
	\end{subfigure}\hfill
	\begin{subfigure}[]{0.2\textwidth}
		\centering
		\fbox{\includegraphics[height=2.2cm]{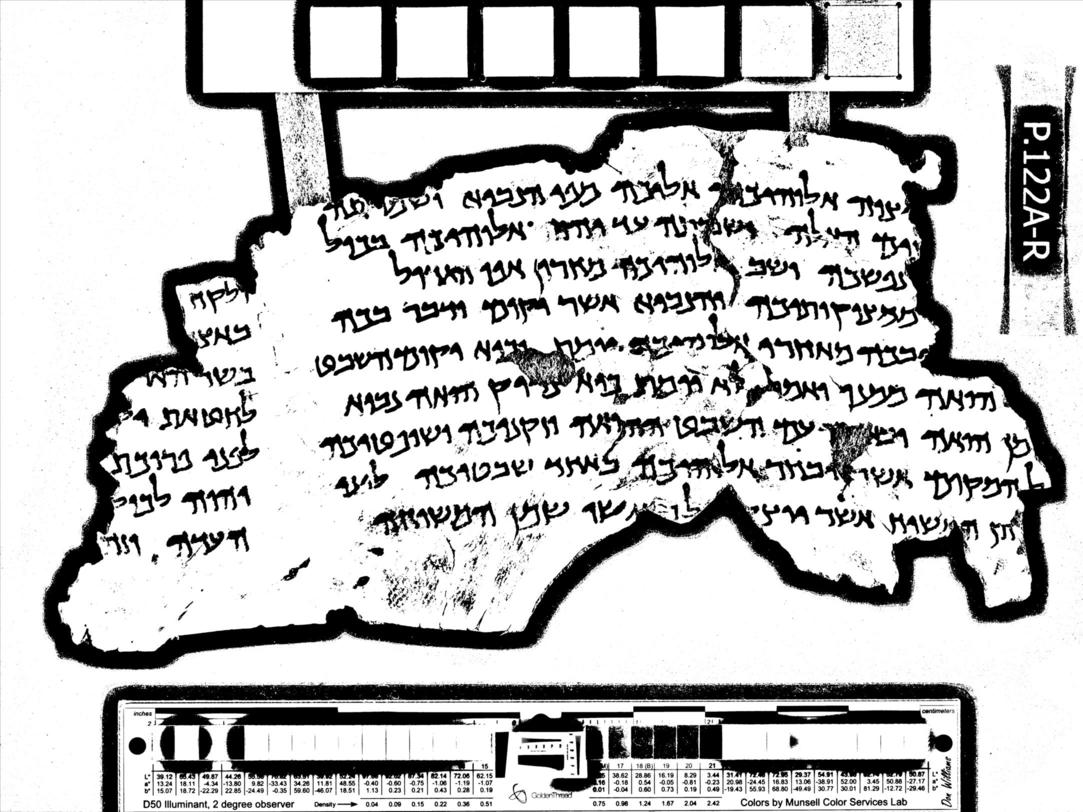}}%
		\caption{Sauvola}
	\end{subfigure}
	\begin{subfigure}[]{0.2\textwidth}
		\centering
		\fbox{\includegraphics[height=2.2cm]{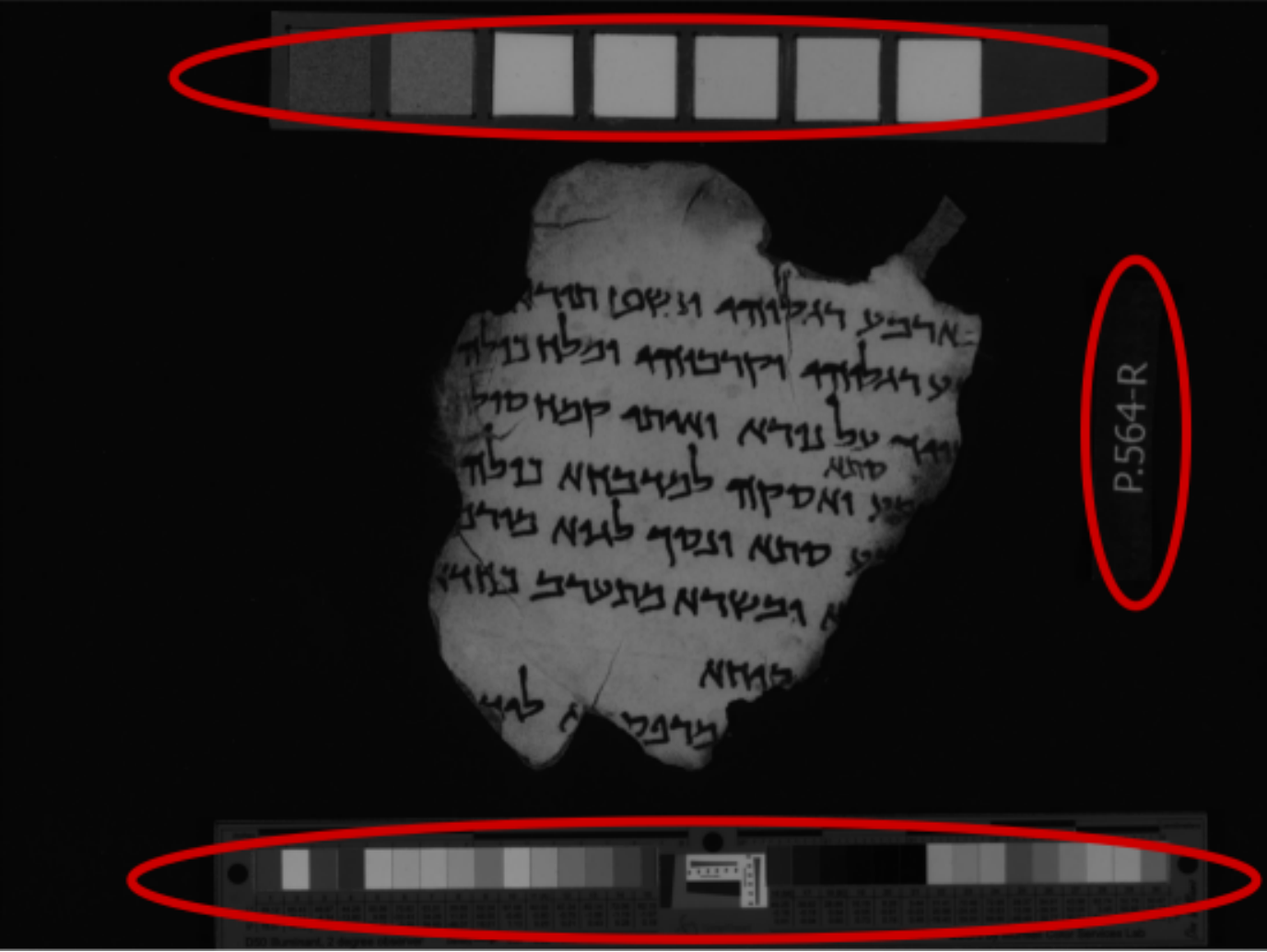}}%
		\caption{Plate \textit{564}, Frag. \textit{3}}
	\end{subfigure}\hfill
	\begin{subfigure}[]{0.2\textwidth}
		\centering
		\fbox{\includegraphics[height=2.2cm]{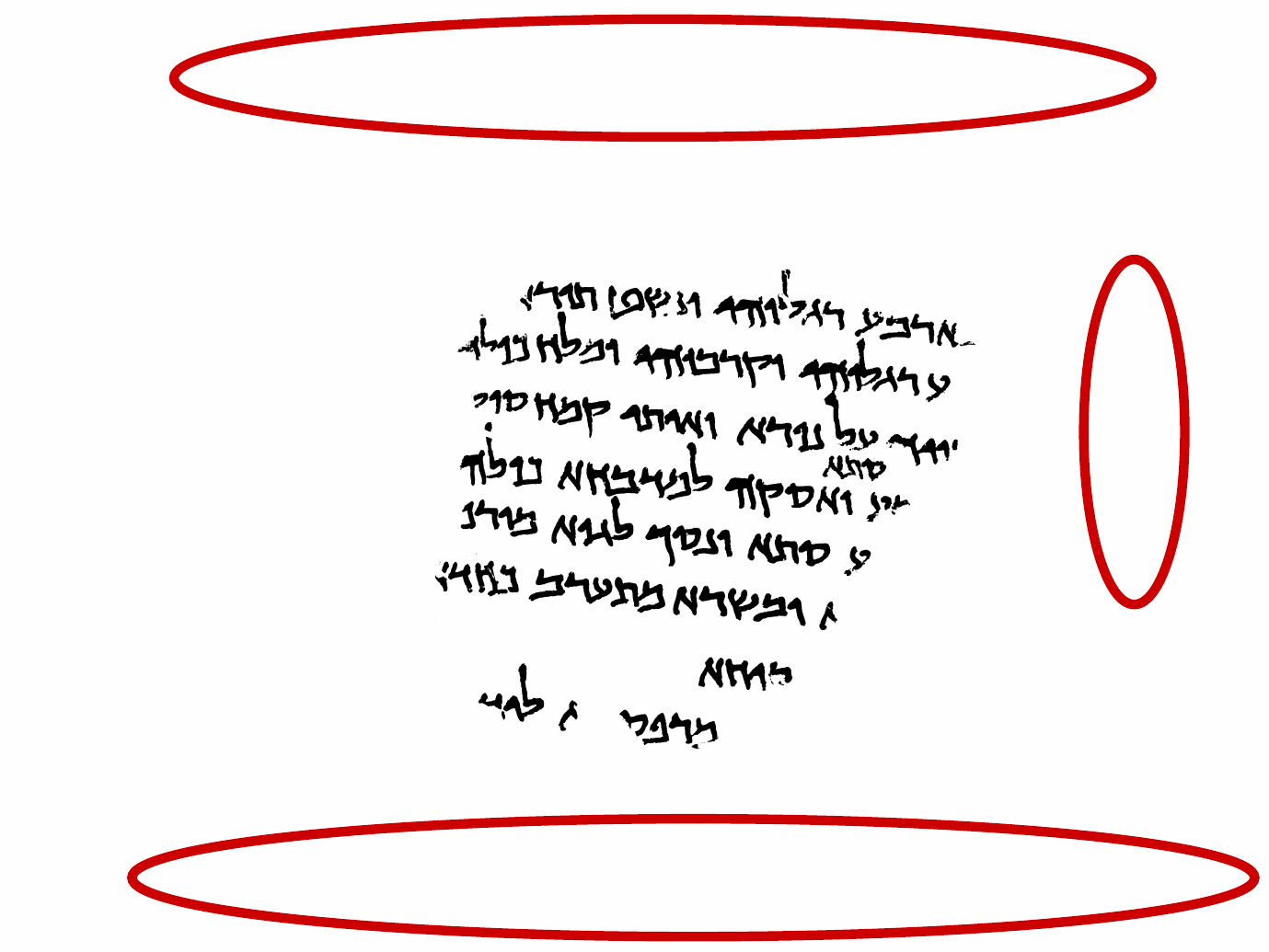}}%
		\caption{Manually labeled}
		\label{subfig:manual2}
	\end{subfigure}\hfill
	\begin{subfigure}[]{0.2\textwidth}
		\centering
		\fbox{\includegraphics[height=2.2cm]{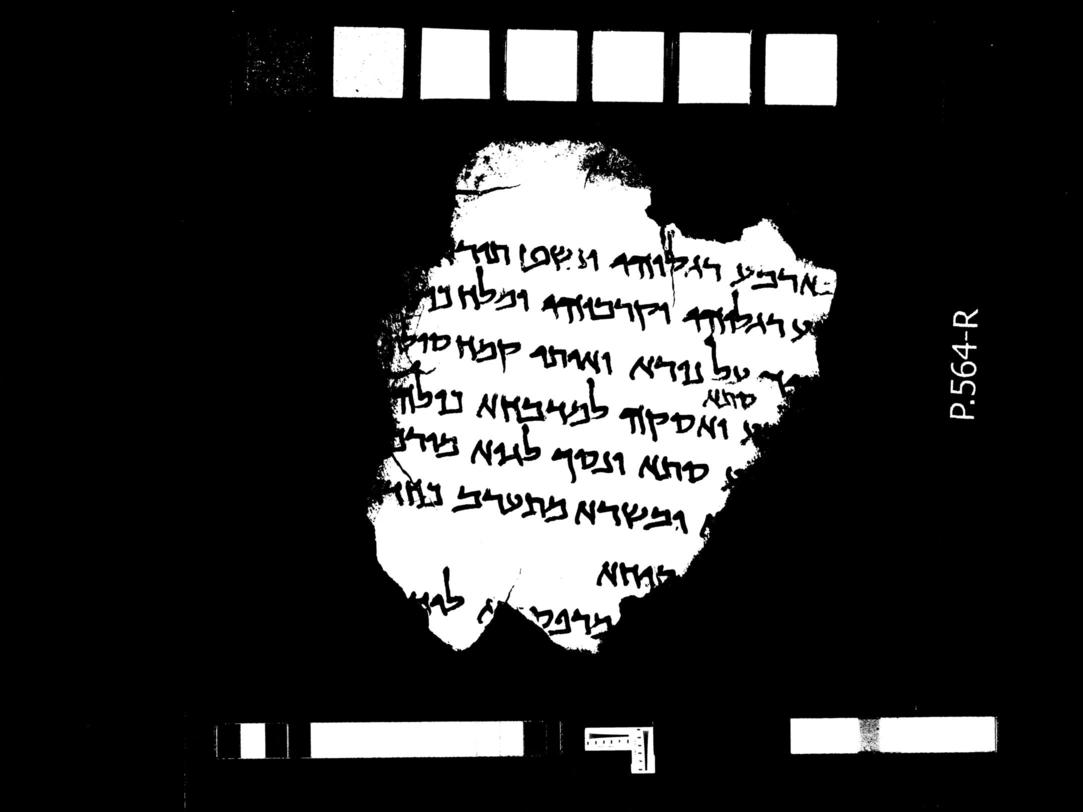}}%
		\caption{Otsu}
	\end{subfigure}\hfill
	\begin{subfigure}[]{0.2\textwidth}
		\centering
		\fbox{\includegraphics[height=2.2cm]{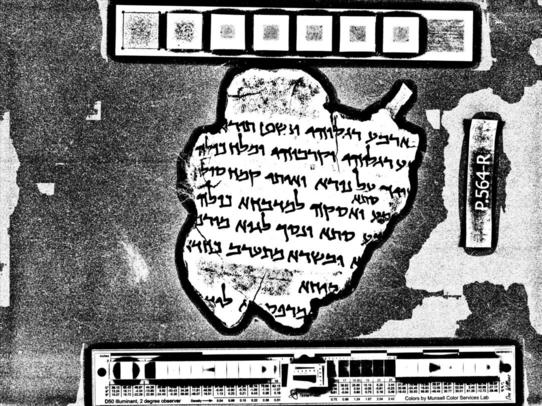}}%
		\caption{Niblack}
	\end{subfigure}\hfill
	\begin{subfigure}[]{0.2\textwidth}
		\centering
		\fbox{\includegraphics[height=2.2cm]{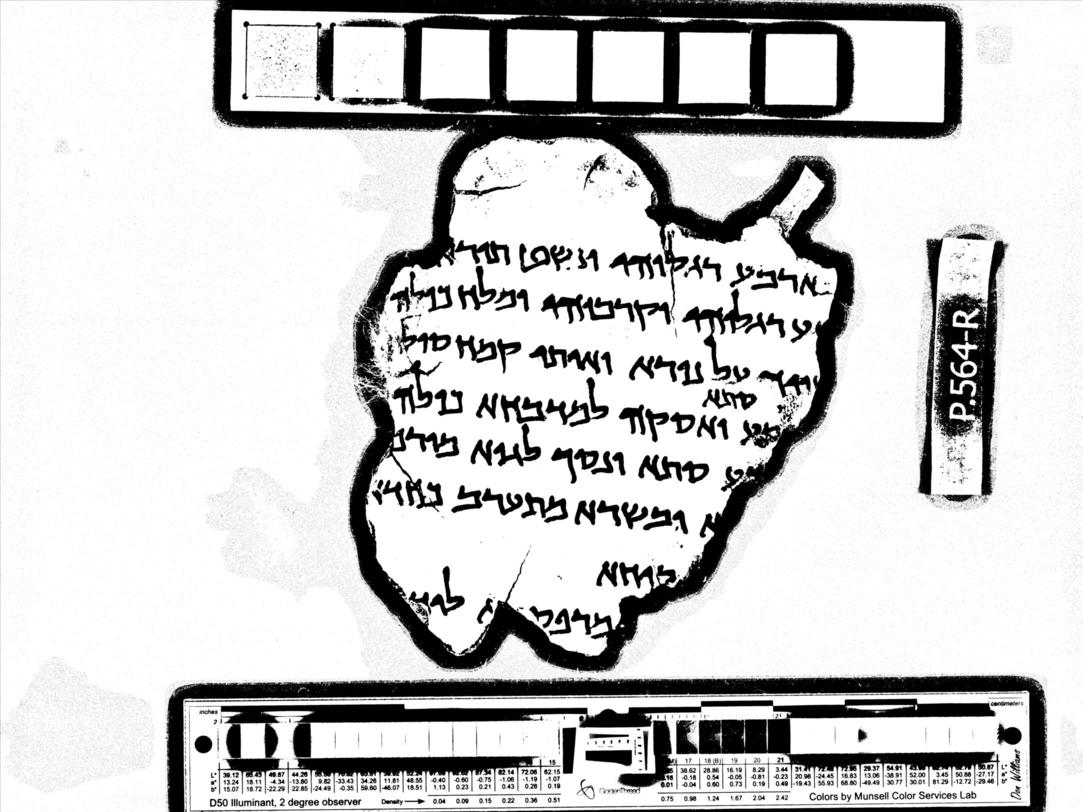}}%
		\caption{Sauvola}
	\end{subfigure}
	\caption{An illustration of popular binarization techniques applied directly to two of the DSS fragment-images. Sub-figures (a) and (f) show the original IR-images (captured in 924-nm wavelength of light). Sub-figures (b) and (g) show the corresponding manually labeled ground truths by human experts. The {\color{red}red}-circled areas show the parts where the human experts ignore the irrelevant contents of the images, such as the color-calibration bars, scales, and numbers. The binarization results of techniques proposed by Otsu \cite{otsu1979threshold}, Niblack \cite{niblack1985introduction}, and Sauvola \cite{sauvola2000adaptive} are presented for both the fragment images. All these three methods fail to provide output images that focus only on the original contents, unlike a human expert.}
	\label{fig:introYardeni}
\end{figure}

Many of the existing binarization techniques are pixel-intensity based. It is already clear that any method working on the pixel-intensity only will struggle in producing excellent results for the DSS images. An intelligent binarization method should accommodate all different variability in the DSS collection and still provide excellent results. Semi-automatic selection of the region of interest or a manual one-off preprocessing technique may obtain good binarization, but it will not serve as a robust solution for the whole corpus. For objective analysis, including writer identification and dating, a non-biasing foreground-background separation is required. Separating foreground-background can be a severe problem to solve but significant for scholarly analysis. This problem can be illustrated using a suggestion put forward by the eminent palaeographer Ada Yardeni. She ascribed fifty-seven or possibly even ninety-three manuscripts to one scribe (\cite{yardeni2007note, lubetski2007new}). Two of such fragment images are shown in Figure \ref{fig:introYardeni}. This figure also presents the binarization results from three of the most famous traditional techniques. In the case of ink separation, the techniques perform considerably well in some regions of the image but fail in most of the remaining areas. Now, for the case of writer identification, the palaeographer who is strictly interested in comparing two hand-writings to check the hypothesis from Yardeni would not accept any of the results based on such binarization data. For writer identification, the binarization technique should be capable enough to focus on the original written content only, not the surrounding material, not even the markings and scale bars. 

\subsection{Goals}
Intensity alone is not a sufficient heuristic in the binarization task. Rather than using a single filtering technique, a bank of trainable filters is needed to solve this problem. The system should be able to ignore the irrelevant information during binarization and should be able to include everything which is part of the original content. A system needs to be adequately intelligent to focus on writing like a human does with reasonable accuracy (see Sub figures \ref{subfig:manual1} and \ref{subfig:manual2}). An artificial neural network, with suitable architecture requiring a small amount of training data, can be the right solution as a multi-filter trainable method for these diverse materials. Towards the goal of both robustness and obtaining optimal results, this article proposes BiNet, an unbiased automatic end-to-end binarization approach for handwritten documents based on the U-Net architecture (\cite{ronneberger2015u}). Inspired by the works of `pix2pix', a general-purpose solution for image-to-image translation problems using conditional adversarial networks (\cite{isola2017image}),  the proposed model includes the encoder-decoder structure without the discriminator part making it a variant of the U-Net. Skip-connections are added between the contracting and the expansive path by simply concatenating all channels from one layer to the other. This concatenation circumvents the bottle-neck issue at the deepest layers of the encoder and ensures the precise positioning of the foreground-background pixels. A simple illustration of the proposed network is provided in Figure \ref{fig:introNetwork}.

This study demonstrates the effectiveness of the proposed model on the binarization task using the collection of the DSS images. Both the RGB-colored images and the pseudo-colored images are used. Pseudo-colored images are fused from grayscale intensity-images of different spectral bands. A simple technique is proposed in generating ground-truth images to train the network. Similar to many historical manuscripts, the DSS collection also lacks in ground-truth labels. It is time consuming and tedious to create ground truth at the pixel level. So the work in this paper ensures that the training data is precise, includes a sufficient amount of variability, and the network can perform well with a small amount of training data. On top of this, transfer learning techniques are introduced so that the proposed model becomes usable for different collections. Additionally, the Handwritten Document Image Binarization Competition (H-DIBCO) datasets are tested to exhibit the effectiveness of the system. Quantitative and qualitative results are presented to compare it with other techniques. BiNet, the proposed model, shows better performance with robustness to the variability of the data. Overall, this article makes the following contributions:
\begin{itemize}
	\item BiNet: the complete framework for a fully automated binarization method for DSS images that allows further analysis of the original contents of the collection.
	\item A network capable of learning the differentiation between relevant and irrelevant information during the training process, and thus providing intelligent and useful binarized outputs.
	\item An in-depth analysis of the proposed binarization tool using comparative studies and quantitative analysis.
	\item Multi-purpose usability of the binarization tool for different manuscript collections (including H-DIBCO images) using effective transfer learning techniques.
	\item A simple and easy technique to generate precise ground-truths (training images) for the DSS collection that can be extended to any degraded historical manuscripts.
	\item A new way to generate fused (pseudo-color) images from multi-spectral bands to yield more information (qualitatively) than any of the individual bands. 
\end{itemize}

%% file: tex/2-background.tex
\section{Related Works}
Document image binarization is one of the most common research problems that has been addressed numerous times in the field of document analysis. Some of these methods have achieved great success in many applications and have become popular over time. Otsu \cite{otsu1979threshold} is one of the most commonly used methods. This unsupervised and non-parametric method automatically selects a global threshold based on the grayscale histogram of a given image with no prior information. The Otsu method is one of the simplest binarization methods that perform well when the image is qualitatively clean with a uniform background. Unfortunately, most of the historical manuscripts do not contain a uniform background or a clear bimodal pattern. Thus, a global thresholding approach is not suitable for these types of documents \cite{kittler1985threshold}. A gradual change in the uniformity of the background can be handled using small local patches of the target image by local adaptive thresholding. Several local thresholding methods have been developed, such as Niblack \cite{niblack1985introduction}, Sauvola \cite{sauvola2000adaptive}, local max-min by Su et al.\cite{su2010binarization}, and AdOtsu \cite{moghaddam2012adotsu}. Descriptive statistics (mean and standard deviation) are calculated on the local area of a pixel to obtain the local threshold. This local thresholding technique is then performed over the whole target image. It performs well compared to the global thresholding techniques, but often shows poor performance in the case of historical manuscripts where the documents are highly degraded with extremely non-uniform backgrounds. To extend further, in cases similar to the DSS collections, both the global and local thresholding methods fail to provide useful results (an example of this can be found in Figure \ref{fig:introYardeni} from the previous section). 

In order to improve the results of threshold-based binarization, several image-processing techniques are used as an enhancement part of the document-processing pipeline along with the binarization itself. Shi et al. used the mathematical morphological operator and region-growing techniques \cite{shi2011image}. In order to compute the final threshold, a Wiener filter \cite{wiener1949back} is used for the background surface by Gatos et al. \cite{gatos2006adaptive}. Instead of using the Wiener filtering, robust regression is used by Vo et al. for document binarization of noisy and non-uniform background \cite{vo2018robust}. Phase-derived features are used for ancient document image binarization in the works of Nafchi et al. \cite{nafchi2014phase}. In order to enhance and reconstruct degraded documents, a method using non-local patch means (NLPM) is proposed by Moghaddam et al. \cite{moghaddam2011beyond}. Bio-inspired models have already been used for text detection in natural images \cite{zagoris2013text}. Similarly, models based on the OFF-center ganglion cells of the human visual system is used for the improvement of document enhancement and binarization \cite{zagoris2017bio}. Different contrast enhancements are performed to adjust local grayscale contrast to improve the binarization results compared to traditional threshold-based techniques on DIBCO datasets \cite{lu2018binarization}. 

Many previous works on binarization have exploited the prior knowledge of texts in the document. The edge pixels of the texts can be extracted by techniques similar to the Canny edge detector \cite{canny1987computational}. This technique is already proposed by Chen et al. in their double threshold image binarization method \cite{chen2008double}. One generalization of edge pixels is transition pixels with extreme transition values. These pixels are calculated on a small neighborhood using the intensity difference, and then the gray-intensity threshold is calculated from the statistical information of the transition set \cite{ramirez2010transition}. On the contrary, structural symmetric pixels (SSPs) are used to determine local thresholds in a neighborhood, and a voting system is utilized for multiple thresholds for the binarization task \cite{jia2018degraded}. Automatic parameter tuning can be done by utilizing a global energy function inspired by a Markov random field model by incorporating edge discontinuities \cite{howe2013document}. All these methods are mostly based on traditional image processing and pattern recognition techniques and may have promising characteristics. However, they are designed to attain good results on certain types of documents and lack in addressing the diversified degradation problems similar to the DSS collection with its wide spectrum of writing-surface materials.

With the success of deep learning in sophisticated image understanding \cite{guo2016deep}, several neural network architectures have been proposed for handwritten document binarization and analysis. A fully convolutional network (FCN) is proposed, which operates at multiple image scales starting from full-resolution \cite{tensmeyer2017document}. A convolutional encoder-decoder is used by Peng et al. \cite{peng2017using} on the LRDE document binarization dataset \cite{lazzara2011scribo, lazzara2014efficient}. In the recent works of Calvo-Zaragoza et al., a selectional autoencoder is used for the binarization task \cite{calvo2019selectional} on a couple of DIBCO datasets \cite{ntirogiannis2014icfhr2014, pratikakis2016icfhr2016}, Balinese palm leaf manuscripts \cite{burie2016icfhr2016}, Persian documents from PHI \cite{ayatollahi2013persian}, and music notations from SAM and ES \cite{calvo2017pixel}. Afzal et al. have proposed the use of the recurrent neural network for document binarization \cite{afzal2015document}. This work has been extended by Westphal et al. \cite{westphal2018document} by using Grid Long Short-Term Memory (Grid LSTM) networks \cite{kalchbrenner2015grid} for the binarization task. A hierarchical deep supervised network (DSN) architecture is proposed, which shows better performance than the Grid LSTM on DIBCO datasets \cite{vo2018binarization}. In the recent work of He et al., an iterative deep learning technique is proposed for document enhancement and binarization \cite{he2019deepotsu} on several DIBCO datasets, the Bickley-diary dataset \cite{deng2010binarizationshop}, the PHIDB dataset \cite{nafchi2013efficient}, and the Synchromedia Multi-spectral dataset \cite{hedjam2015icdar}. 

\begin{figure}[h!]
	\centering
	\includegraphics[height=4.1cm]{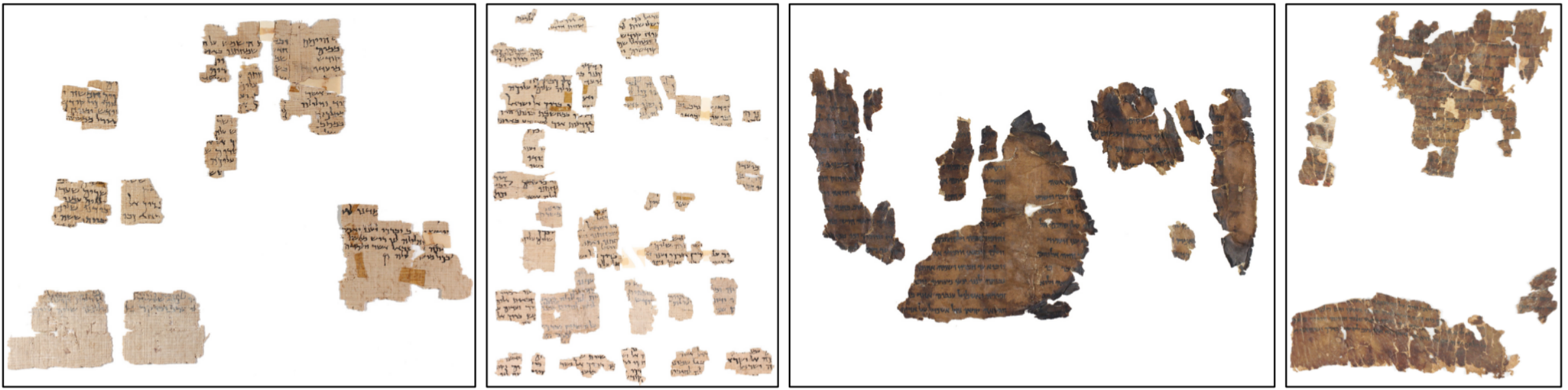}
	\caption{The evidence for ink versus background resides only partly in the intensity of an individual pixel. The evidence is heavily present in external features (of papyrus and parchment), as well. The figure shows four RGB-color images of full plates from the DSS collection (from \textit{left} to \textit{right}: $463A, 464, 1080,$ and $1082$). The first two have papyrus as a surface material for writing, and the latter two have parchment. Both the materials show distinctive degradation and decaying of characters (inks). The binarization tool should be able to find the separation of ink from the surface materials explicitly.}
	\label{fig:backgroundPapyParch}
\end{figure}

All these neural network-based techniques are useful, and they present improved performances in many cases. However, most of the time, the datasets used do not pose extreme cases of degradation along with diverse material textures like parchment and papyrus (see Figure \ref{fig:backgroundPapyParch}). A texture modeling can be performed using the Markov random field (MRF) \cite{cross1983markov}, but it will be extremely complicated in the case of DSS images. Explicit foreground and background modeling have been proposed by Sriman et al. in the classification of text blocks in the scene images \cite{sriman2015explicit}, but the binarization task needs precise localization of each of the ink pixels. On top of this, in the case of the fragment images of the DSS collection, the unnecessary elements and modern number tags should be ignored in the outcome of the binarization (as discussed in the previous section; see Figure \ref{fig:introYardeni}). The desired binarization tool, on the one hand, should be robust enough to handle extremely degraded historical manuscripts written in parchments and papyrus, like the DSS collection, and, on the other hand, should perform well in general document cases including the DIBCO images.

Although many of the previous works already provided us with different tools for benchmark datasets, a robust tool is yet to be designed that performs not only outstanding binarization for severe cases like the DSS collection but also shows consistent performance in general cases. The implementations of U-Net \cite{ronneberger2015u} and pix2pix \cite{isola2017image} methods are particularly relevant here. Though U-Net was initially designed for biomedical-image segmentation, this has been used already for accurate pixel classification in one of the recent competitions on document image binarization (DIBCO 2017 \cite{pratikakis2017icdar2017}). On the other hand, pix2pix is proposed initially as a general-purpose solution to image-to-image translation problems using conditional adversarial networks \cite{goodfellow2014generative, michelsanti2017conditional} and is not designed to perform document binarization tasks. However, the inspiration lies in the performance of the pix2pix method on image-to-image translation tasks with highly-structured graphical outputs by learning a loss adapted to the task.  The proposed method in this article is based on a similar idea where the outputs are precise and straightforward representations of highly complex inputs. Hence, the proposed model is inspired by pix2pix and is a variant of the general U-Net approach.

%% file: tex/3-methods.tex
\begin{figure}[h!]
	\centering
	\resizebox{\textwidth}{!}{%
		\includegraphics[height=4cm]{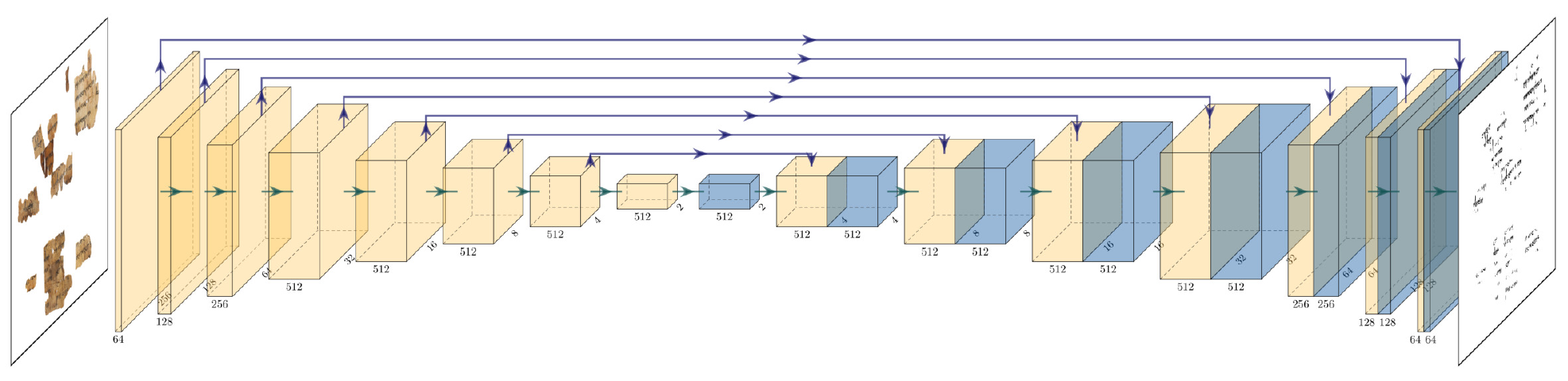}}
	\caption{The proposed network architecture shows the encoder (contracting path) at the left-half and the decoder (expanding path) at the right-half of the image. Each step in the decoder part receives a concatenation with the corresponding feature-map from the encoder part through the skip connections. This concatenation circumvents the bottle-neck issue at the deepest layers of the encoder and ensures the precise localization of the foreground-background pixels.}
	\label{fig:introNetwork}
\end{figure}

\section{Methodology}
In this section, we will briefly explain the proposed model and the complete methodology. We will present the BiNet framework, the network architecture with technical details, hyperparameters, the transfer learning techniques, and the datasets.

\subsection{BiNet}
The document image binarization task is simply a two-class classification problem. For handwritten documents, we define the two classes as foreground and background. The simple solution is to decide labels for each of the pixels of an image, one by one. The background study already shows several methods working in this direction. However, for the cases of extremely degraded historical manuscripts, this is a problematic direction, even with local thresholding in a small neighborhood. We need a trainable network that learns the target content and classifies each pixel by taking into account its neighbors in a local region. In this article, we propose a new approach utilizing a network architecture, the BiNet, that works end-to-end, providing the binarization in just one single step taking into account the knowledge of the original content. Rather than classifying each pixel separately, the BiNet efficiently works on the whole input image to generate a binarized output image of the same size.  

The implementation of BiNet is a variant of the U-Net architecture \cite{ronneberger2015u} that is capable of doing complex binarization tasks. The original U-Net was designed for biomedical image segmentation with precise localization and can be trained end-to-end with very few images. Due to these two traits, we follow the typical shape of U-Net with skip connections to build our model. Thus, our model becomes an image-generator (which shares similarity with pix2pix \cite{isola2017image} but differs in the adversarial parts) that produces a binarized image from an input DSS image. The original content (the texts; ground-truth) of a DSS image is accompanied by several factors including the texture of writing-surface, various degradation, and irrelevant materials (numbers, scale-bars, and the surface of the platform). The BiNet learns a mapping from the input image $x$ to output image $y$: 
\begin{equation}
x = y + \delta
\end{equation}
where $x$ is the DSS image as it is with degradation and other factors, $y$ is the latent original content (ink from the original writing; the ground truth), and $\delta$ is the noise that comprises all the information except the original content. The network is trained by minimizing the classification error from the $L1$-loss function:
\begin{equation}
L_{1} = \sum_{i=1}^{n} |y_{true}-y_{predicted}|
\end{equation} 
where $n$ is the number of pixels in the input image, $y_{true}$ is the ground-truth, and $y_{predicted}$ is the prediction from the network. As the binarized output is less complex than the input image, simply using the $L1$-regression is enough. Additionally, the  $L1$ loss function is less affected by the outliers, making it a preferable choice over $L2$ loss function for the DSS collection. The $L2$ loss has the tendency of producing the border effects.

\begin{table}[h!]
	\centering
	\caption{Detailed description of the BiNet architecture. Here, $Conv (f,h,w)$ denotes as a convolutional operator with $f$ filters and $h\times w$ kernel sizes; $BNorm()$ refers to batch normalization; $Dropout(r)$ denotes dropout operation with ratio of $r$ connections at each time; $LeakyReLU$ refers to leaky rectified linear unit; $Tanh$ refers to Tanh activation.}
	\label{tab:methodNetwork}
	\begin{tabular}{lllll}
		\hline \hline
		\multicolumn{1}{c}{\textbf{Original input}} & \multicolumn{1}{c}{\textbf{\begin{tabular}[c]{@{}c@{}}Input at \\ Encoder \end{tabular}}} & \multicolumn{1}{c}{\textbf{Encoding layers}}                                               & \multicolumn{1}{c}{\textbf{Decoding layers}}                                                                & \multicolumn{1}{c}{\textbf{Final output}}    \\ \hline \hline
		& 256x256                                                                                         & \begin{tabular}[c]{@{}l@{}}Conv (64,4,4,2)\\ Actv (LeakyReLU 0.2)\end{tabular}             & \begin{tabular}[c]{@{}l@{}}Conv (512,4,4,2)\\ BNorm ()\\ Actv (LeakyReLU 0.2)\\ Dropout (0.5)\end{tabular}  &                                        \\ \cline{2-4}
		& 128x128                                                                                         & \begin{tabular}[c]{@{}l@{}}Conv (128,4,4,2)\\ BNorm ()\\ Actv (LeakyReLU 0.2)\end{tabular} & \begin{tabular}[c]{@{}l@{}}Conv (1024,4,4,2)\\ BNorm ()\\ Actv (LeakyReLU 0.2)\\ Dropout (0.5)\end{tabular} &                                        \\ \cline{2-4}
		${[}0,255{]}^{256\times256}$    & 64x64                                                                                           & \begin{tabular}[c]{@{}l@{}}Conv (256,4,4,2)\\ BNorm ()\\ Actv (LeakyReLU 0.2)\end{tabular} & \begin{tabular}[c]{@{}l@{}}Conv (1024,4,4,2)\\ BNorm ()\\ Actv (LeakyReLU 0.2)\\ Dropout (0.5)\end{tabular} &                                        \\ \cline{2-4}
		or, & 32x32                                                                                           & \begin{tabular}[c]{@{}l@{}}Conv (512,4,4,2)\\ BNorm ()\\ Actv (LeakyReLU 0.2)\end{tabular} & \begin{tabular}[c]{@{}l@{}}Conv (1024,4,4,2)\\ BNorm ()\\ Actv (LeakyReLU 0.2)\end{tabular}                 &                    ${[}0,1{]}^{256\times256}$                    \\ \cline{2-4}
		${[}0,255{]}^{256\times256\times3}$    & 16x16                                                                                           & \begin{tabular}[c]{@{}l@{}}Conv (512,4,4,2)\\ BNorm ()\\ Actv (LeakyReLU 0.2)\end{tabular} & \begin{tabular}[c]{@{}l@{}}Conv (1024,4,4,2)\\ BNorm ()\\ Actv (LeakyReLU 0.2)\end{tabular}                 &  \\ \cline{2-4}
		& 8x8                                                                                             & \begin{tabular}[c]{@{}l@{}}Conv (512,4,4,2)\\ BNorm ()\\ Actv (LeakyReLU 0.2)\end{tabular} & \begin{tabular}[c]{@{}l@{}}Conv (512,4,4,2)\\ BNorm ()\\ Actv (LeakyReLU 0.2)\end{tabular}                  &                                        \\ \cline{2-4}
		& 4x4                                                                                             & \begin{tabular}[c]{@{}l@{}}Conv (512,4,4,2)\\ BNorm ()\\ Actv (LeakyReLU 0.2)\end{tabular} & \begin{tabular}[c]{@{}l@{}}Conv (256,4,4,2)\\ BNorm ()\\ Actv (LeakyReLU 0.2)\end{tabular}                  &                                        \\ \cline{2-4}
		& 2x2                                                                                             & \begin{tabular}[c]{@{}l@{}}Conv (512,4,4,2)\\ BNorm ()\\ Actv (LeakyReLU 0.2)\end{tabular} & \begin{tabular}[c]{@{}l@{}}Conv (128,4,4,2)\\ BNorm ()\\ Actv (LeakyReLU 0.2)\end{tabular}                  &                                        \\ \cline{2-4}
		&                                                                                                 &                                                                                            & \begin{tabular}[c]{@{}l@{}}Conv (1,4,4)\\ Actv (Tanh)\end{tabular}                                          &                                        \\ \hline \hline
	\end{tabular}
\end{table}

\subsection{Network Architecture}
The network architecture is illustrated in Figure \ref{fig:introNetwork}. The technical details of each of the layers can be found in Table \ref{tab:methodNetwork}. In the encoder part, we have Convolution-BatchNorm-LeakyReLU layers with different number of filters. The decoder part consists of Convolution-BatchNorm-Dropout-LeakyReLU with a $50\%$ dropout in the first three layers, then the remaining layers consist of a Convolution-BatchNorm-LeakyReLU structure. We used the leaky rectified linear unit (LeakyReLU, \cite{maas2013rectifier}) as the activation function. The convolutions are $4\times4$ spatial filters applied with stride $2$ and padding $1$. The hyperparameters are set empirically through grid search. No max-pooling layer is used. Experimental results show that BiNet is one of the optimal topologies for DSS image binarization. The model works with a fixed image size of $256\times256$ and accepts both color ($3$ channel) and grayscale ($1$ channel) images, and always outputs a $256\times256$ binary image. Please note that the document image can be much larger and of variable sizes. The implementation processes all the input images by dividing them into equal pieces of $256\times256$ patches (padding is performed if necessary) to provide the input images to the network and combine the individual outputs to get the full binarized image.

Instead of ReLU, we used LeakyReLU to avoid collapsing gradient. Parametric ReLU has the same advantage with one difference that the slope of the output for negative inputs is a learnable parameter while in the Leaky ReLU it is a hyperparameter. Convolutional layers replace max-pooling with increased stride making it a sub-sampling step. The network is large enough for the DSS images to be trained on, and it can learn all the necessary invariances without using max-pooling layers, without any loss in accuracy \cite{springenberg2014striving}. 

\subsection{Transfer Learning} \label{subsec:transferlearn}
The BiNet structure is initially proposed for binarizing the DSS fragment images.  Now, in the case of full plate images, the whole model that is pre-trained from scratch using DIBCO images can be retrained to update the weights of the network. This retraining can be done using a small number of ground-truth plate images due to the high similarity of the plate images to the DIBCO images. A BiNet architecture can also be used in different historical manuscript collections using this simple transfer learning technique with only a small amount of training data.    

\begin{figure}[h!]
	\centering
	\includegraphics[width=0.9\linewidth]{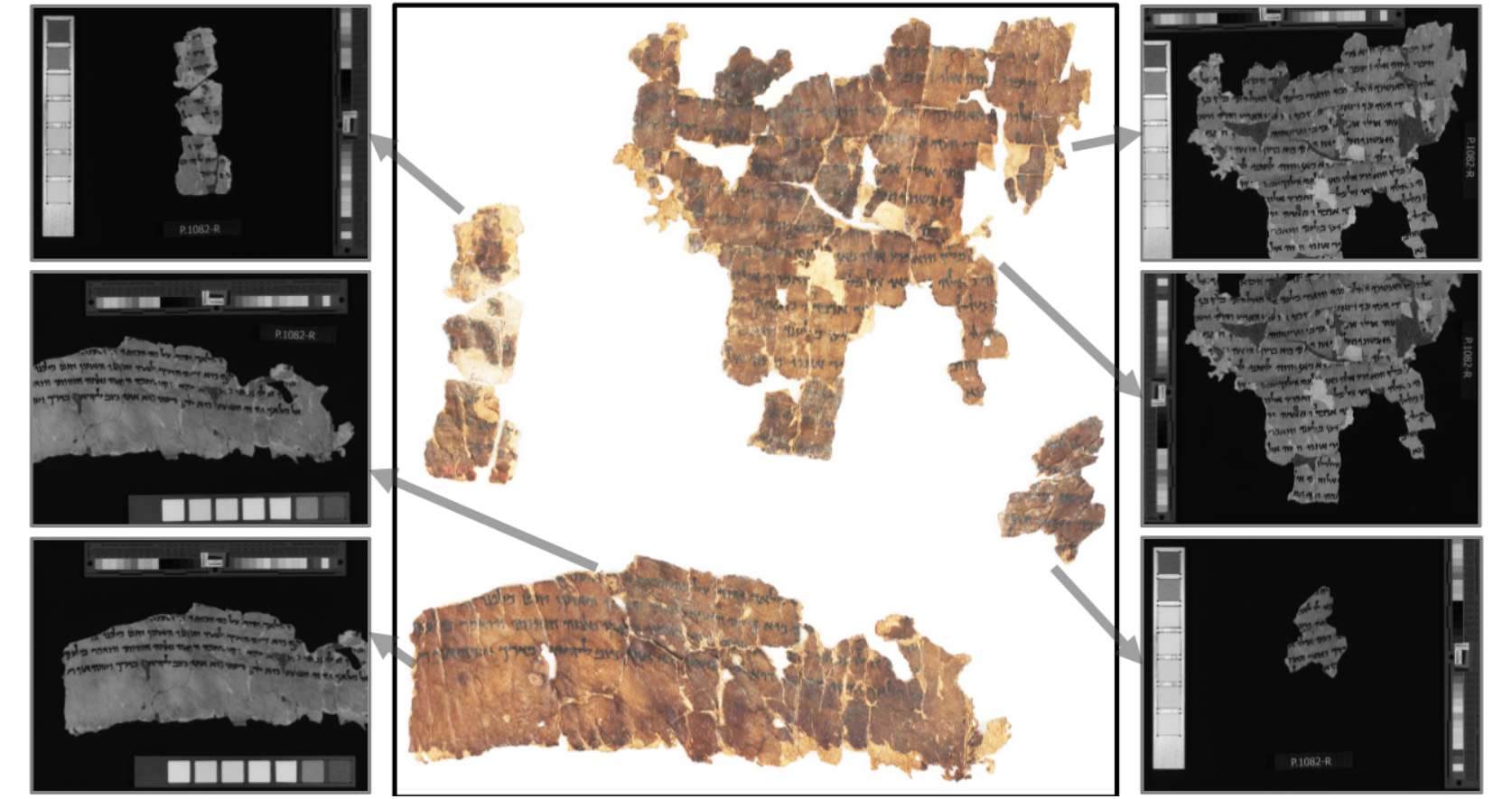}
	\caption{The RGB-color image of the full plate 1082 and the grayscale intensity images ($924nm$ wavelength) of all the four fragments of this plate. Two separate images have been captured for each of the two large fragments to cover the entire areas.}
	\label{fig:expPlateFrag}
\end{figure}

\subsection{Datasets} \label{subsec:dataset}
As our primary dataset, we will use the DSS image collection. There are several sources for the digital images of the DSS manuscripts. In this study, we will use the high-resolution multi-spectral images kindly provided to us by the Israel Antiquities Authority (IAA), which derive from their Leon Levy Dead Sea Scrolls Digital Library project. These images are offered to scholars and the general public on their website \cite{dssllnet}. For each of the scrolls fragments, the IAA produces one color-image and several multi-spectral images on both recto and verso in 28 different exposures with a resolution of $1215$ pixels per inch (PPI) at 1:1 ratio \cite{shor2014leon}. In addition to the fragment images, there are also color images of the full plates where the fragments are physically preserved (see Figure \ref{fig:expPlateFrag}). Depending on the arrangement, a full plate may contain one fragment or several different fragments. In this study, we will first use the fragment images to train and test the model. Later, we will use the plate images through the transfer learning technique (as described in \ref{subsec:transferlearn}). 

The \textbf{fragment images} have a dimension of $5412\times7216$ or $7216\times5412$ pixels, depending on the orientation of the physical fragment. We scaled this down to $50\%$, to speed up the training and testing process. The resulting dimensions are therefore $2706\times3608$ or $3608\times2706$ pixels respectively. The proposed network, BiNet, takes input image with a dimension of $256\times256$. We, therefore, divide the input image into small-images of $256\times256$ pixels. This way, we end up with $165$ small images per original image. As the dimensions of the images are not divisible by $256$, the cuts from the edges of the images have a smaller size. We change these images to size $256\times256$ by padding them. This same procedure can be followed to accommodate any image size at the input of the network. We will train three models with different inputs of fragment images, so we can see which of the three images gives the best binarization result:
\begin{itemize}
	\item RGB-color images (captured in visible light; $445nm$ wavelength)
	\item Grayscale intensity images (captured in infrared; $924nm$ wavelength)
	\item Fused images (details in Subsection \ref{subsec:fused}) 
\end{itemize}

The \textbf{plate images} have variable dimensions of \textasciitilde $3000\times4000$ pixels. They all are RGB-color images. We follow the same procedure as in the case of fragment images to produce small-images of $256\times256$ from the plate images. 

In addition to the DSS images, we will also use the publicly available \textbf{(H-)DIBCO} datasets from the document image binarization competitions of years $2009$ through $2018$ \cite{gatos2009icdar, pratikakis2010h, pratikakis2012icfhr, pratikakis2013icdar, ntirogiannis2014icfhr2014, pratikakis2016icfhr2016, pratikakis2017icdar2017, pratikakis2018icfhr2018}). Finally, in order to check the robustness of the system, we will use grid images produced from several different historical manuscripts (non-DSS) from the Monk system \cite{monknet}.

\begin{figure}[h!]
	\centering
	\includegraphics[width=\linewidth]{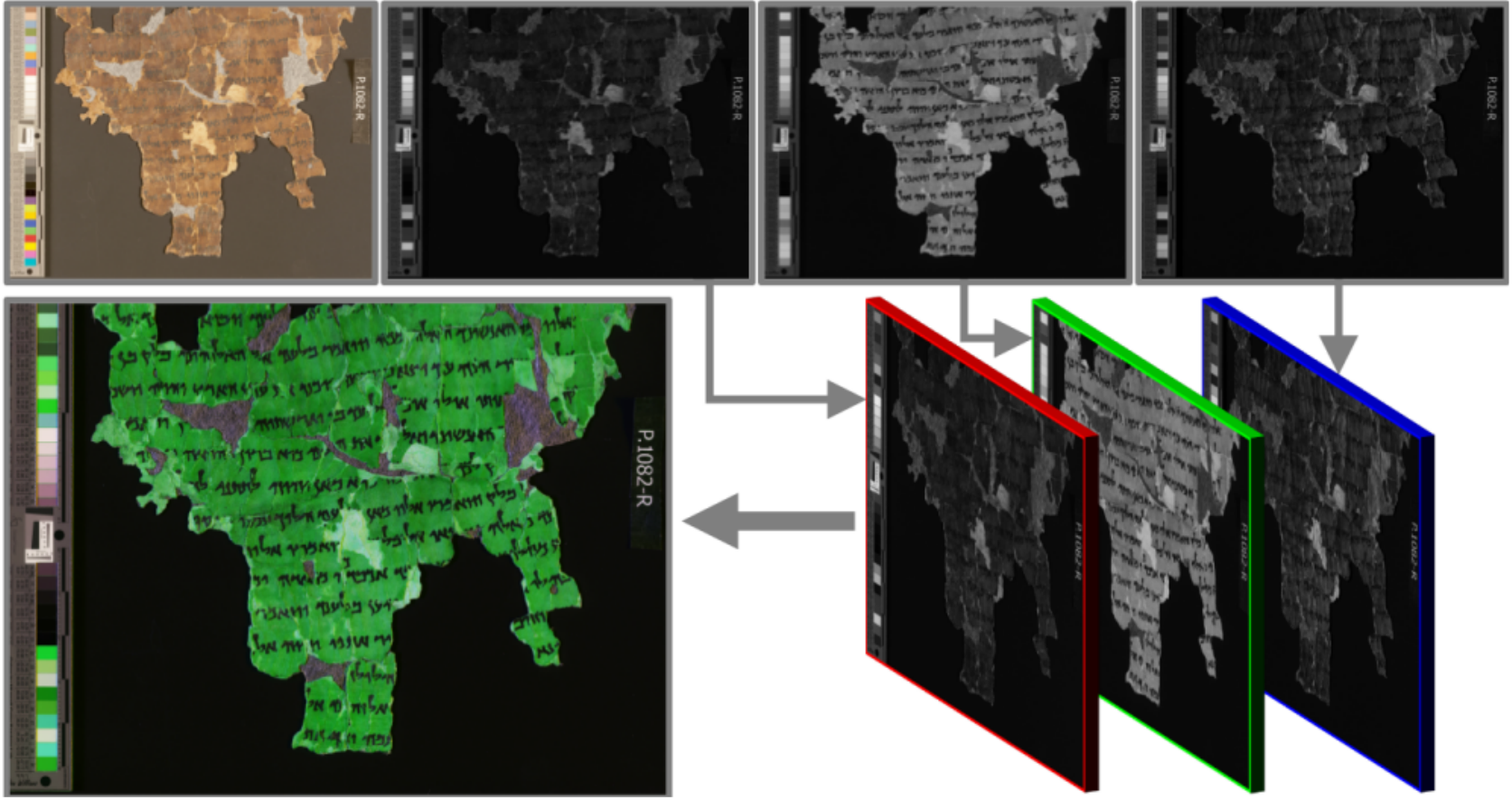}
	\caption{An illustration of the image fusion technique. In the top row: the first image from right is the RGB-color image of fragment $1$ from plate $1082$. The next three are the band images with wavelength of $595nm$, $924nm$, and $638nm$ respectively. In the bottom row: at the right side the formation of three channels from the grayscale images are shown, and at the left is the resultant pseudo-color image (fused image).}
	\label{fig:expFused}
\end{figure}

\subsection{Image Fusion} \label{subsec:fused}
We use the grayscale image resulting from the intensity of light at each pixel in the wavelength of $924nm$. This image is an interesting choice from our end. We select this particular wavelength because the resulting grayscale image shows the maximum visible contrast between the ink and the background compared to any other wavelengths. To extend our work further and to improve the ink-background separation, we take advantage of other multi-spectral band-images. We propose here an image-fusion technique to create a pseudo-color image. We take grayscale intensity images from three separate wavelengths: $595nm$, $638nm$, and $924nm$. With these three images, we produce a new fused image (pseudo-color image) with three channels. Image with wavelength $595nm$ goes to the R-channel, $924nm$  to the G-channel, and $638nm$ to the B-channel. Figure \ref{fig:expFused} shows an example of three grayscale images and the resulting pseudo-color RGB-image from image-fusion. By doing this, we hope to capture more details that emerge from various lighting conditions to improve the binarization result.

\subsection{Ground Truth} \label{subsec:gt}
One of the biggest challenges in working with DSS images is the lack of ground-truths. Besides, the DSS collection is not a structured or complete dataset, like many other historical manuscripts. In order to use trainable networks, we first need labeled train images. To create the ground-truths, we used GIMP, a free and open-source raster graphics editor (version 2.8.16 \cite{gimp}). To establish the credibility of the ground-truths, palaeographic experts labeled the images (see Section \ref{sec:acknowledge}).

We have proposed a simple method for the labeling task using the GIMP tool. First, a transparent layer is created on the top of the fragment-image that would be used for training the network. This layer is of the same dimension as the original image. Now, by zooming into the vicinity of the characters, the palaeographic experts mark the inks-pixels in red, capturing the characters from the bottom layer (original image) to the transparent layer (new image: the ground truth) with a pen of $1$ pixel accuracy. This work is similar to creating the carbon copy of a document, but the other way around. In this task, the palaeographic expert will overwrite the entire original content on the transparent layer. Due to choosing a pen size of $1$ pixel, we ensure that only the ink pixels are marked red, and everything else remains transparent. Once the task is complete, the transparent layer is taken out and saved as a separate image where the red pixels are converted to black and everything else to white (ground truth). As the task is done using a computer mouse, it is simple but time-consuming. It should be noted that the palaeographic experts were cautious to produce the ink-pixel labels as accurately as possible, conservatively avoiding to mark any non-ink pixels. This labeling task took around 4-8 hours for each of the images. We selected 51 fragment images to be labeled for this experiment. The selection was made carefully to accommodate maximum diversities to represent the whole collection (a full list is attached in the Appendix \ref{appen:train-test-DSS}).

For the plate image, we used the already labeled fragment images (for that plate) by manually putting them on a transparent layer created on top of the plate image. We used 17 plates to re-train the model (a full list is attached in the Appendix \ref{appen:train-plate}). 

\begin{figure}[h!]
	\centering
	\begin{subfigure}[]{0.33\textwidth}
		\centering
		\fbox{\includegraphics[height=3.8cm]{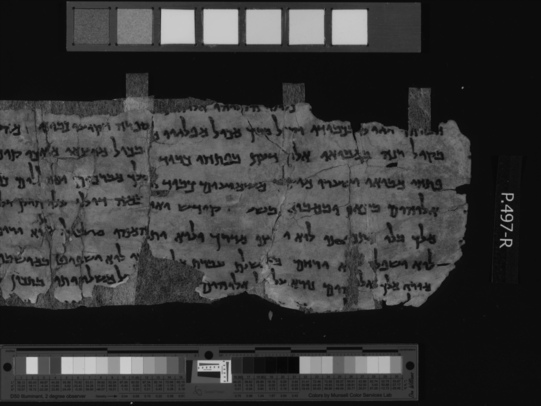}}%
		\caption{Original grayscale image}
	\end{subfigure}\hfill
	\begin{subfigure}[]{0.33\textwidth}
		\centering
		\fbox{\includegraphics[height=3.8cm]{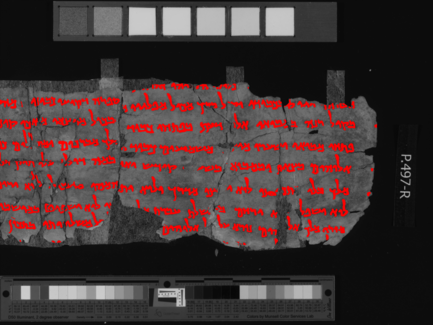}}%
		\caption{Marked ink pixels in {\color{red}red}}
	\end{subfigure}\hfill
	\begin{subfigure}[]{0.33\textwidth}
		\centering
		\fbox{\includegraphics[height=3.8cm]{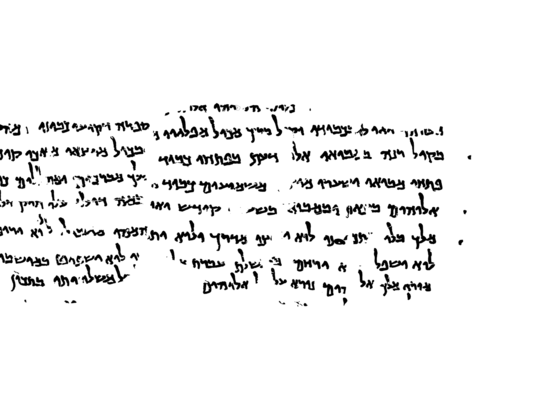}}%
		\caption{Extracted layer (ground truth)}
	\end{subfigure} \hfill
	\caption{An illustration of the procedure to create ground-truth data from the DSS fragment images using GIMP tool. The image is from fragment 9 of plate 497 (column 2, row 1).}
	\label{fig:gimpGT}
\end{figure}

%% file: tex/4-experiments.tex
\section{Experiments}
In this section, we will briefly discuss different aspects of the experiment including the training procedures and the evaluation matrices for quantitative analysis.

\subsection{Training}
From the labeled fragment images, we use $40$ images for training and $11$ for testing and evaluation. Each of these images contains one fragment and has been manually binarized. We train this network to minimize the L1-loss. For this, we use Adam optimizer with an initial learning rate of 0.0002. The model is trained for 200 epochs. We train both the proposed framework (BiNet) and the original CGAN model (from pix2pix \cite{isola2017image}) on color, grayscale, and fused fragment images separately (from scratch) as described in Subsection \ref{subsec:dataset}. Additionally, we train our model on the DIBCO datasets from scratch. We used the DIBCO datasets from 2009 to 2014 as training data and the datasets from 2016 and 2017 as test data. For plate images, we used a pre-train model and updated the weights by re-training it for another 200 epochs using the 16 manually labeled plate images. The system runs on a personal workstation with a single GPU (NVIDIA GTX 1060 with 6GB memory; details can be found in the Appendices \ref{appen:comp-config} and \ref{appen:time}).

\subsection{Evaluation Measures} \label{subsec:metrics}
For the purpose of quantitative analysis, we use evaluation metrics that are commonly used in the (H-)DIBCO \cite{gatos2009icdar, pratikakis2010h, pratikakis2012icfhr, pratikakis2013icdar, ntirogiannis2014icfhr2014, pratikakis2016icfhr2016, pratikakis2017icdar2017}. The metrics are suitable in the context of document analysis. We will use four metrics: F-measure ($F$), pseudo-F-measure ($F_{ps}$), peak signal-to-noise ratio ($PSNR$), and distance reciprocal distortion (DRD). Brief descriptions for each of them are provided in the following subsections.

\subsubsection{F-measure}
F-measure (also F1-score or F-score) is the measure of a test's accuracy. It is defined as:
\begin{equation} \label{eq:fm}
Fmeasure = \frac{2 \times Recall \times Precesion}{Recall + Precision}
\end{equation}
where, $Recall = \frac{TP}{TP+FN}$ and $Precision = \frac{TP}{TP+Fp}$; $TP$, $FP$, $FN$ refer to the True Positive, False Positive, and False Negative values, respectively.

\subsubsection{pseudo-F-measure}
Pseudo F-measure follows the same formula of F-measure (Eq. \ref{eq:fm}), but it uses pseudo-Recall and pseudo-precision \cite{ntirogiannis2012performance}.  
\begin{equation} \label{eq:fm}
pseudoFmeasure = \frac{2 \times pseudoRecall \times pseudoPrecesion}{pseudoRecall + pseudoPrecision}
\end{equation}
Both these pseudo metrics use distance weights with respect to the contour of the ground-truth characters. In the case of pseudo-Recall, the weights of the ground-truth inks are normalized according to the local stroke width. These weights are defined between [$0,1$]. In the case of pseudo-Precision, the weights are constrained within an area that expands to the background of the ground-truth taking into account the stroke width of the nearest ground-truth component. Inside this area, the weights are greater than one (generally between [$1,2$]) while outside this area they are equal to one.

\subsubsection{Peak Signal-to-Noise Ratio (PSNR)}
Peak signal-to-noise ratio is the measure of how close an image is to another one. The higher the value of PSNR, the higher the similarity of the two images that are being compared.
\begin{eqnarray}
PSNR = 10\log (\frac{C^2}{MSE})
\end{eqnarray}
where, $C$ is the difference between foreground and background, and $MSE$ is the the mean squared error and defined as:
\begin{eqnarray}
MSE = \frac{\sum_{x=1}^{M}\sum_{y=1}^{N} (I_{bin}(x,y) - I^{'}_{bin}(x,y))^2}{MN}
\end{eqnarray} 
where, $M$ x $N$ is the dimension of the image, $I_{bin}$ is the original ground-truth image and $I^{'}_{bin}$ is the test-output of the ground-truth from the model.

\subsubsection{Distance Reciprocal Distortion (DRD)}
The distance reciprocal distortion metric correlates with the human visual perception system and measures the distortion for all the $S$-flipped pixels as:
\begin{eqnarray}
DRD = \frac{\sum_{k=1}^{S} {DRD}_{k}}{NUBN}
\end{eqnarray}
where, ${DRD}_{k}$ is the distortion of the $k$-th flipped pixel that is calculated using $5$ x $5$ normalized weight matrix. The weight matrix $W_{Nm}$ is defined by Lu et al. \cite{lu2004distance}, where they used the DRD to measure the visual distortion in binary document images. $NUBN$ is the number of non-uniform $8$ x $8$ blocks in the ground-truth image.

%% file: tex/5-results.tex
\section{Results}
In this section, we present the experimental results based on four distinct evaluation measures presented in Subsection \ref{subsec:metrics}. The results are obtained using three different test sets. The \textbf{first test set} consists of $11$ fragment images of the DSS collection for which the corresponding ground truths were built manually, as described in Subsection \ref{subsec:gt}. The selection of the test images was made to accommodate maximum diversity and different degradation (a list is attached in Appendix \ref{appen:train-test-DSS}). The \textbf{second test set} contains $40$ images from H-DIBCO 2016, DIBCO 2017, and H-DIBCO 2018 datasets. Both H-DIBCO 2016 and 2018 have ten handwritten document images each, and DIBCO 2017 has twenty document images: ten machine-printed and ten handwritten. Finally, the \textbf{third test set} consists of $3$ RGB-colored full plate images from the DSS collection.%

\begin{figure}[H]
	\centering
	\begin{subfigure}{.2\textwidth}
		\fbox{\includegraphics[height=2.2cm]{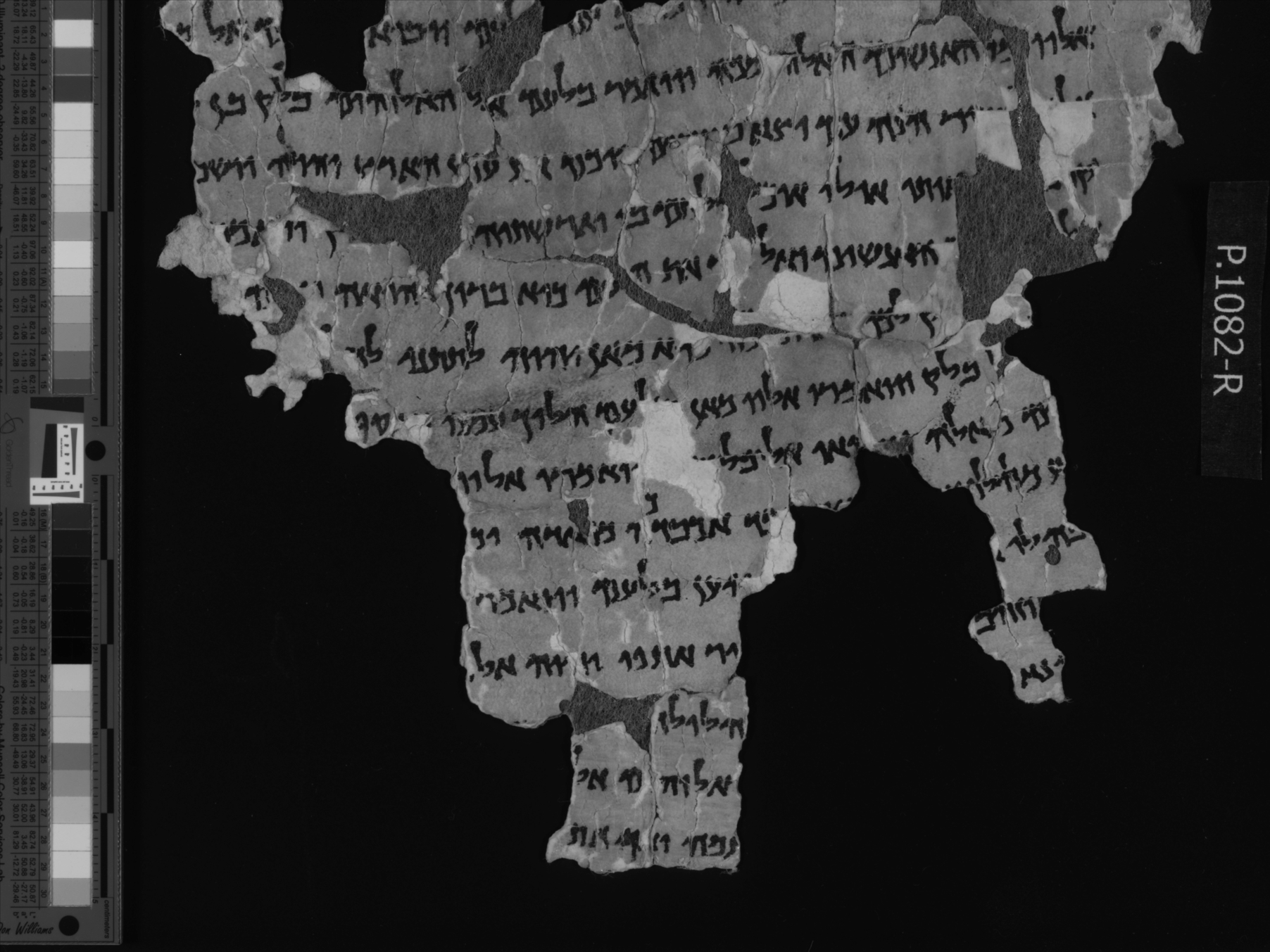}}
		\caption{Original grayscale}
	\end{subfigure}\hfill
	\begin{subfigure}{.2\textwidth}
		\fbox{\includegraphics[height=2.2cm]{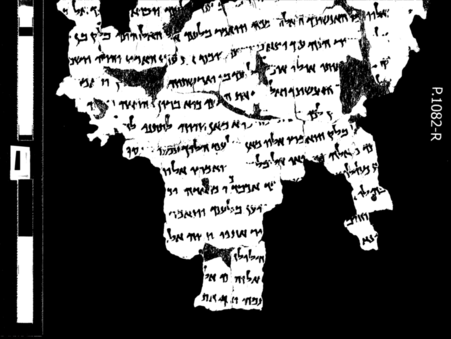}}
		\caption{Otsu}
	\end{subfigure}\hfill
	\begin{subfigure}{.2\textwidth}	
		\fbox{\includegraphics[height=2.2cm]{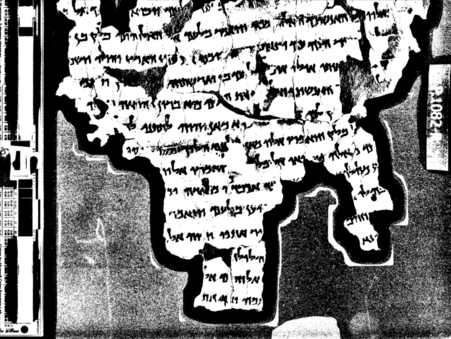}}
		\caption{Niblack}
	\end{subfigure}\hfill
	\begin{subfigure}{.2\textwidth}	
		\fbox{\includegraphics[height=2.2cm]{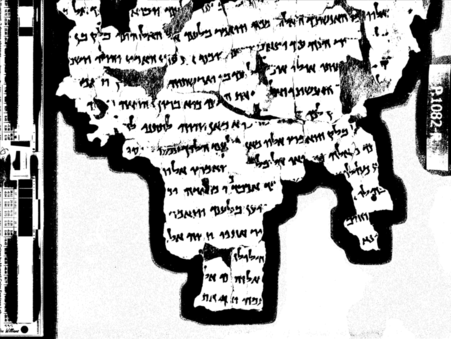}}
		\caption{Sauvola}
	\end{subfigure}\hfill
	\begin{subfigure}{.2\textwidth}	
		\fbox{\includegraphics[height=2.2cm]{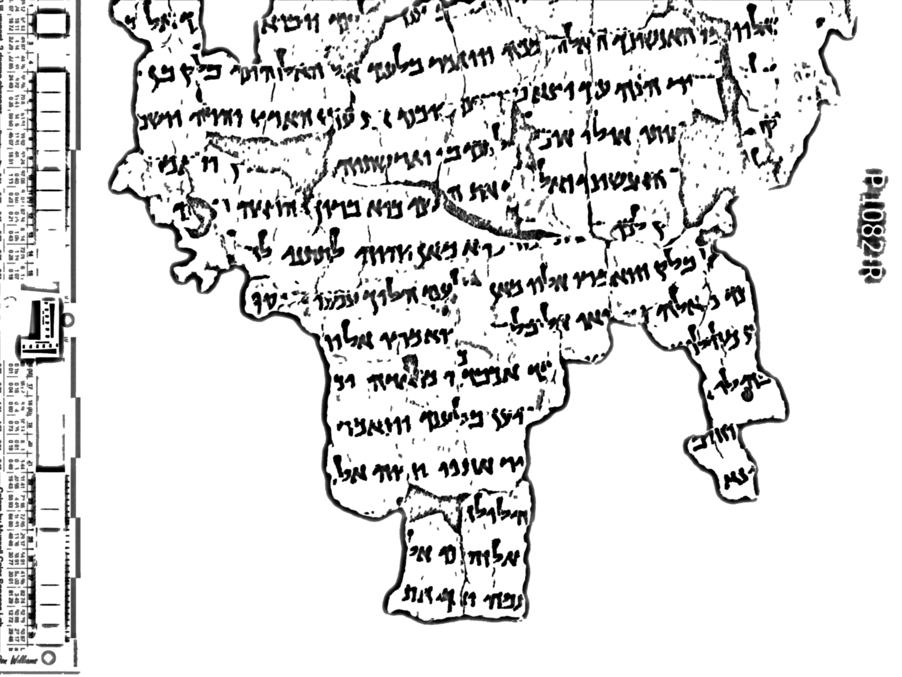}}
		\caption{Otsu (local)}	
	\end{subfigure}
	\begin{subfigure}{.2\textwidth}
		\fbox{\includegraphics[height=2.2cm]{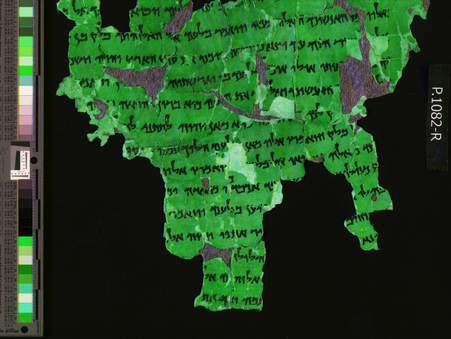}}
		\caption{Original fused}
	\end{subfigure}\hfill
	\begin{subfigure}{.2\textwidth}
		\fbox{\includegraphics[height=2.2cm]{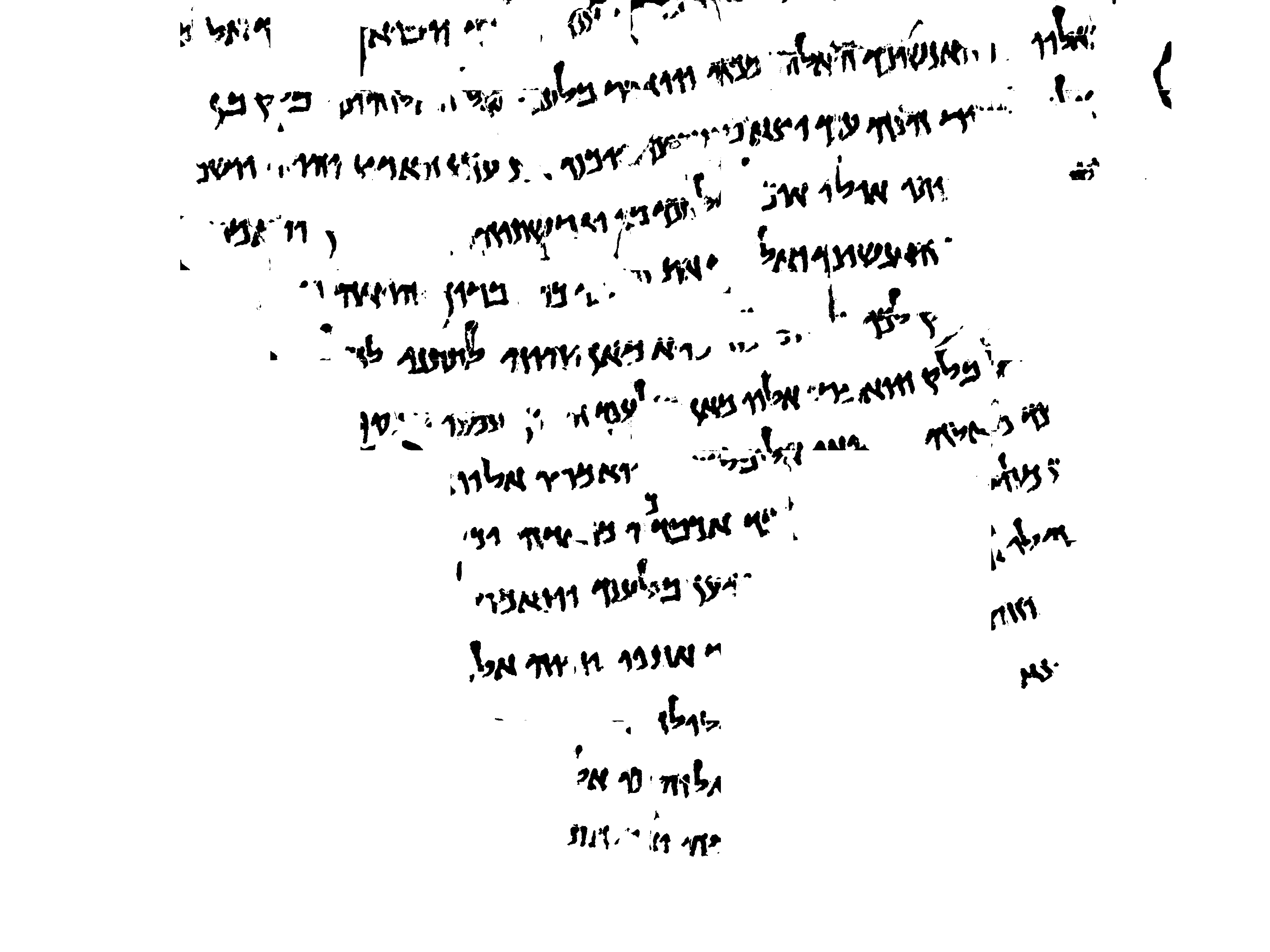}}
		\caption{CGAN (fused)}
	\end{subfigure}\hfill
	\begin{subfigure}{.2\textwidth}	
		\fbox{\includegraphics[height=2.2cm]{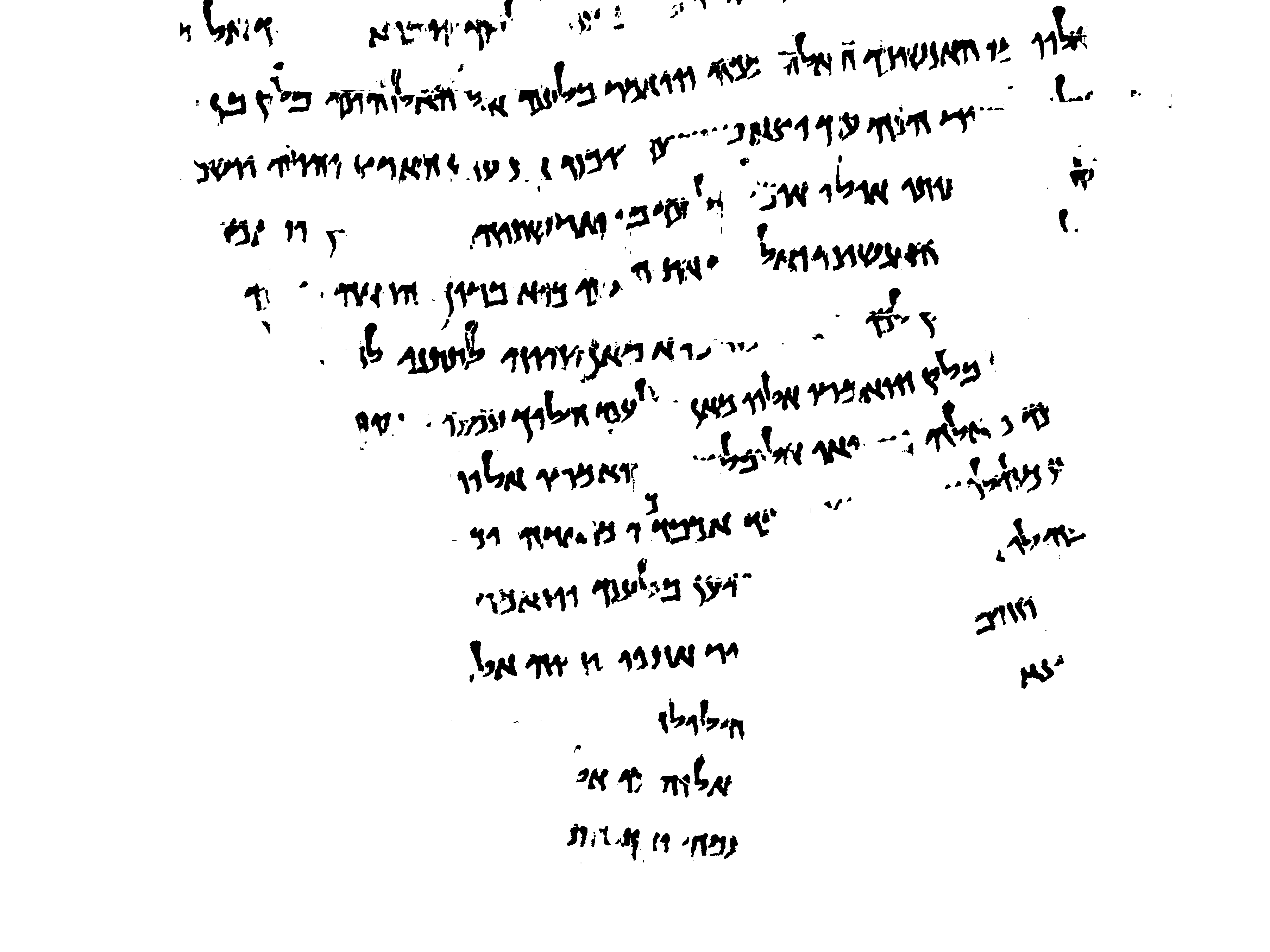}}
		\caption{BiNet (color)}
	\end{subfigure}\hfill
	\begin{subfigure}{.2\textwidth}	
		\fbox{\includegraphics[height=2.2cm]{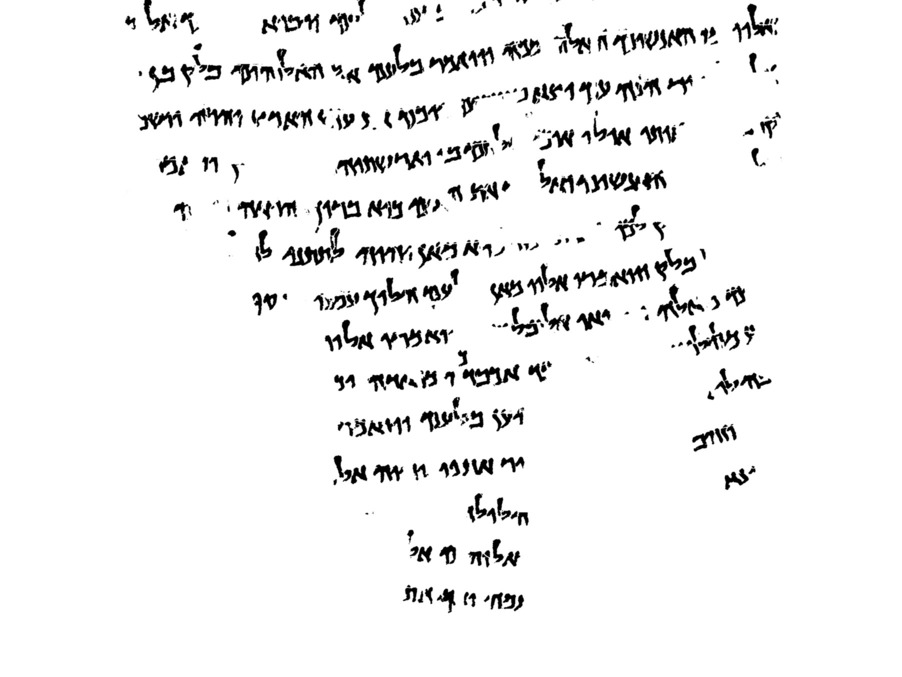}}
		\caption{BiNet (fused)}
	\end{subfigure}\hfill
	\begin{subfigure}{.2\textwidth}	
		\fbox{\includegraphics[height=2.2cm]{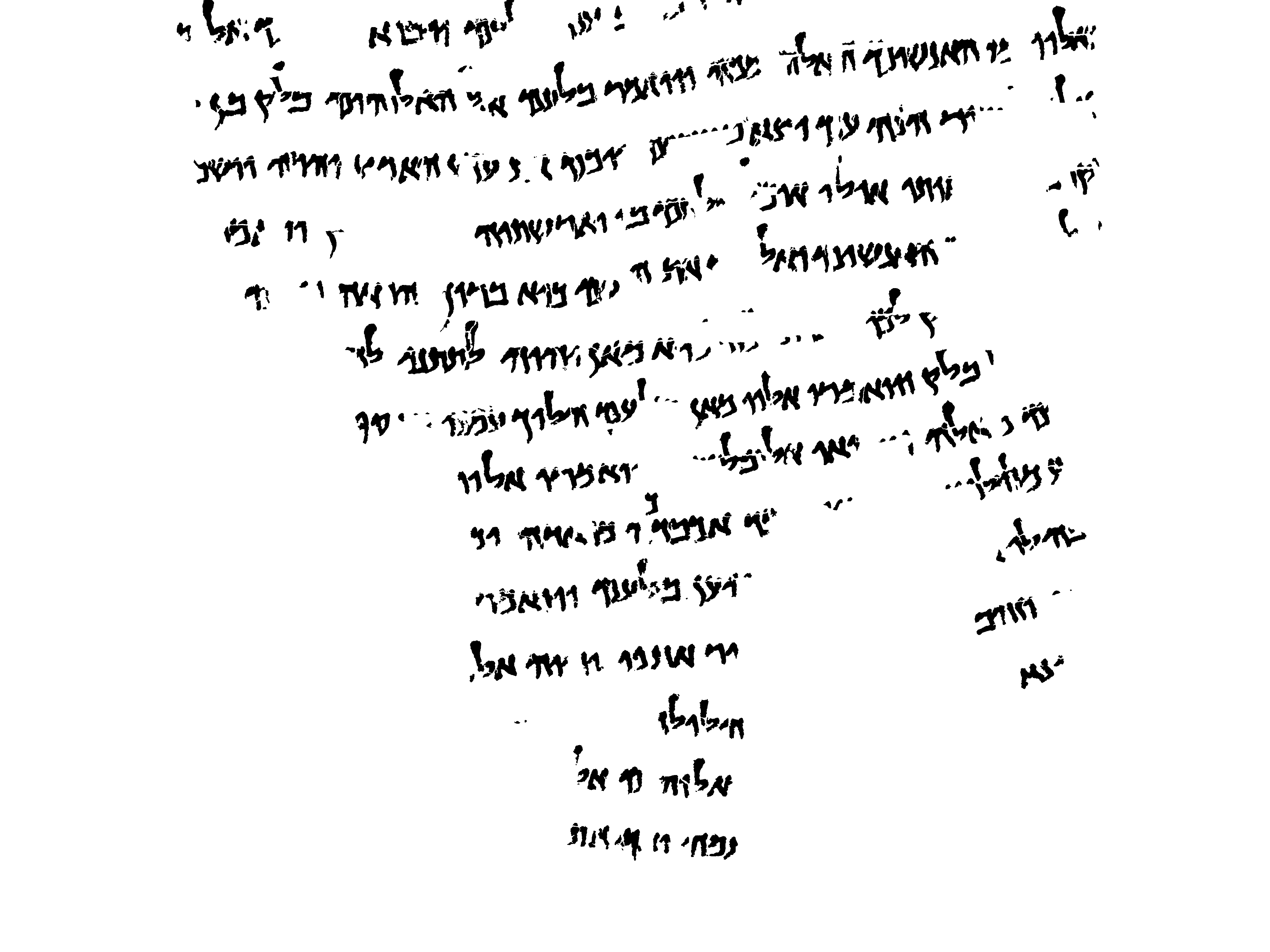}}
		\caption{Ground truth}	
	\end{subfigure}
	\caption{A comparative illustration of test results using different traditional methods along with the proposed BiNet model on fragment 1 of plate 1082. Please note the successful removal of the color-calibrator strip and machine-printed number-tag.}
	\label{fig:resultfrag}
\end{figure}%

\begin{figure}[h!]
	\centering
	\begin{subfigure}{.2\textwidth}
		\fbox{\includegraphics[height=3cm]{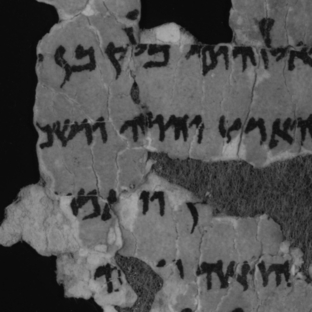}}
		\caption{Original grayscale}
	\end{subfigure}\hfill
	\begin{subfigure}{.2\textwidth}
		\fbox{\includegraphics[height=3cm]{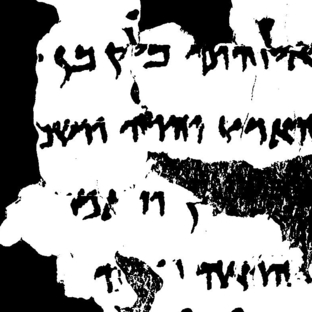}}
		\caption{Otsu}
	\end{subfigure}\hfill
	\begin{subfigure}{.2\textwidth}	
		\fbox{\includegraphics[height=3cm]{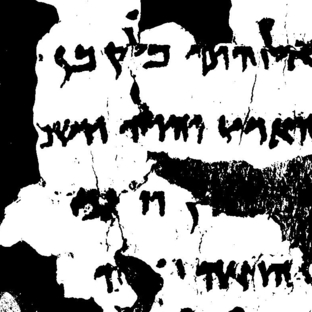}}
		\caption{Niblack}
	\end{subfigure}\hfill
	\begin{subfigure}{.2\textwidth}	
		\fbox{\includegraphics[height=3cm]{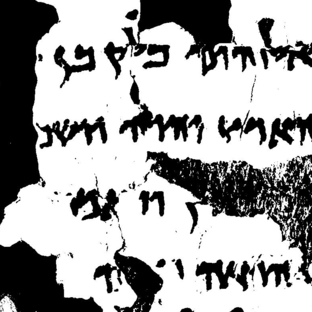}}
		\caption{Sauvola}
	\end{subfigure}\hfill
	\begin{subfigure}{.2\textwidth}	
		\fbox{\includegraphics[height=3cm]{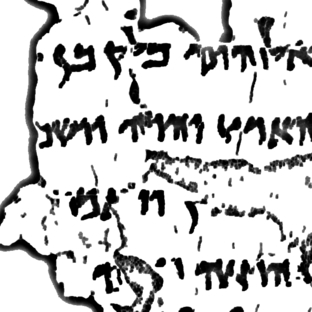}}
		\caption{Otsu (local)}	
	\end{subfigure}
	\begin{subfigure}{.2\textwidth}
		\fbox{\includegraphics[height=3cm]{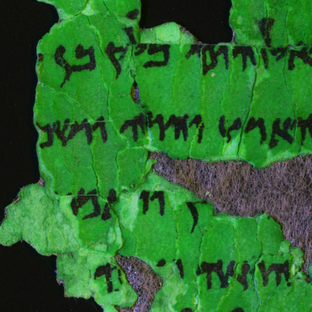}}
		\caption{Original fused}
	\end{subfigure}\hfill
	\begin{subfigure}{.2\textwidth}
		\fbox{\includegraphics[height=3cm]{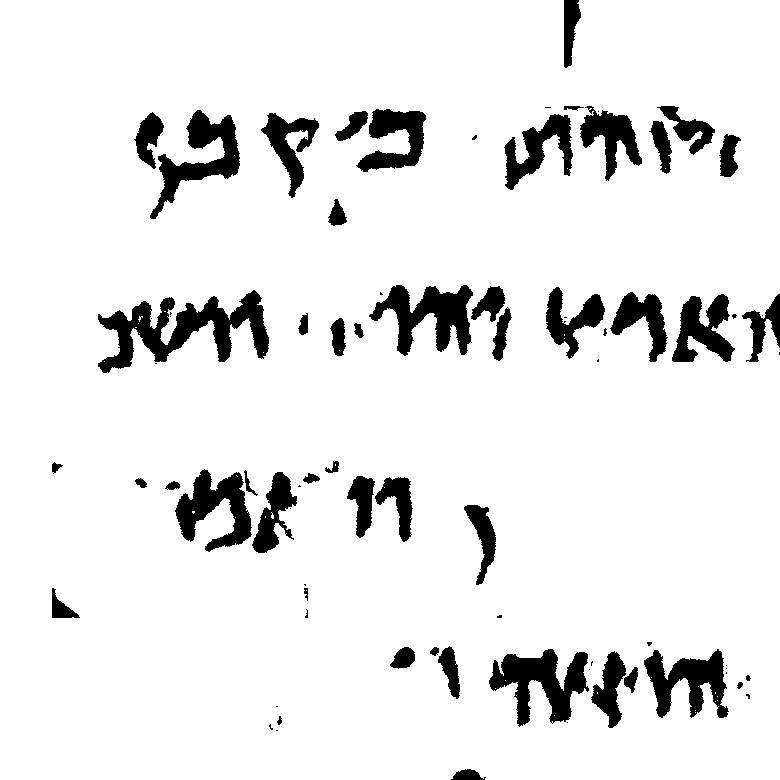}}
		\caption{CGAN (fused)}
	\end{subfigure}\hfill
	\begin{subfigure}{.2\textwidth}	
		\fbox{\includegraphics[height=3cm]{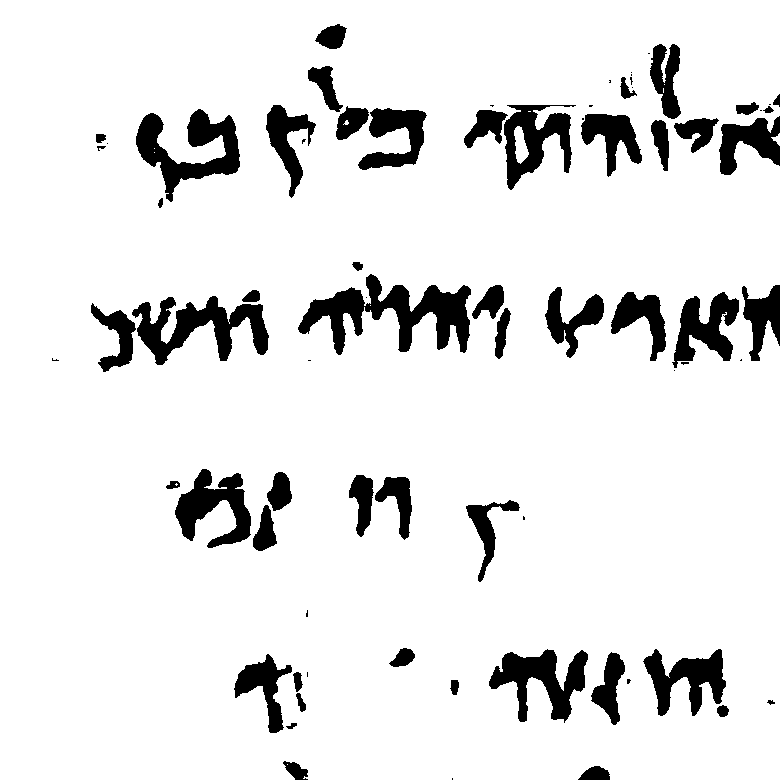}}
		\caption{BiNet (color)}
	\end{subfigure}\hfill
	\begin{subfigure}{.2\textwidth}	
		\fbox{\includegraphics[height=3cm]{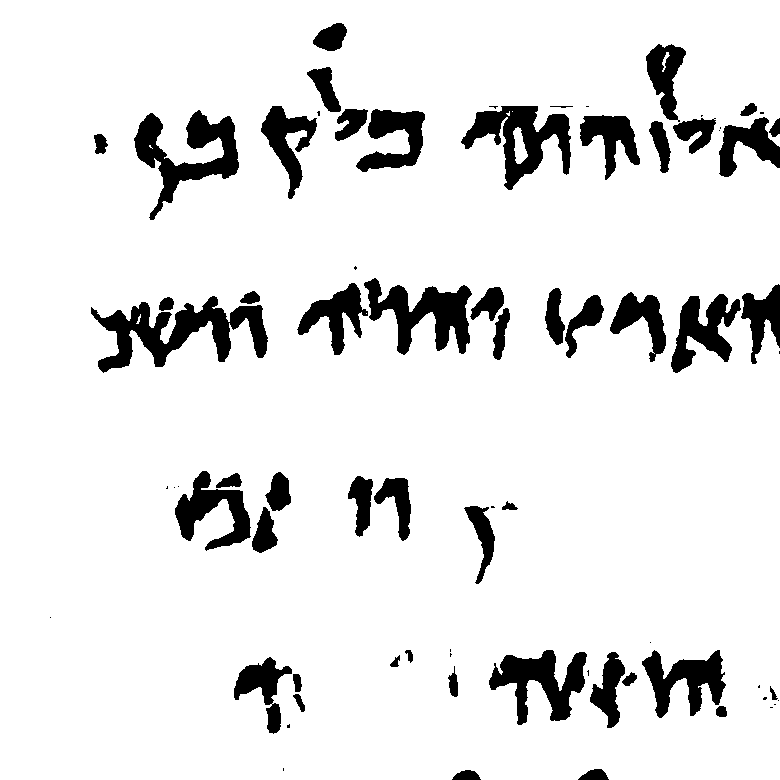}}
		\caption{BiNet (fused)}
	\end{subfigure}\hfill
	\begin{subfigure}{.2\textwidth}	
		\fbox{\includegraphics[height=3cm]{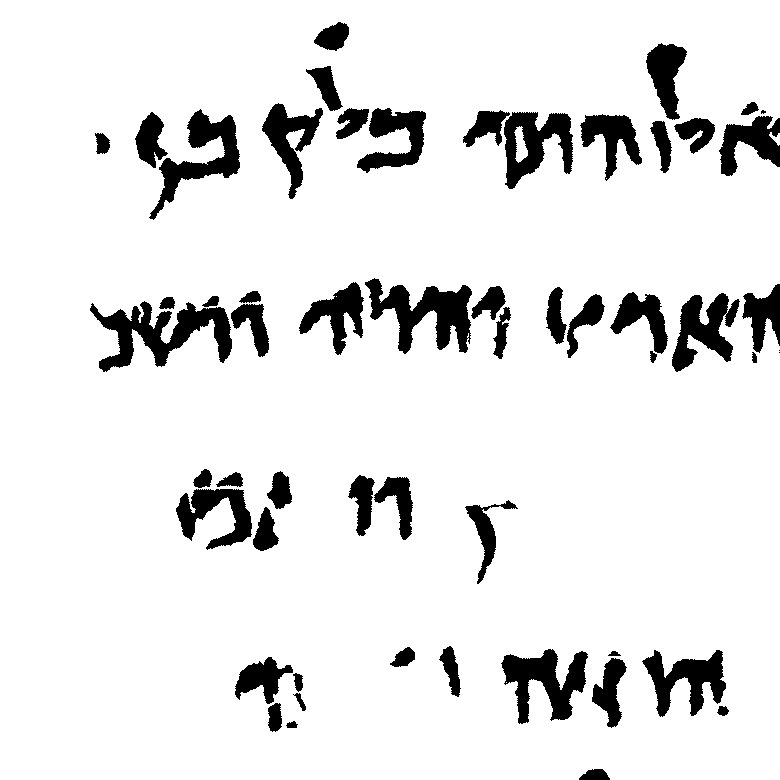}}
		\caption{Ground truth}	
	\end{subfigure}
	\caption{Results of BiNet and traditional methods presented on a zoomed-in part of fragment 1 of plate 1082. The zoomed-in (enlarged) part is taken from the pixel position of $(460,150)$ with a window size of $780 \times 780$ pixels. A visual inspection shows that the BiNet (fused) output is the closest match to the ground truth.}
	\label{fig:resultzoom}
\end{figure}%

For the \textbf{first test-set}, we trained two different models: the proposed BiNet and the conditional adversarial network (CGAN) as proposed in pix2pix (the image-to-image translation work \cite{isola2017image}). The reason behind training a CGAN is to evaluate the potential of the adversarial network on learning complex tasks with smaller training data. Additionally, it is worthwhile to present the results of CGAN as the BiNet shares similar generative architecture. Both the CGAN and BiNet model are trained on three different types of images: grayscale, color, and fused (pseudo-color). Thus, we tested the first set of images on six different networks (2 models trained from scratch on three different types of training data). 

In order to present a quantitative evaluation, we use four traditional thresholding methods to perform binarization: Otsu \cite{otsu1979threshold}, Niblack \cite{niblack1985introduction}, Sauvola \cite{sauvola2000adaptive}, and a local implementation of Otsu with $70\times70$ windows. The quantitative results for the first set, the DSS images, are presented in Table \ref{tableresult:dssfrag}. The CGAN models show promising performance compared to the traditional methods. But all the BiNet models outperform the rest. BiNet on fused is the best performing model in all the four evaluation measures. 
\begin{table}[h!]
	\centering
	\caption{Detailed evaluation results on the DSS fragment images. The results are presented as $mean \pm std. dev.$ on the whole test set. The proposed BiNet outperforms other methods (best performance in \textbf{bold}).}
	\label{tableresult:dssfrag}	
	\begin{tabular}{lllll}
		\hline \hline
		\textbf{Method}            & \textbf{F-measure} & \textbf{pF-measure} & \textbf{PSNR} & \textbf{DRD}     \\ \hline
		Otsu (global) on grayscale & 7.3 $\pm$3.5       & 7.3 $\pm$3.5        & 1.6 $\pm$0.5  & 1236 $\pm$644.8  \\ \hline
		Niblack on grayscale       & 15.7 $\pm$8.8      & 15.9 $\pm$9.1       & 4.9 $\pm$1.0  & 589.4 $\pm$317.7 \\ \hline
		Sauvola on grayscale       & 19.1 $\pm$10.3     & 19.5 $\pm$10.7      & 6.3 $\pm$1.1  & 429.5 $\pm$228.1 \\ \hline
		Otsu (local) on grayscale  & 51.3 $\pm$13.4     & 54.9 $\pm$13.6      & 14.8 $\pm$1.2 & 53.7 $\pm$25.8   \\ \hline
		CGAN on grayscale          & 63.6 $\pm$16.6     & 65.1 $\pm$15.7      & 17.3 $\pm$2.2 & 29.8 $\pm$20.2   \\ \hline
		CGAN on fused              & 68.4 $\pm$18.2     & 70.5 $\pm$16.7      & 17.8 $\pm$2.3 & 28.6 $\pm$23.1   \\ \hline
		CGAN on color             & 71.5 $\pm$8.7      & 72.9 $\pm$8.4       & 17.2 $\pm$2.0 & 28.9 $\pm$15.2   \\ \hline
		BiNet on grayscale         & 80.3 $\pm$20.7     & 82.6 $\pm$19.2      & 20.5 $\pm$3.8 & 18.7 $\pm$24.6   \\ \hline
		BiNet on color            & 83.5 $\pm$9.9      & 85.8 $\pm$9.0       & 20.3 $\pm$2.9 & 15.2 $\pm$14.4   \\ \hline
		\textbf{BiNet on fused}            & \textbf{86.7} $\pm$\textbf{9.4}      & \textbf{89.3} $\pm$\textbf{8.3}       & \textbf{21.3} $\pm$\textbf{3.4} & \textbf{13} $\pm$\textbf{14.8}     \\ \hline \hline
	\end{tabular}
\end{table}%

\begin{table}[h!]
	\centering
	\caption{Detailed evaluation results on (H-)DIBCO datasets. The BiNet performs equally well with the best-performing methods from each of the competitions (best performance in \textbf{bold}). The output images are attached in the Appendices \ref{appen:dibco2018}, \ref{appen:dibco2017}, and \ref{appen:dibco2016}.}
	\label{tableresult:dibco}
	\begin{tabular}{lllll}
		\hline \hline
		& \textbf{F-measure} & \textbf{pF-measure} & \textbf{PSNR} & \textbf{DRD} \\ \hline
		\multicolumn{5}{c}{H-DIBCO 2016}                                                                                                                    \\ \hline
		Otsu (global)                                                             & 86.7               & 90                  & 17.8          & 5.5          \\ \hline
		Niblack                                                                   & 56.2               & 56.3                & 9.6           & 57.8         \\ \hline
		Sauvola                                                                   & 79.9               & 81.7                & 14.8          & 11.4         \\ \hline
		Best method at H-DIBCO'16 \cite{pratikakis2016icfhr2016}  &\textbf{ 87.6}               & \textbf{91.3}                & 18.1          & 5.2          \\ \hline
		BiNet                                                                     & 85.6 ({\color{red}-2.0})                & 90.7 ({\color{red}-0.6})                & \textbf{18.3} ({\color{green}+0.2})          & \textbf{4.9} ({\color{green}-0.3})          \\ \hline \hline
		\multicolumn{5}{c}{DIBCO 2017}                                                                                                                      \\ \hline
		Otsu (global)                                                             & 77.7               & 80.1                & 13.8          & 15.8         \\ \hline
		Niblack                                                                   & 57.4               & 57.6                & 8.8           & 44           \\ \hline
		Sauvola                                                                   & 80                 & 83                  & 14.3          & 9.6          \\ \hline
		Best method at DIBCO'17 \cite{pratikakis2017icdar2017}    & \textbf{91.0}               	   & 92.9                & \textbf{18.3}          & \textbf{3.4}          \\ \hline
		BiNet                                                                     & 90.9 ({\color{red}-0.1})               & \textbf{93.3} ({\color{green}+0.4})                & \textbf{18.3} ({\color{green}+0.0})         & 3.6 ({\color{red}+0.2})          \\ \hline \hline
		\multicolumn{5}{c}{H-DIBCO 2018}                                                                                                                    \\ \hline
		Otsu (global)                                                             & 51.4               & 53.4                & 9.7           & 59.5         \\ \hline
		Niblack                                                                   & 45.7               & 45.9                & 7.7           & 80.8         \\ \hline
		Sauvola                                                                   & 56.3               & 58.7                & 10.9          & 36           \\ \hline
		Best method at H-DIBCO'18 \cite{pratikakis2018icfhr2018}  & \textbf{88.3}              & \textbf{90.2}                & \textbf{19.1}          & \textbf{4.9}          \\ \hline
		BiNet                                                                     & 84.7 ({\color{red}-3.6})               & 87.1 ({\color{red}-3.1})                & 17.4 ({\color{red}-1.7})          & 7.5 ({\color{red}+2.6})          \\ \hline \hline
	\end{tabular}
\end{table}%

\begin{table}[h!]
	\centering
	\caption{Detailed evaluation results on the DSS full plate images. The results are presented as $mean \pm std. dev.$ on the whole test set (best performance in \textbf{bold}). The output images are attached in the Appendix \ref{appen:fig:plateimages}.}
	\label{tableresult:dssplate}	
	\begin{tabular}{lllll}
		\hline \hline
		\textbf{Method} & \textbf{F-measure}      & \textbf{pF-measure}     & \textbf{PSNR}           & \textbf{DRD}           \\ \hline
		Otsu            & 27.2 $\pm$ 5.8          & 27.2 $\pm$ 5.8          & 9.5 $\pm$ 2.1           & 206.4 $\pm$ 101.8      \\ \hline
		Niblack         & 7.3 $\pm$ 5.9           & 7.3 $\pm$ 5.9           & 1.9 $\pm$ 1.4           & 1375.4 $\pm$ 872.7     \\ \hline
		Sauvola         & 47.9 $\pm$ 16.1         & 48 $\pm$ 16.2           & 13.8 $\pm$ .08          & 80.9 $\pm$ 43.6        \\ \hline
		\textbf{BiNet}  & \textbf{86.9 $\pm$ 3.1} & \textbf{91.1 $\pm$ 2.4} & \textbf{22.9 $\pm$ 2.6} & \textbf{6.3 $\pm$ 0.2} \\ \hline \hline
	\end{tabular}
\end{table}
For the \textbf{second test set} of (H-)DIBCO images, we used the BiNet model only. The quantitative results of the second set are shown in Table \ref{tableresult:dibco}. For the (H-)DIBCO results, we also present the performances of the winning methods from each year. Though our model is originally designed for DSS images, it shows high performance in the case of (H-)DIBCO images. For the datasets of 2016 and 2017, BiNet outperforms the best-performing method in a couple of evaluation measures. For the cases of 2018-dataset, BiNet is almost as good as the best-performing method (Table \ref{tableresult:dibco}).

For the \textbf{third test set} of RGB-colored full plate images, we used the transfer learning technique on the BiNet model that was initially trained on (H-)DIBCO images. The results are presented in Table \ref{tableresult:dssplate}. The BiNet outperforms the traditional methods by large differences and obtains high evaluation measures. This performance shows the excellent usability of the system on different datasets with a small number of training data by transfer learning techniques. 
\begin{figure}[h!]
	\centering
	\begin{subfigure}[]{0.24\textwidth}
		\centering
		\fbox{\includegraphics[height=4.8cm]{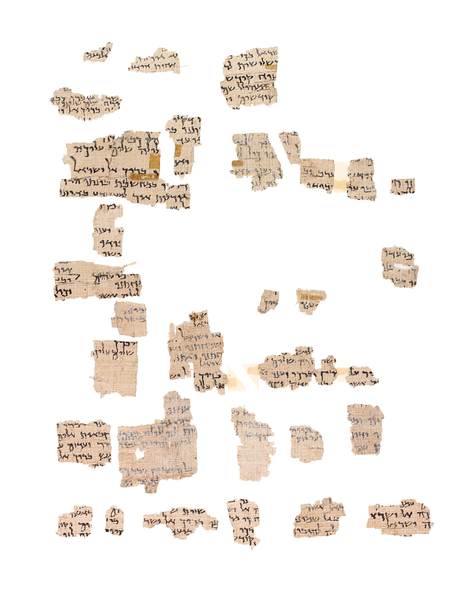}}%
		\caption{Plate \textit{464}}
	\end{subfigure}\hfill
	\begin{subfigure}[]{0.24\textwidth}
		\centering
		\fbox{\includegraphics[height=4.8cm]{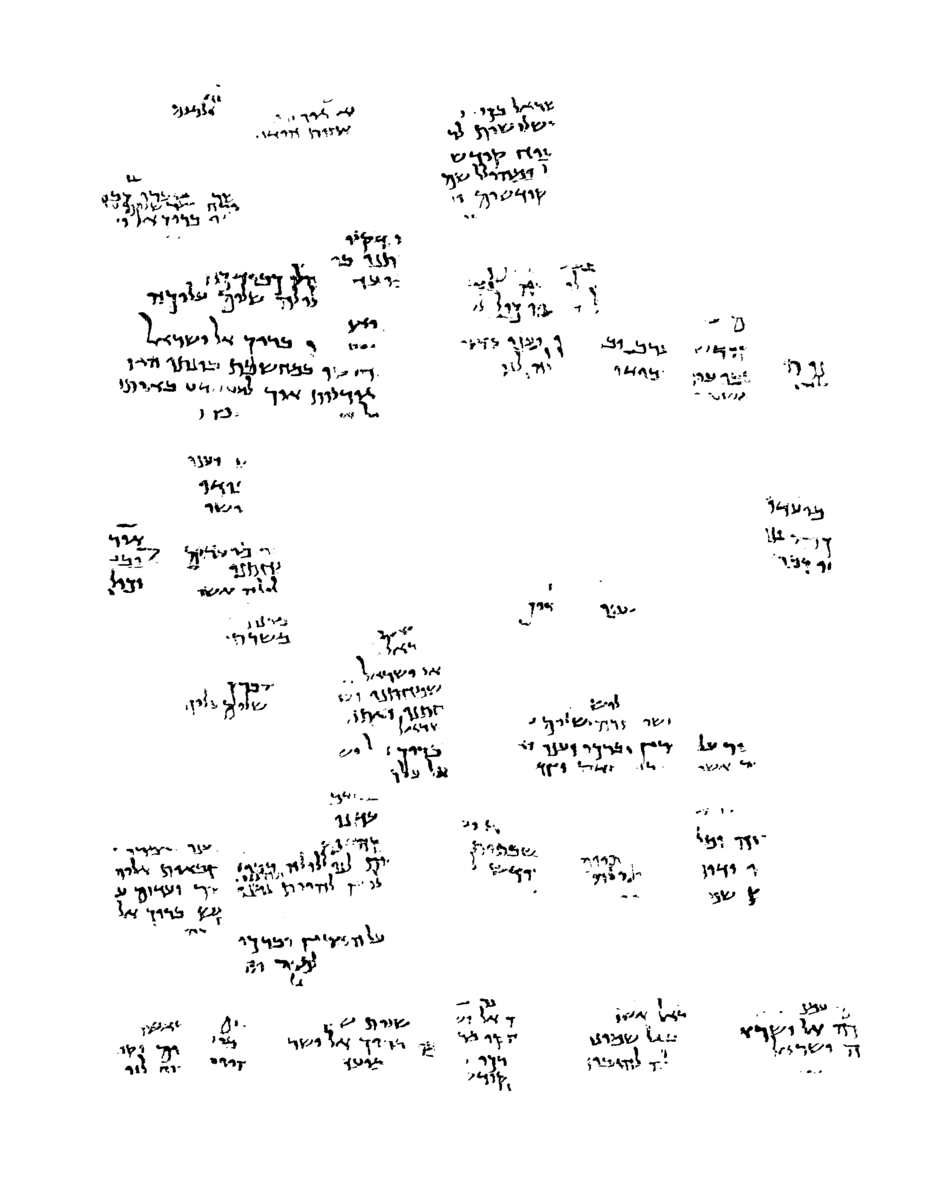}}%
		\caption{BiNet}
	\end{subfigure}\hfill
	\begin{subfigure}[]{0.24\textwidth}
		\centering
		\fbox{\includegraphics[height=4.8cm]{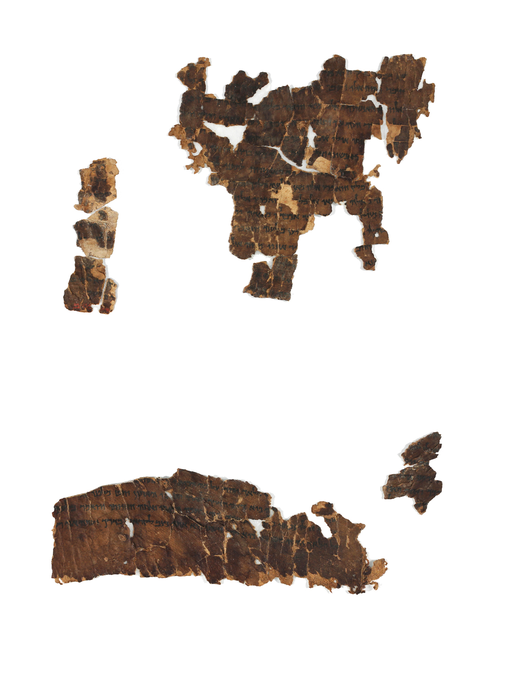}}%
		\caption{Plate \textit{1082}}
	\end{subfigure} \hfill
	\begin{subfigure}[]{0.24\textwidth}
		\centering
		\fbox{\includegraphics[height=4.8cm]{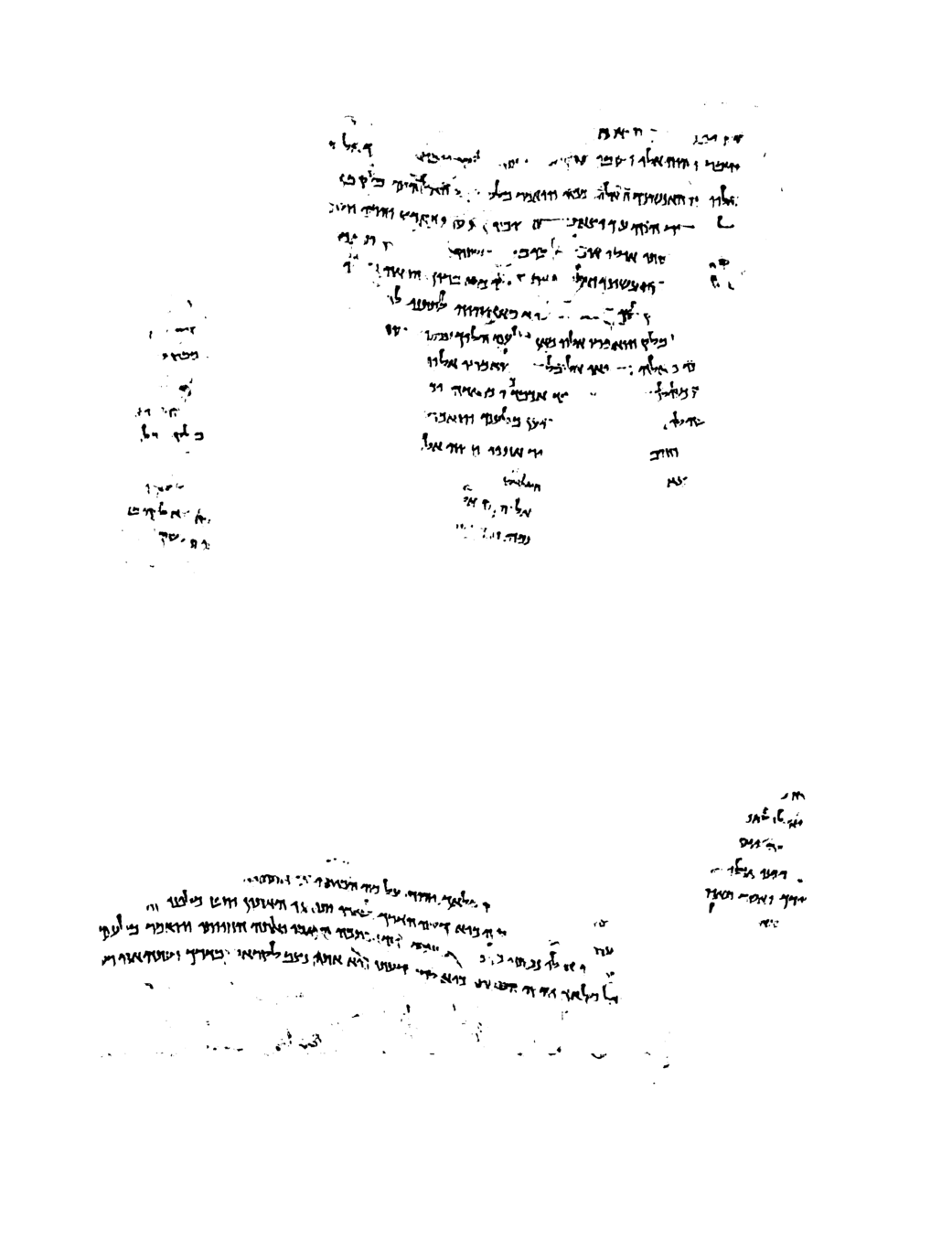}}%
		\caption{BiNet}
	\end{subfigure}
	\caption{Binarization results of two full plate images from the DSS collection using BiNet (trained on DIBCO images, then updated by transfer learning using sixteen manually labeled plate images).}
	\label{fig:fullPlateResults}
\end{figure}

\begin{figure}[h!]
	\centering
	\begin{subfigure}{.25\textwidth}
		\includegraphics[height=5.3cm]{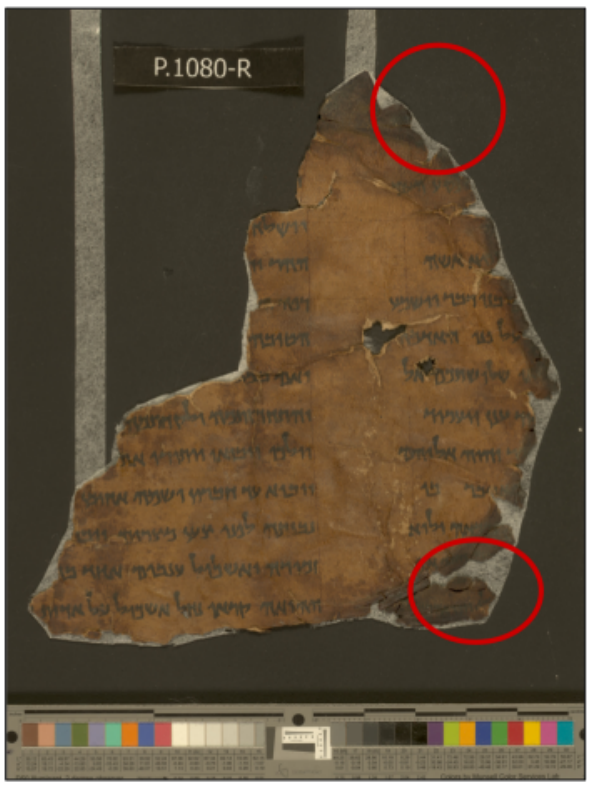}
		\caption{Original colour}
	\end{subfigure}\hfill
	\begin{subfigure}{.25\textwidth}
		\includegraphics[height=5.3cm]{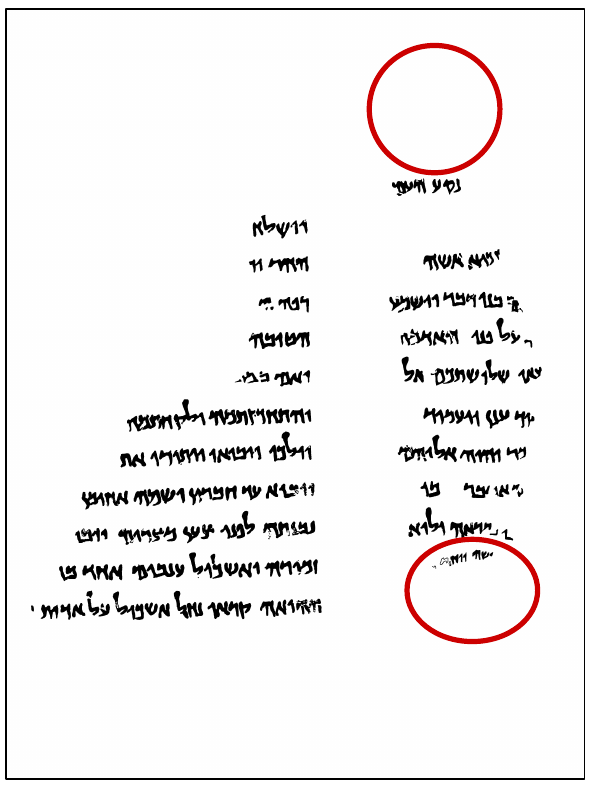}
		\caption{Manually labelled}
	\end{subfigure}\hfill
	\begin{subfigure}{.25\textwidth}	
		\includegraphics[height=5.3cm]{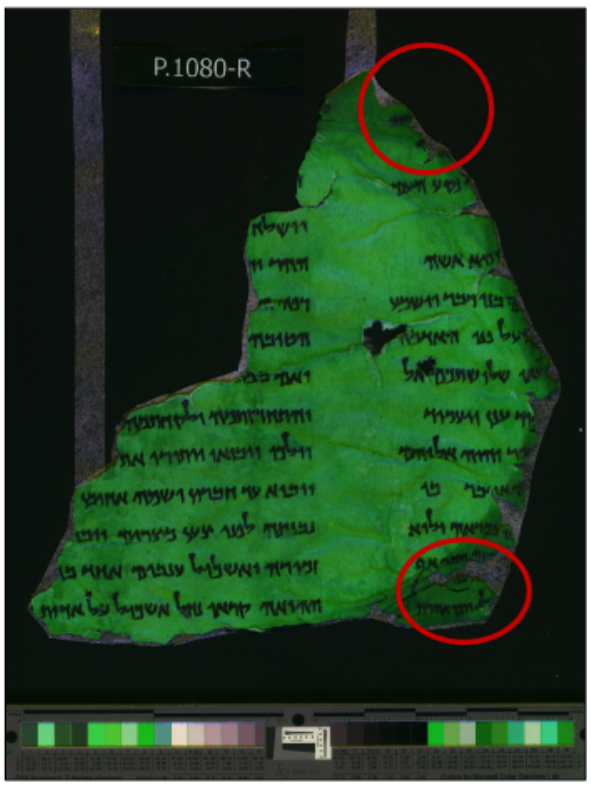}
		\caption{Original fused}
	\end{subfigure}\hfill
	\begin{subfigure}{.25\textwidth}	
		\includegraphics[height=5.3cm]{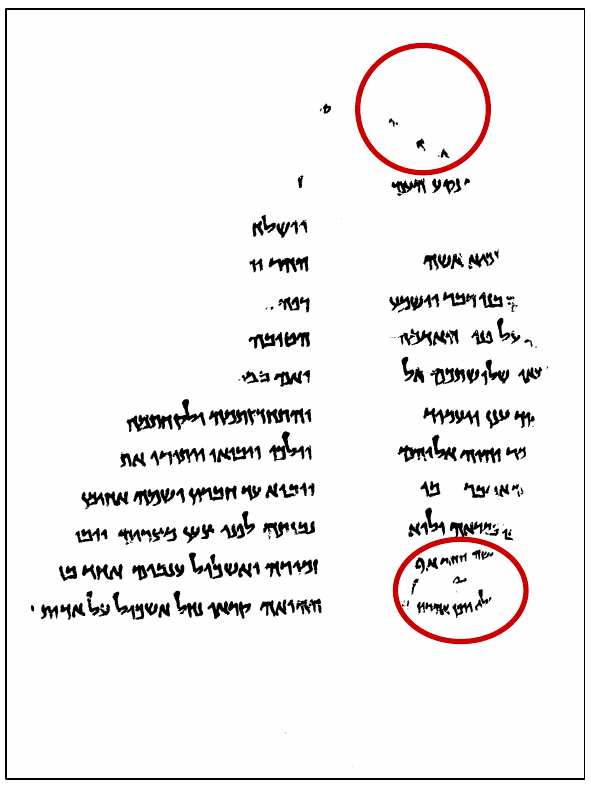}
		\caption{BiNet}	
	\end{subfigure}
	\caption{More characters (marked in {\color{red}red} circles) are extracted than the ground truth during the binarization process using BiNet on pseudo-color image (fused image) of fragment 6 from plate 1080. }
	\label{fig:resultfuse}
\end{figure}

The resulting images from different methods are presented in Figure \ref*{fig:resultfrag}. For a better qualitative analysis, a zoomed-in portion of the methods can be found in Figure \ref{fig:resultzoom}. BiNet is extremely good in binarizing the original content and labeling everything else as background. Inside the area of the original content, BiNet is remarkably similar to the ground truth (Figure \ref{fig:resultzoom}). An interesting finding in the results from the fused images can be seen in Figure \ref{fig:resultfuse}. During the labeling process of the images, the human expert only labeled the visible inks in the fragments. However, in the fused images, parts of ink might become visible, making it extractable during the binarization process. Thus a fused image binarization can reveal more characters than a visible color image (Figure  \ref{fig:resultfuse}). Please note that the phenomenon might reduce the quantitative performance as we are comparing the output with the ground truth. Nevertheless, this is extraordinary and useful in real applications. 

\begin{figure}[h!]
	\centering
	\includegraphics[height=5.4cm]{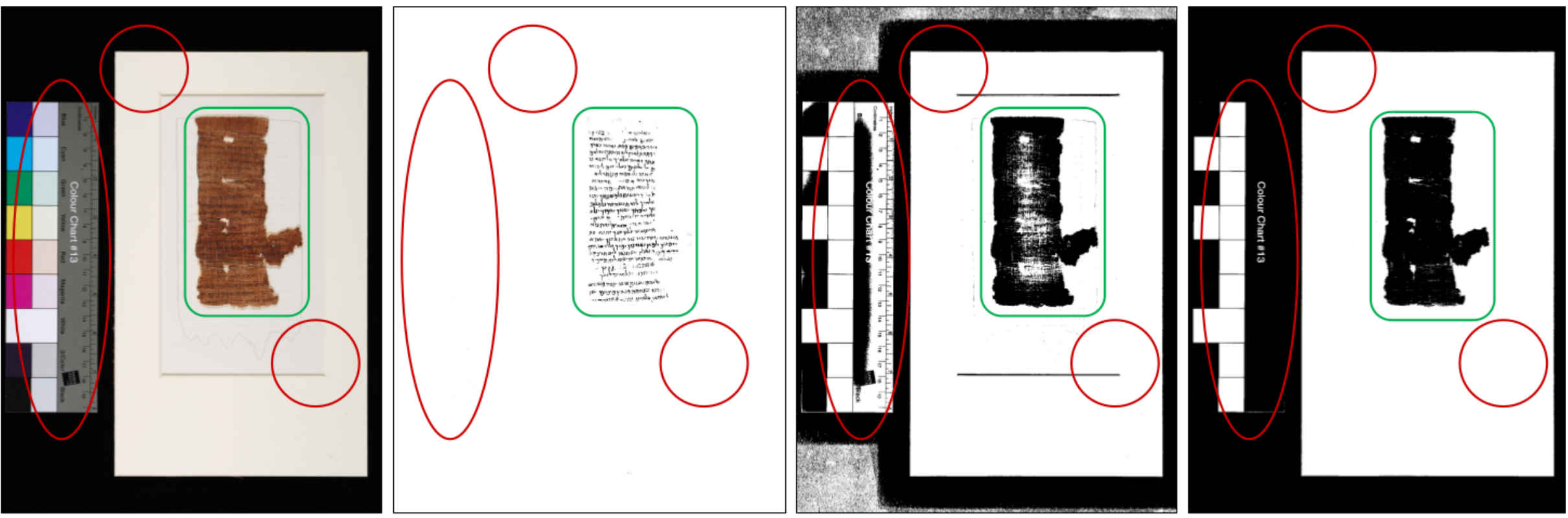}
	\caption{The binarization result from the proposed BiNet on a completely unseen (non-DSS) data (the Nash papyri) to illustrate the robustness of the system. From left to right: the original image, BiNet output, Sauvola, and Otsu. BiNet outputs only the original content from the papyrus fragment (marked in {\color{green}green} circle) and removes all the irrelevant components such as the picture frame and the machine-printed color-calibrators (marked in {\color{red}red} circles).}
	\label{fig:nash}
\end{figure}

Finally, to test the robustness of the model, we collected some additional materials from different collections to perform the binarization using BiNet. One such image is of the famous Nash Papyrus, which was kindly provided to us by Ben Outhwaite of the Genizah Research Unit at Cambridge University Library. We used the pretrained BiNet model from the DSS color-images to binarize the Nash papyrus and the result is illustrated in Figure \ref{fig:nash}. The binarization result shows yet another case where the BiNet can extract the original content by segmenting all the irrelevant materials as background pixels. We also performed tests on some grid images produced from several different historical manuscripts (non-DSS) from the Monk system. These results are presented in the Appendix \ref{appen:fig:monkgrid}.

%% file: tex/6-conclusions.tex
\section{Conclusions}
In this article, we proposed a complete framework, the BiNet, to efficiently binarize one of the most degraded historical manuscript collections, the Dead Sea Scrolls. The method can work with both the full-plate color images and the grayscale intensity images of the individual fragments. The network we used is a variant of U-Net and is capable of learning what is essential and what is irrelevant. Compared to traditional methods, the proposed BiNet can focus on and binarize the original written content of the document with remarkably high performance, which is crucial for getting as close as possible to the original writer to be able to perform writer identification and document dating. For a strictly handwriting based prediction, all non-ink information needs to be removed in these applications. One of the significant features of the network is the ability to segment everything except the original writing-contents into the background semantically. The binarization results of BiNet demonstrate the robustness and multi-purpose usability of the network on different degraded manuscripts of diverse document textures and layouts. 

In order to facilitate the training of the neural network, we utilized a simple and effective ground truth labeling technique. We proposed an image fusion technique to produce a pseudo-color image from multi-spectral image-bands. This technique improved the binarization results. In several cases, these improved results from fused images were able to extract more of the original contents than the ground truth itself. Though this phenomenon might lead to lower performance on the quantitative analysis, this additional extraction is much more desirable in the real-world applications. Similar to BiNet, the fusion technique can also be used in different collections if multi-spectral images are available. Though our framework was initially designed for the DSS images, we have performed experiments with (H-)DIBCO datasets as well. The BiNet delivered excellent results for these datasets with high performance-measures, similar to the best-performing methods. This success shows the prospect of using BiNet as a one-off tool for manuscript binarization. 

In this article, we have worked on the binarization part only. Additionally, both our ground-truth labeling and binarization results were restrictive. We were careful enough to avoid any automatic filling or smoothing of the characters. This conservative approach made the output of BiNet crisp and accurate to the ink of each character. As most of the historical manuscripts show extreme degradation even on the character level, a reconstruction-based binarization needs to be explored in the future. More research can be done on the image fusion technique itself by fusing more than three channels. In cases of a larger image, we divided the images into smaller ones ($256\times256$). During the binarization process, the border areas sometimes lack in obtaining perfect binarization. This problem can be improved by further work in the area and might be solved by using overlapping patches of images. We used a variant of U-Net due to the low amount of training data with high complexity. If more training data becomes available, a more in-depth network similar to ResNet \cite{he2016deep} or even a dense network similar to DenseNet \cite{huang2017densely} can be explored in the future. Though we propose a complete binarization framework, for now, additional research can always be performed for further improvement of the technique.

%% file: tex/acknowledgements.tex
\section{Acknowledgements} \label{sec:acknowledge}
The authors would like to thank Mladen Popovi\'{c} (Principal investigator of the ERC project, Faculty of Theology and Religious Studies, University of Groningen),   Eibert Tigchelaar (Research Professor, Faculty of Theology and Religious Studies, KU Leuven) for their valuable inputs and time in labeling and creating the ground truths of the DSS images. The authors would also like to thank Drew Longacre (Post-doctoral fellow, ERC project), Gemma Hayes (Ph.D. candidate, ERC project), Ayhan Aksu (Ph.D. candidate, NWO/FWO project), Marwin Van Dijk (Research assistant, ERC project), and Christine van der Veer (Research assistant, ERC project) for their time and effort in labeling a number of the ground-truth images. 

This work has been supported by an ERC Starting Grant of the European Research Council (EU Horizon 2020): \textit{The Hands that Wrote the Bible: Digital Palaeography and Scribal Culture of the DSS} (HandsandBible \# 640497).

%% file: tex/7-appendix.tex
\begin{figure}[h!]
	\centering
	\begin{subfigure}{.45\textwidth}
		\fbox{\includegraphics[width=.8\linewidth]{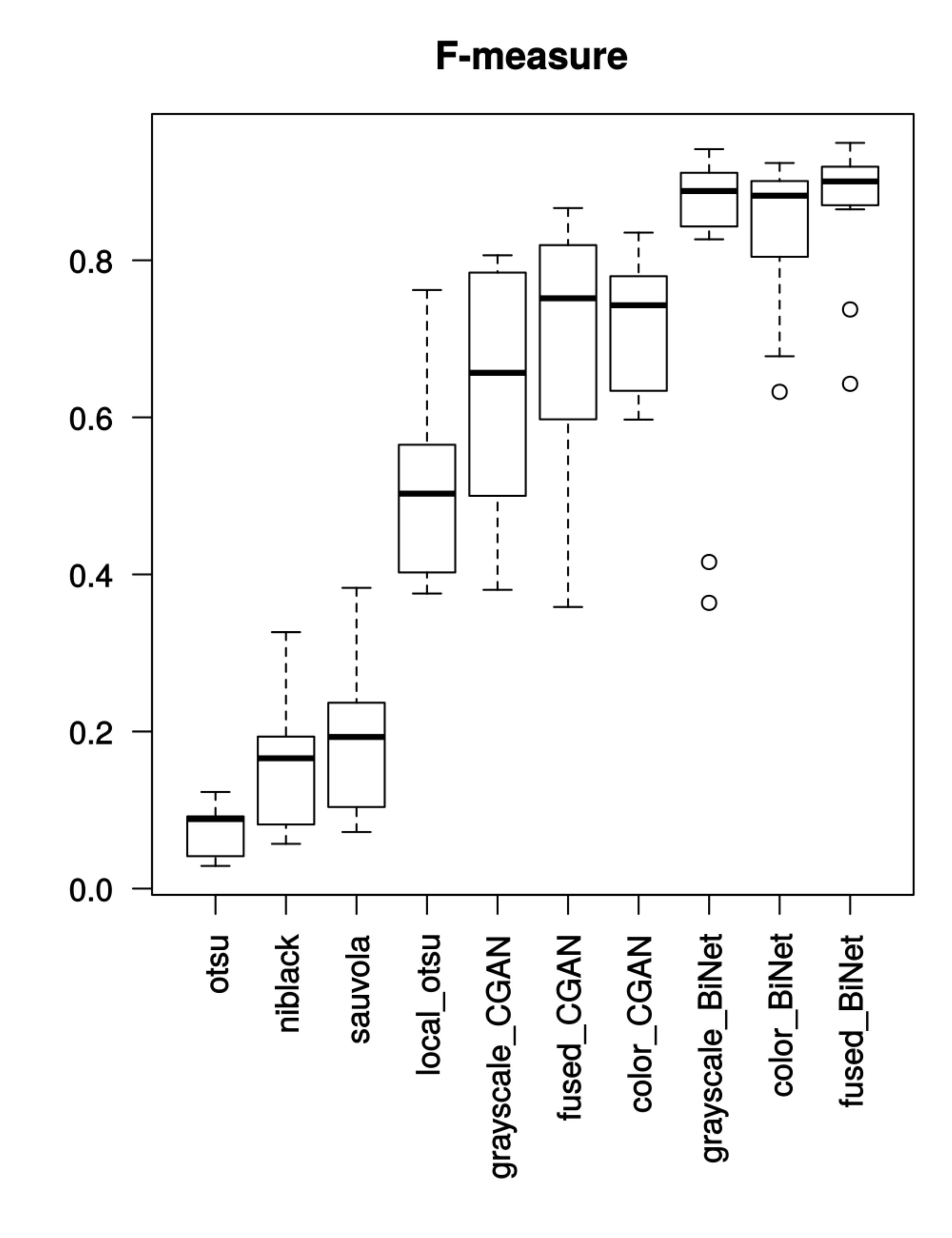} }
		\subcaption{F-measure}
	\end{subfigure} \hfill
	\begin{subfigure}{.45\textwidth}
		\fbox{\includegraphics[width=.8\linewidth]{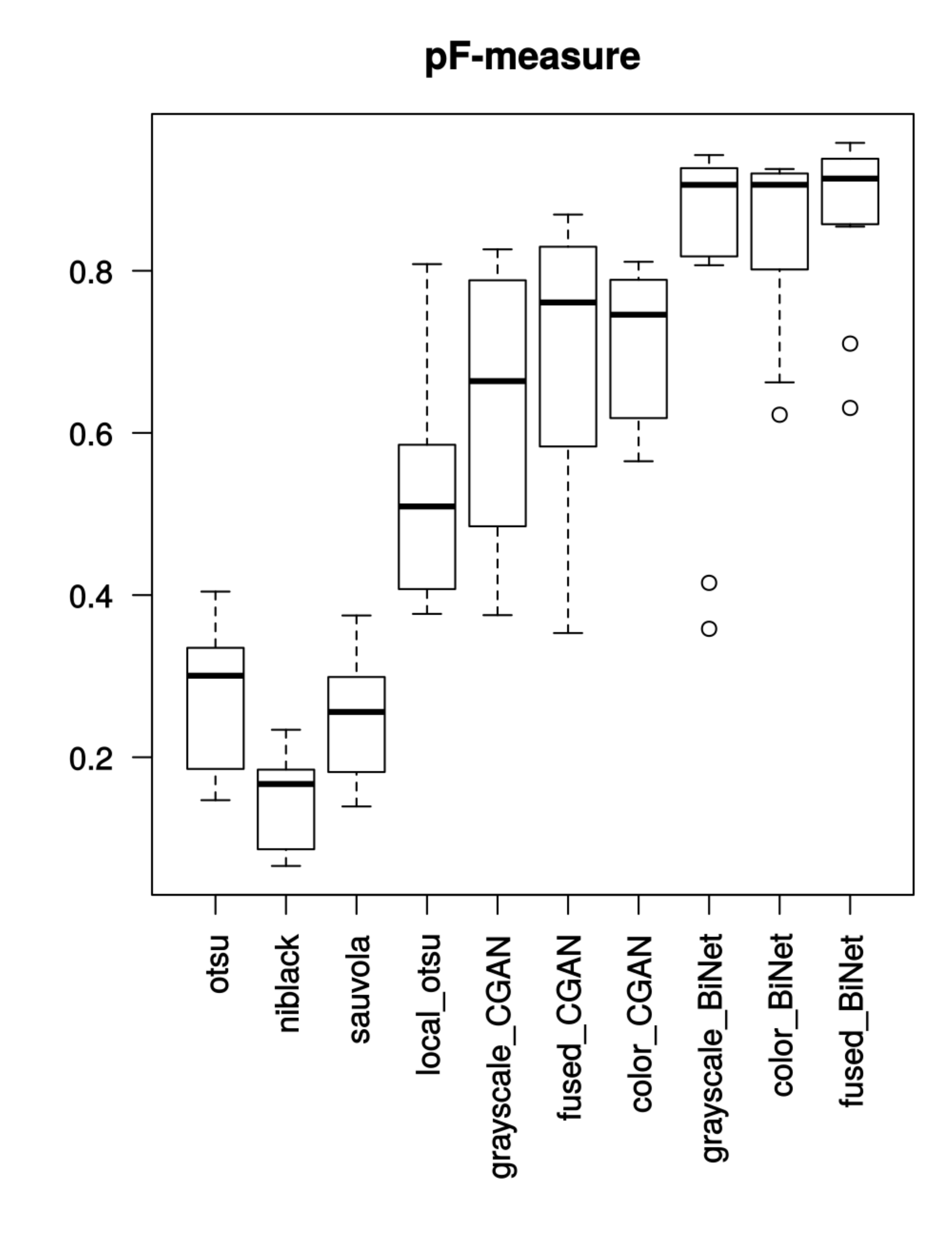} }
		\subcaption{pF-measure}
	\end{subfigure}
	\begin{subfigure}{.45\textwidth}
		\fbox{\includegraphics[width=.8\linewidth]{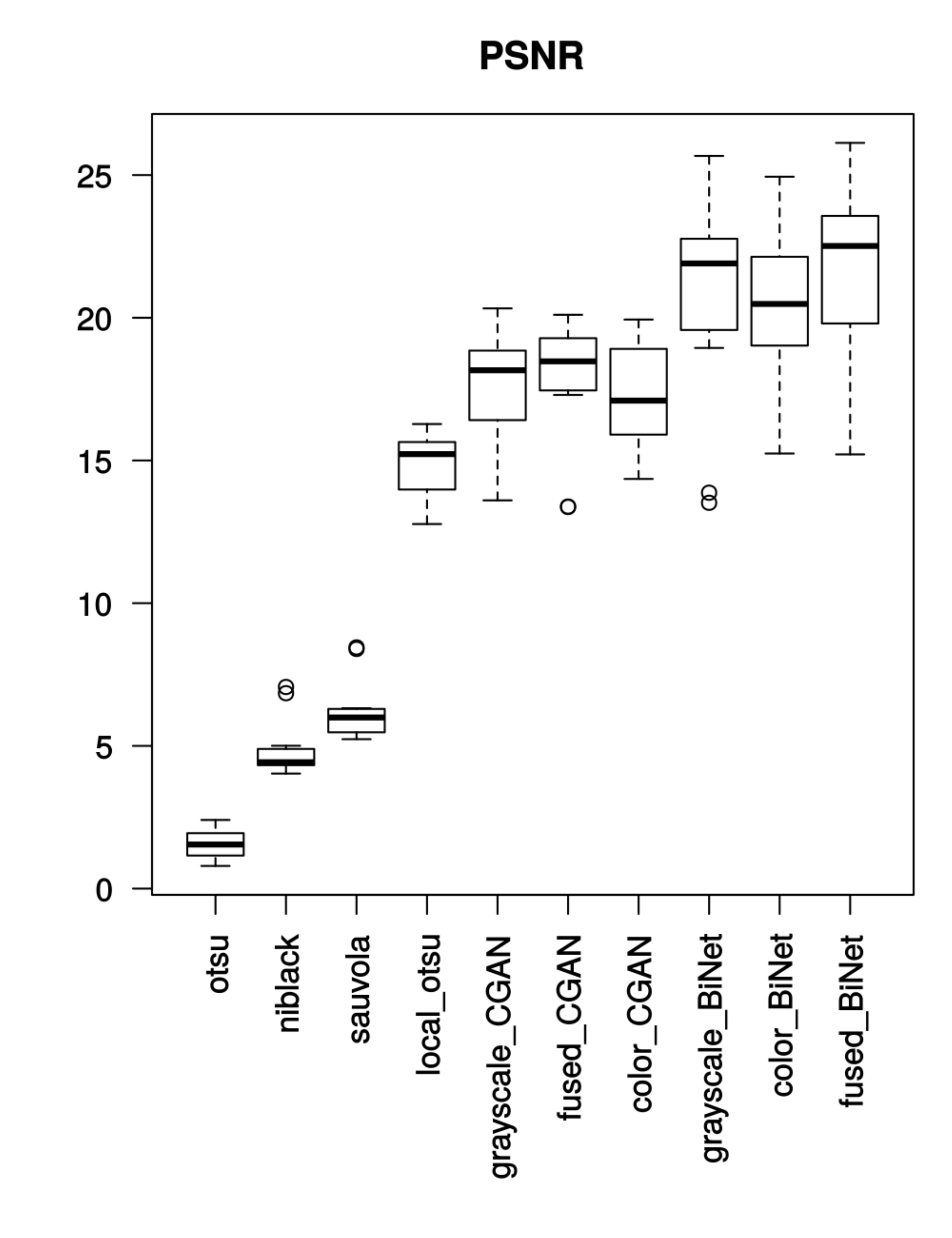} }
		\subcaption{PSNR}	
	\end{subfigure} \hfill
	\begin{subfigure}{.45\textwidth}	
		\fbox{\includegraphics[width=.8\linewidth]{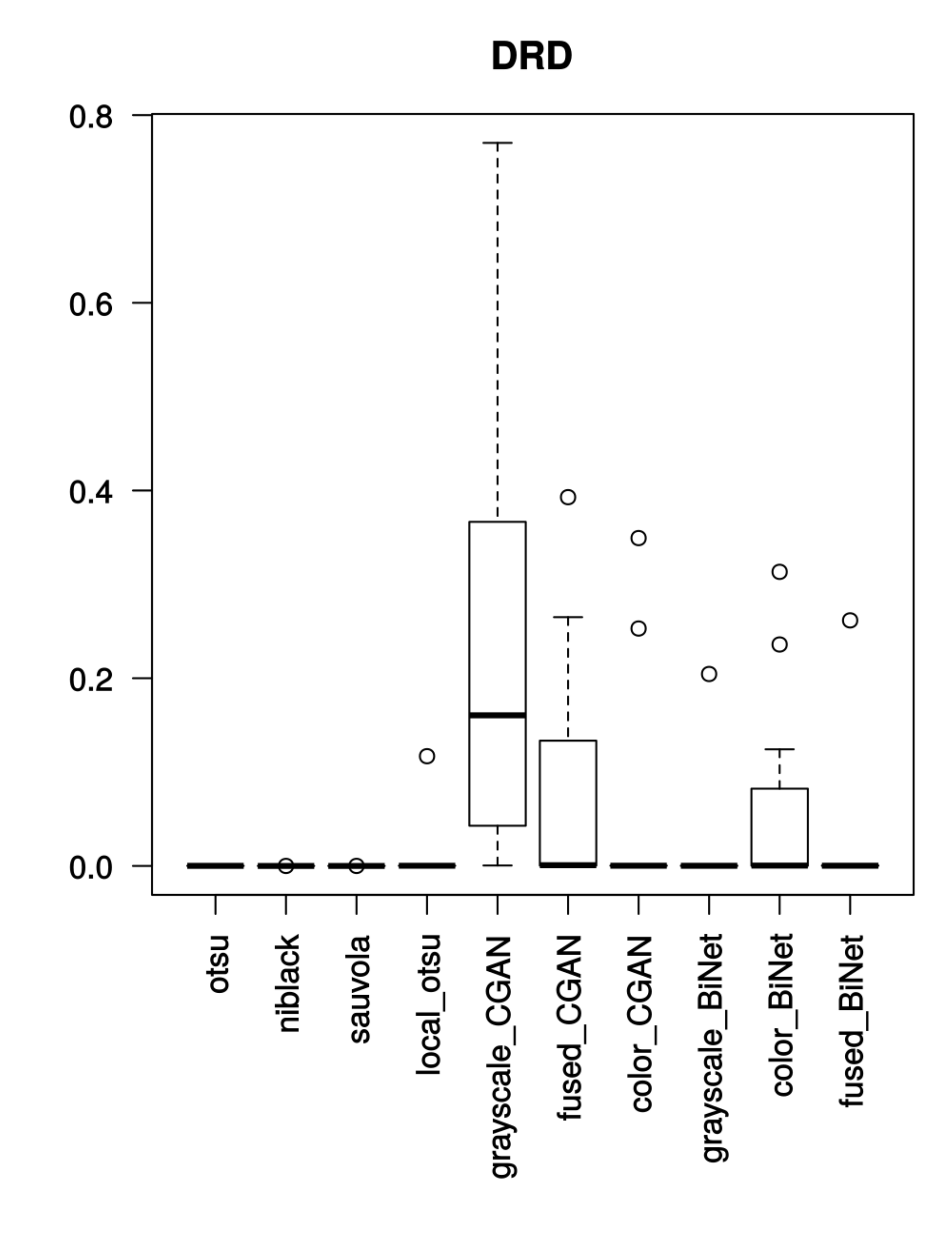} }
		\subcaption{DRD}
	\end{subfigure}
	\caption{Box-plots showing the distribution of test data (DSS fragment-images) for the four different evaluation metrics.}
	\label{appen:fig:boxplot}
\end{figure}
\begin{figure}[H]
	\centering
	\begin{subfigure}{.3\textwidth}
		\fbox{\includegraphics[height=3.5cm]{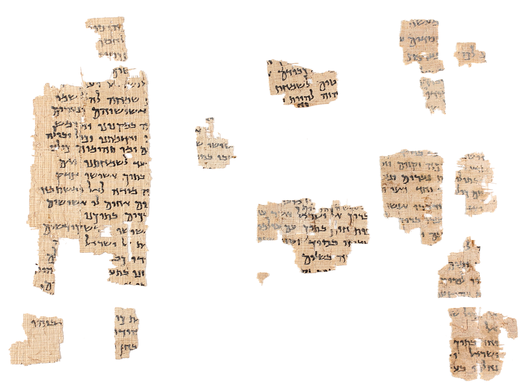} }
		\subcaption{Plate 1}
	\end{subfigure} \hfill
	\begin{subfigure}{.3\textwidth}
		\fbox{\includegraphics[height=3.5cm]{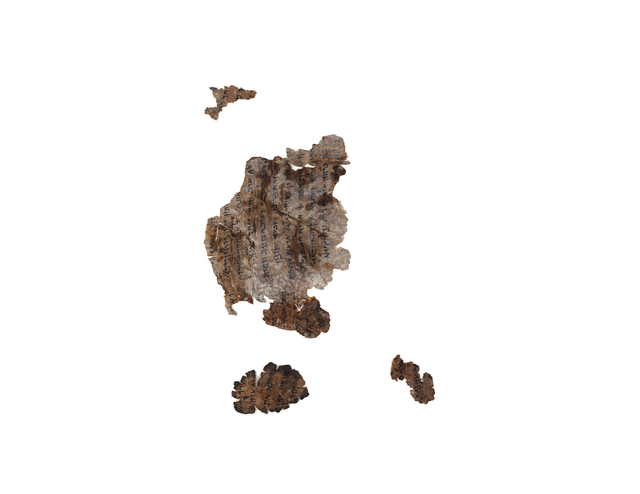} }
		\subcaption{Plate 386}
	\end{subfigure} \hfill
	\begin{subfigure}{.3\textwidth}
		\fbox{\includegraphics[height=3.5cm]{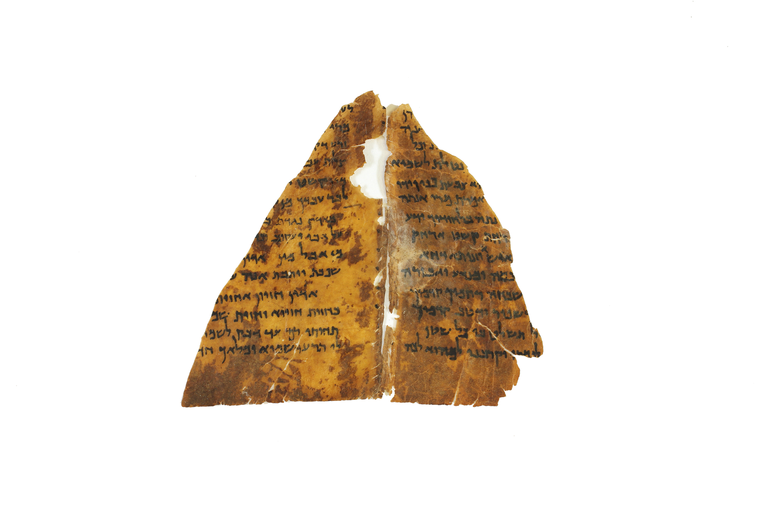} }
		\subcaption{Plate 997}	
	\end{subfigure} 
	\begin{subfigure}{.3\textwidth}
		\fbox{\includegraphics[height=3.5cm]{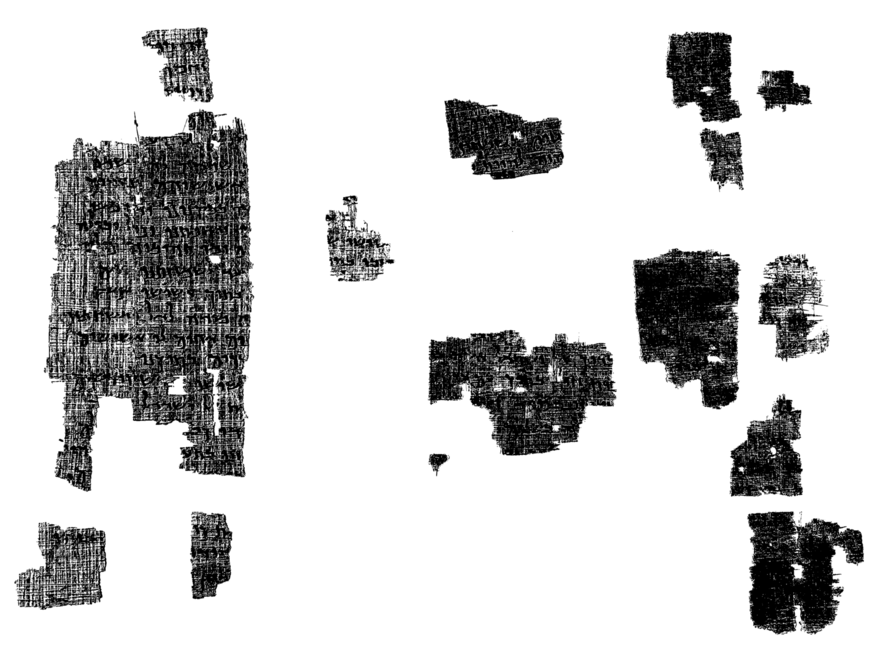} }
		\subcaption{Otsu of Plate 1}
	\end{subfigure} \hfill
	\begin{subfigure}{.3\textwidth}
		\fbox{\includegraphics[height=3.5cm]{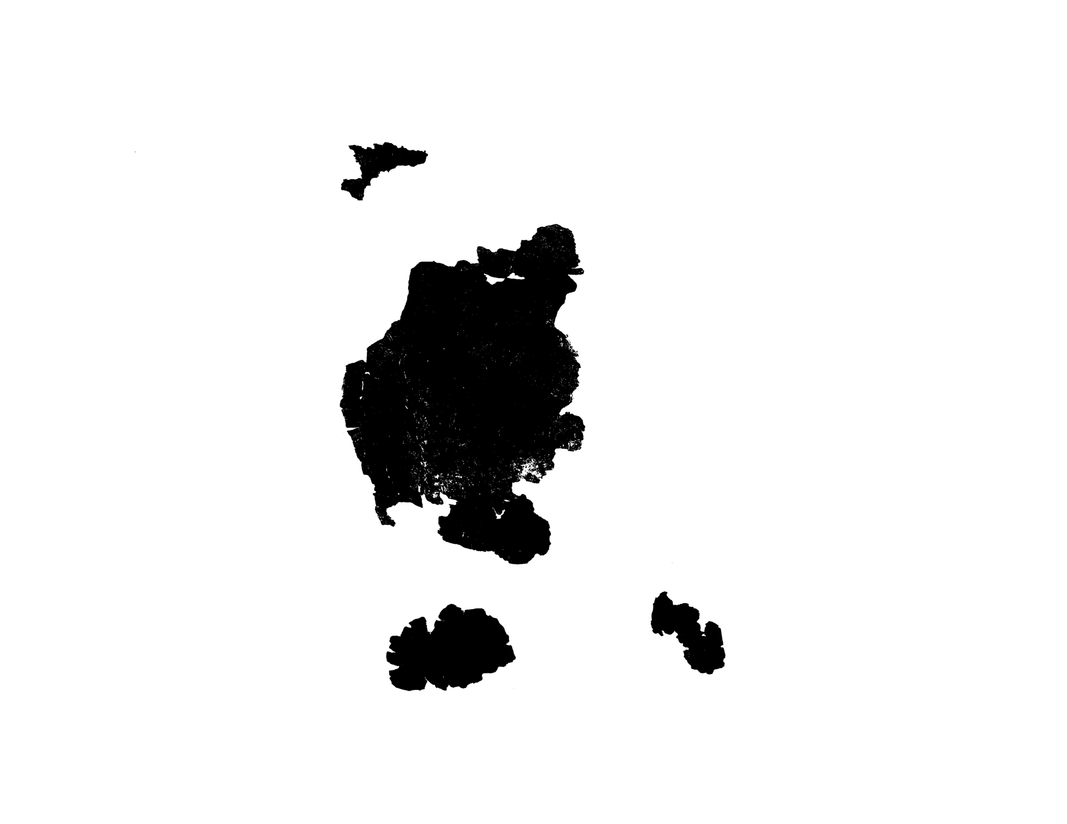} }
		\subcaption{Otsu of Plate 386}
	\end{subfigure} \hfill
	\begin{subfigure}{.3\textwidth}
		\fbox{\includegraphics[height=3.5cm]{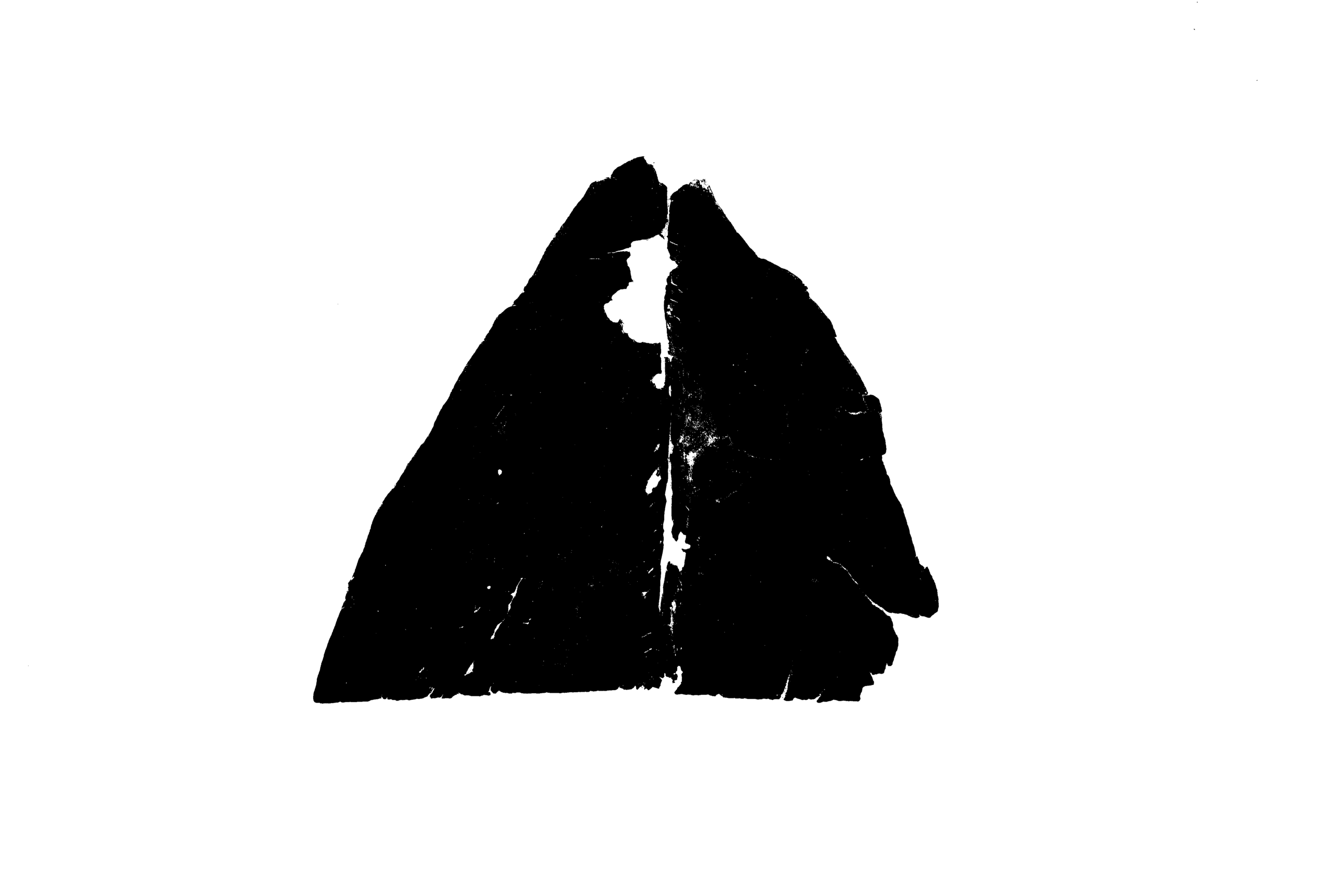} }
		\subcaption{Otsu of Plate 997}	
	\end{subfigure} 
	\begin{subfigure}{.3\textwidth}
		\fbox{\includegraphics[height=3.5cm]{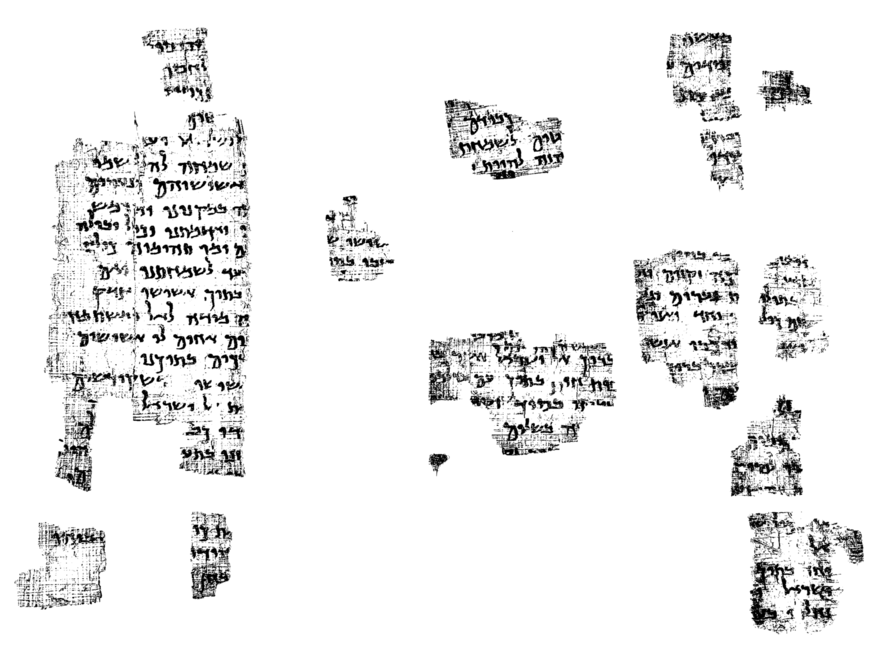} }
		\subcaption{Sauvola of Plate 1}
	\end{subfigure} \hfill
	\begin{subfigure}{.3\textwidth}
		\fbox{\includegraphics[height=3.5cm]{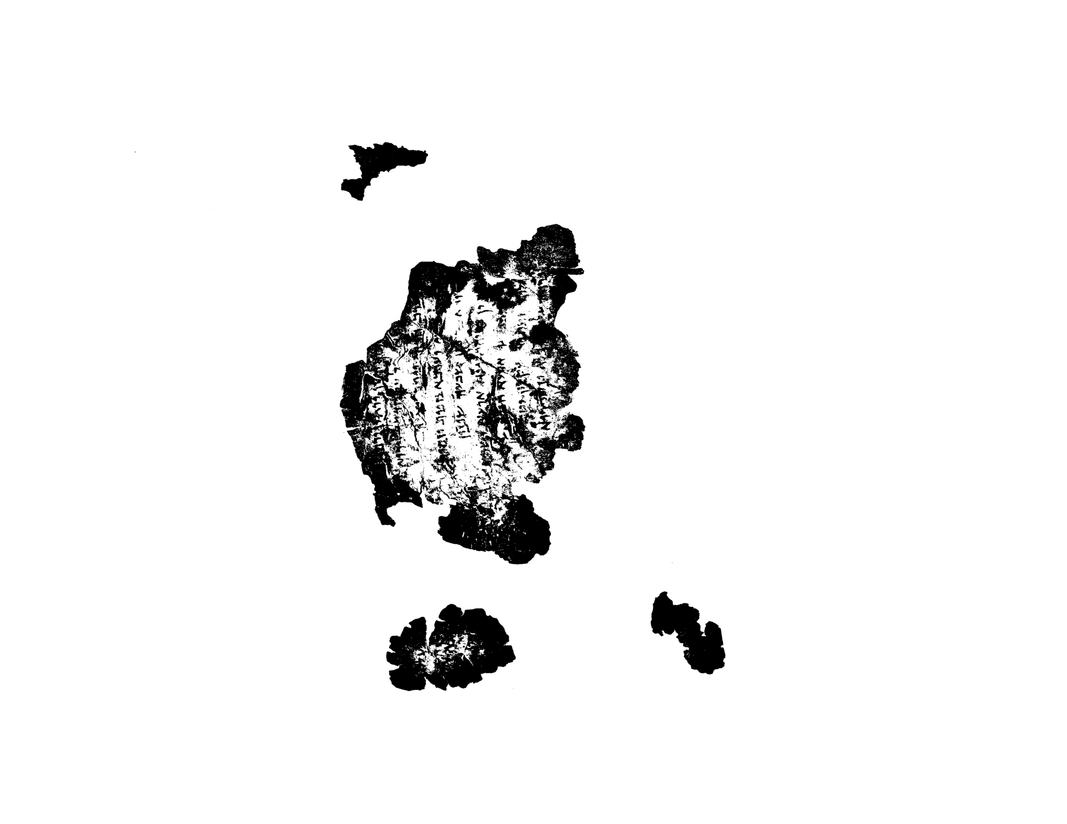} }
		\subcaption{Sauvola of Plate 386}
	\end{subfigure} \hfill
	\begin{subfigure}{.3\textwidth}
		\fbox{\includegraphics[height=3.5cm]{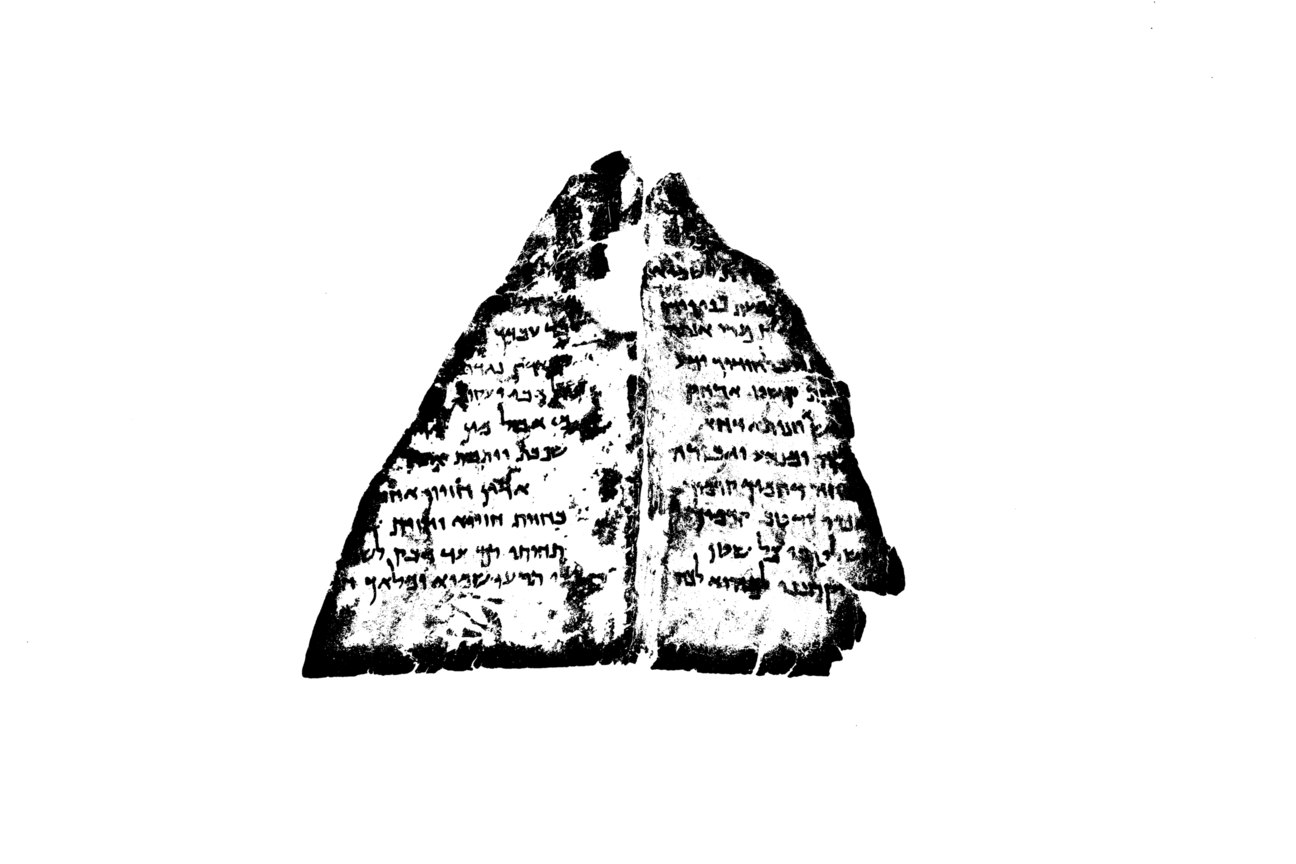} }
		\subcaption{Sauvola of Plate 997}	
	\end{subfigure} 
	\begin{subfigure}{.3\textwidth}
		\fbox{\includegraphics[height=3.5cm]{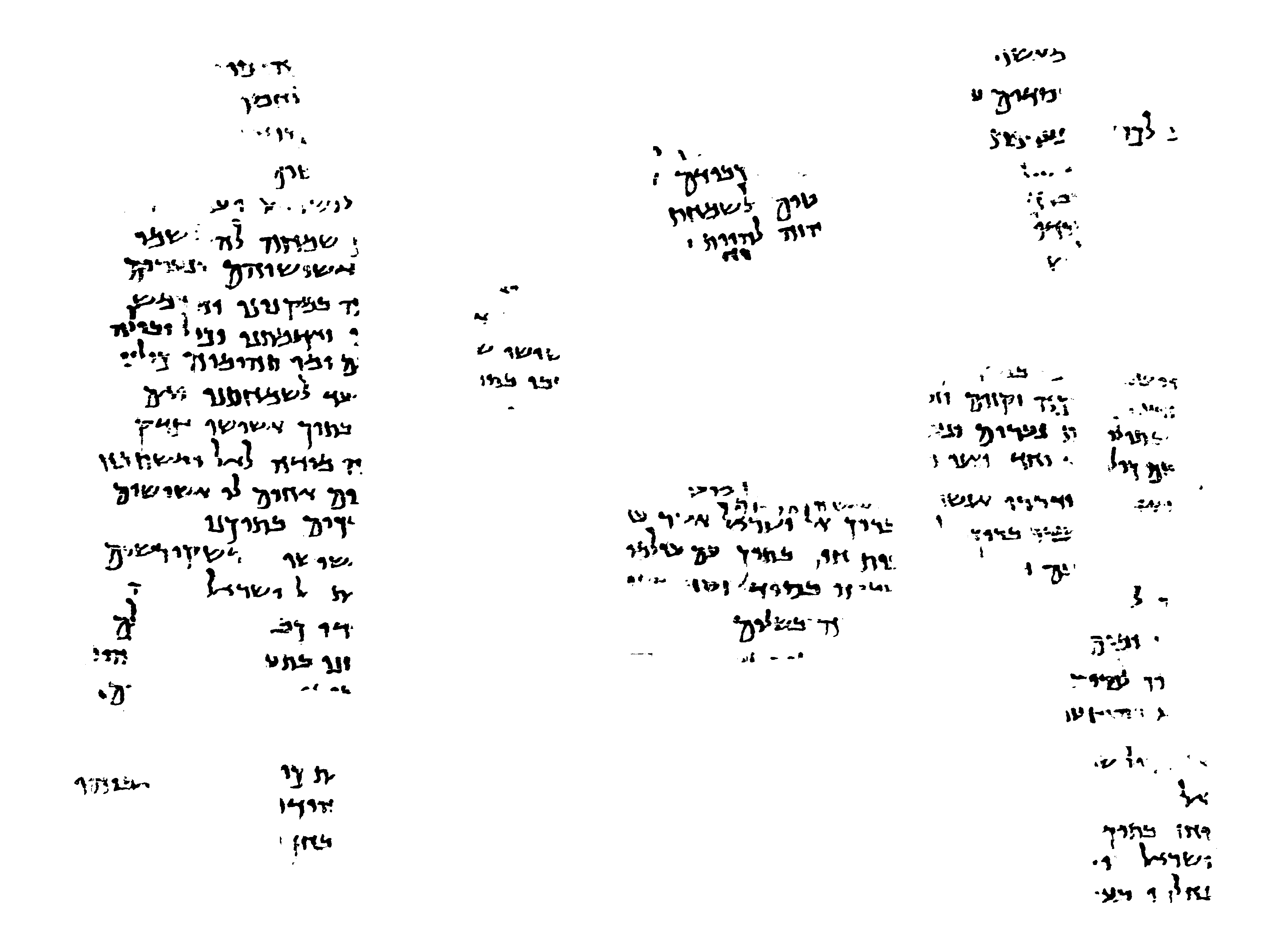} }
		\subcaption{BiNet on Plate 1}
	\end{subfigure} \hfill
	\begin{subfigure}{.3\textwidth}
		\fbox{\includegraphics[height=3.5cm]{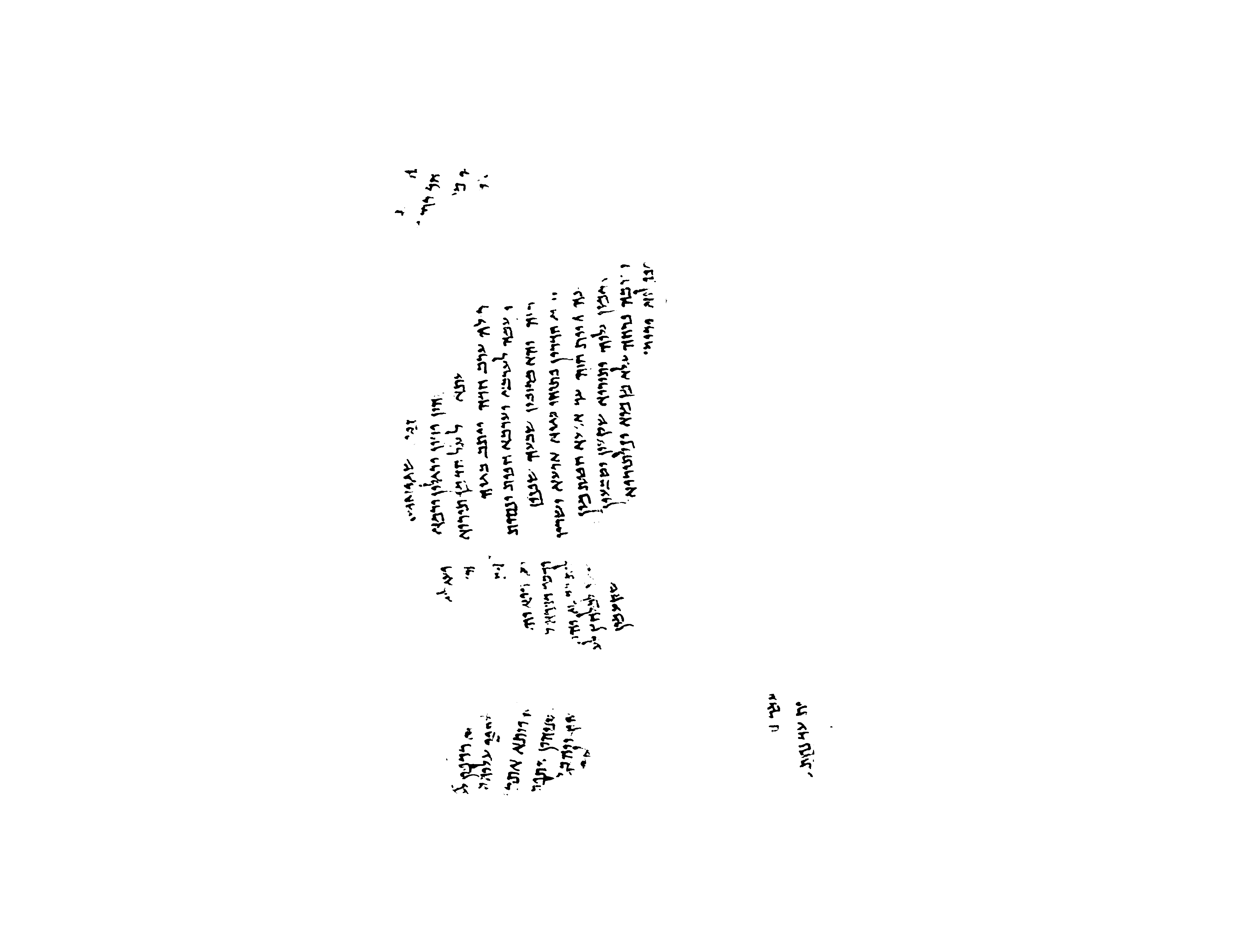} }
		\subcaption{BiNet on Plate 386}
	\end{subfigure} \hfill
	\begin{subfigure}{.3\textwidth}
		\fbox{\includegraphics[height=3.5cm]{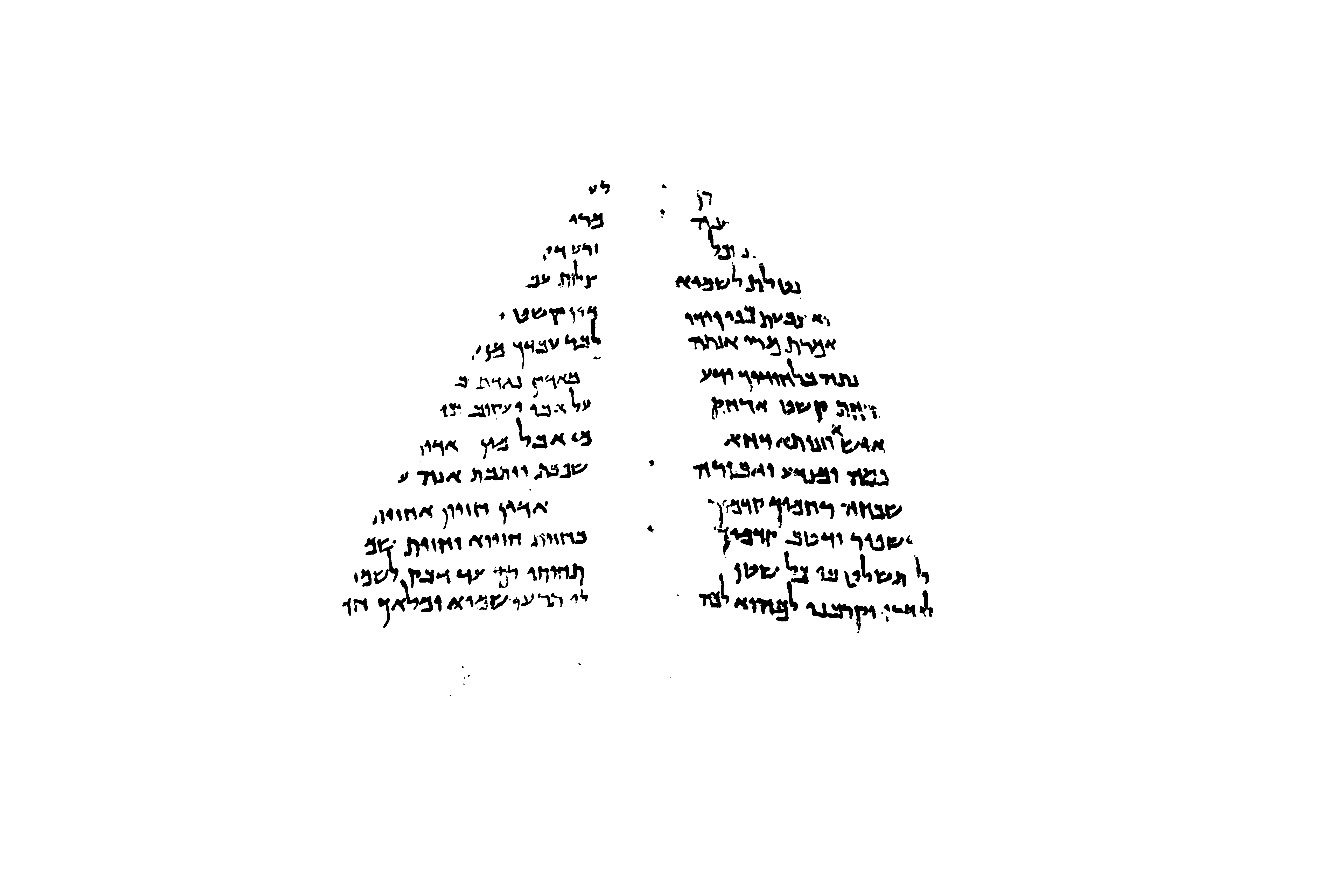} }
		\subcaption{BiNet on Plate 997}	
	\end{subfigure} 
	\begin{subfigure}{.3\textwidth}
		\fbox{\includegraphics[height=3.5cm]{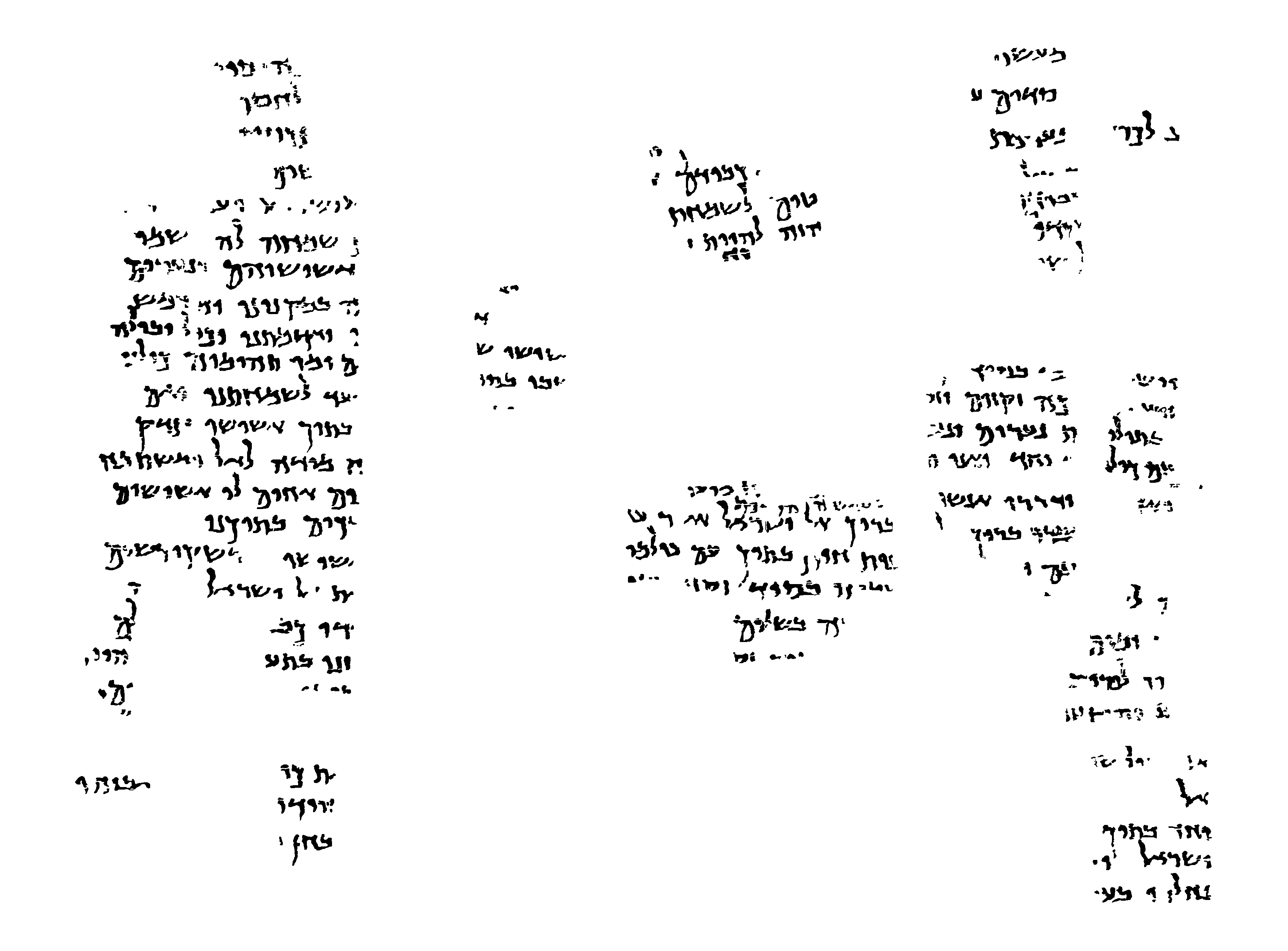} }
		\subcaption{Ground-truth of Plate 1}
	\end{subfigure} \hfill
	\begin{subfigure}{.3\textwidth}
		\fbox{\includegraphics[height=3.5cm]{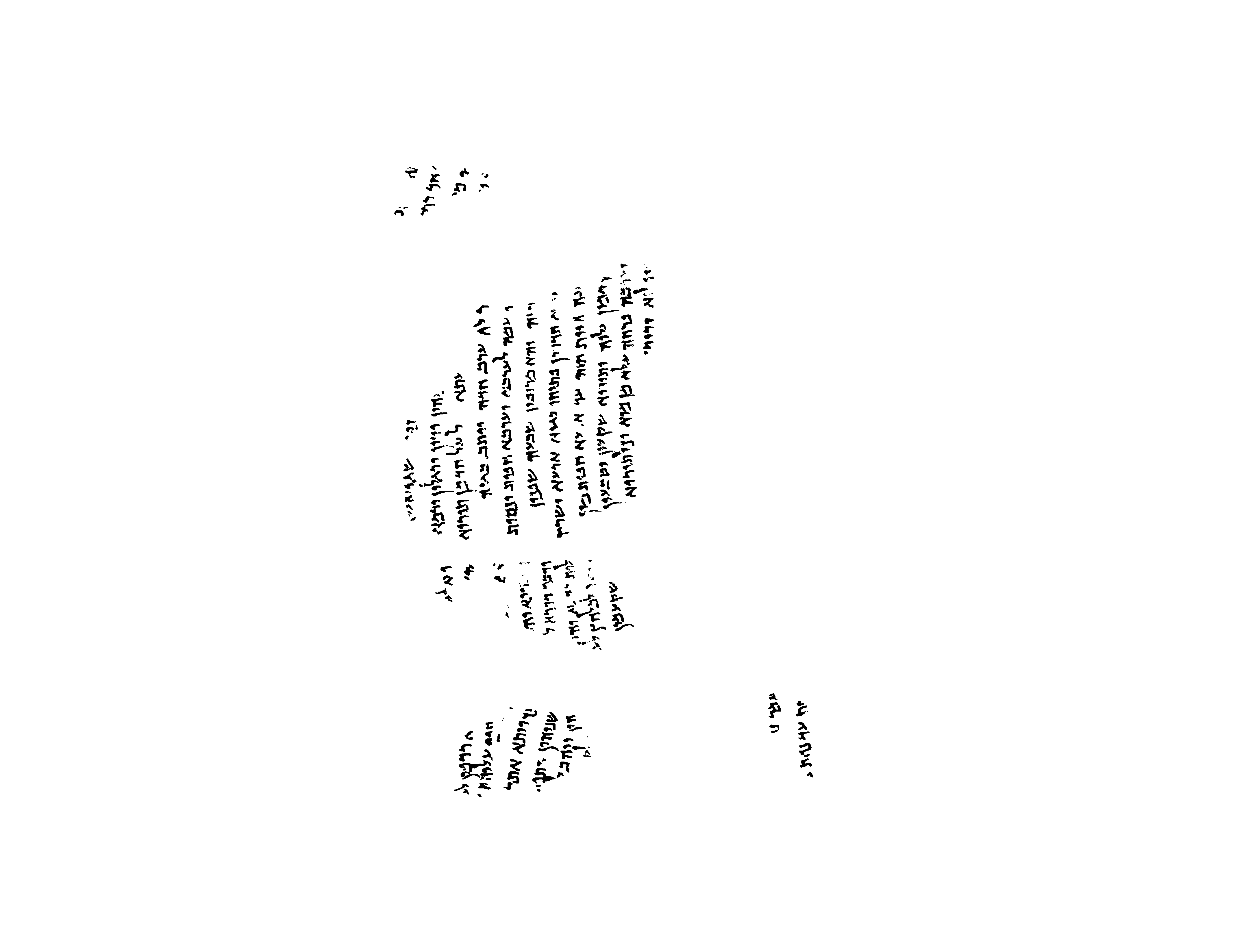} }
		\subcaption{Ground-truth of Plate 386}
	\end{subfigure} \hfill
	\begin{subfigure}{.3\textwidth}
		\fbox{\includegraphics[height=3.5cm]{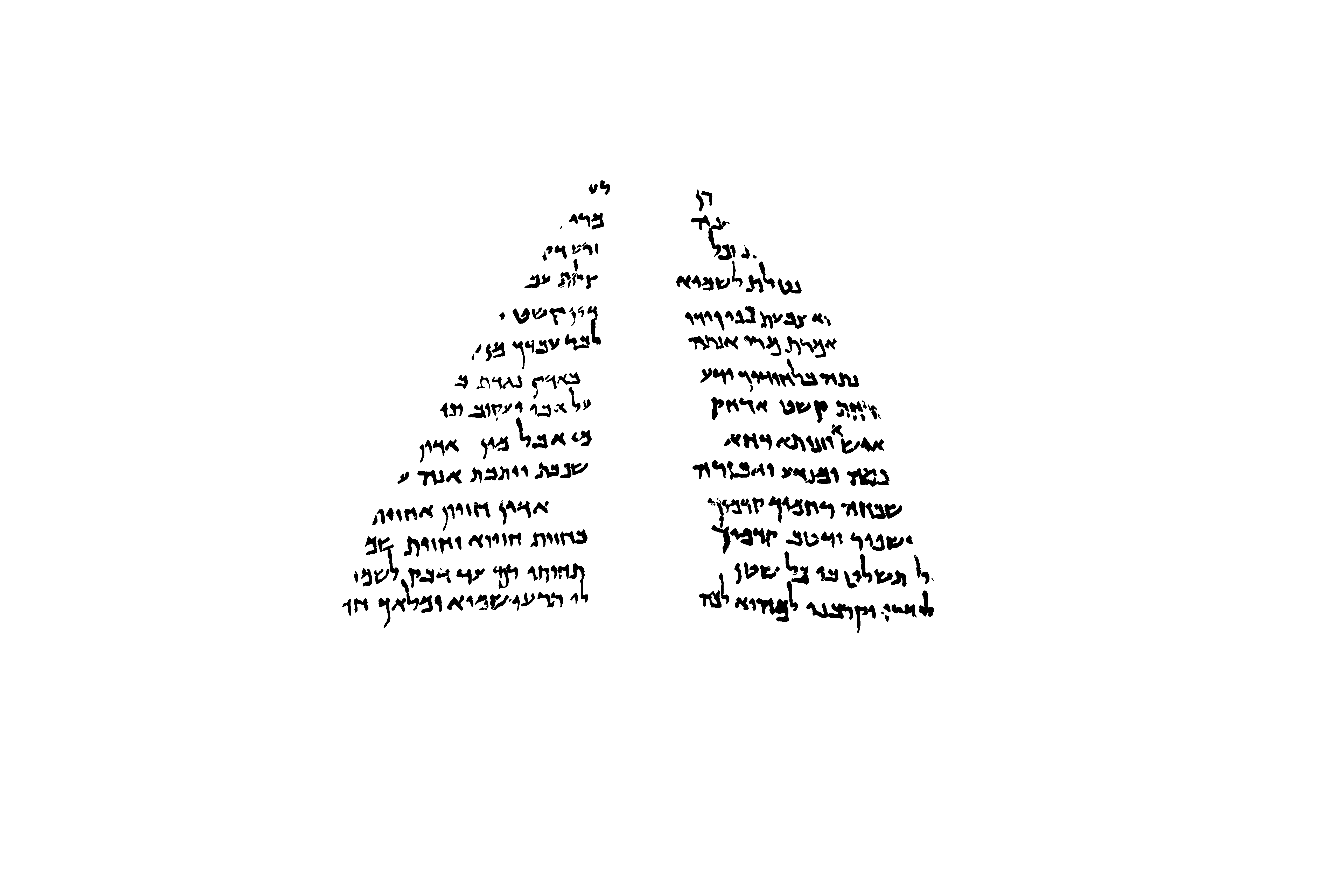} }
		\subcaption{Ground-truth of Plate 997}	
	\end{subfigure} 
	\caption{Binarization results of three test images (DSS full-plate images) using BiNet (trained on DIBCO images, then updated by transfer learning using sixteen manually labeled plate images).}
	\label{appen:fig:plateimages}
\end{figure}
\begin{figure}[h!]
	\centering
	\begin{subfigure}{.23\textwidth}
		\fbox{\includegraphics[width=.99\linewidth]{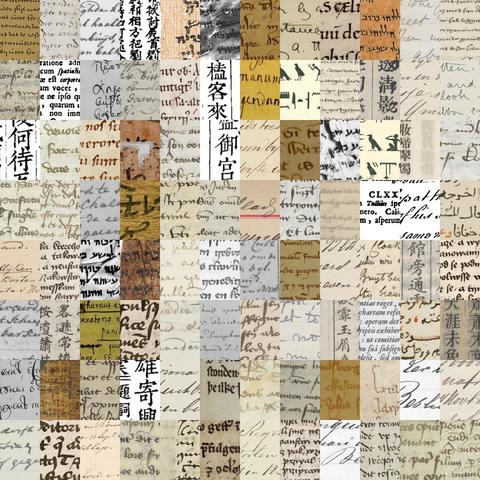}}
		\subcaption{Original color}
	\end{subfigure} \hfill
	\begin{subfigure}{.23\textwidth}
		\fbox{\includegraphics[width=.99\linewidth]{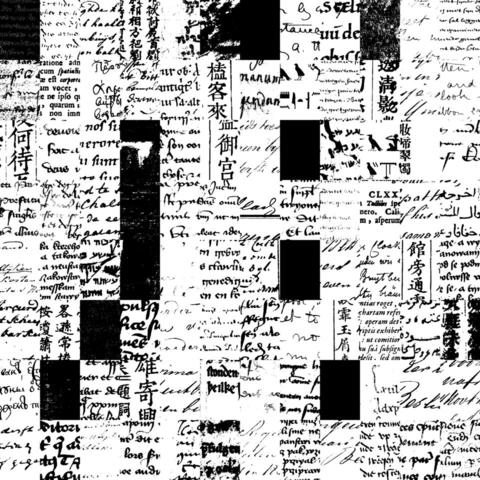}}
		\subcaption{Otsu}
	\end{subfigure} \hfill
	\begin{subfigure}{.23\textwidth}
		\fbox{\includegraphics[width=.99\linewidth]{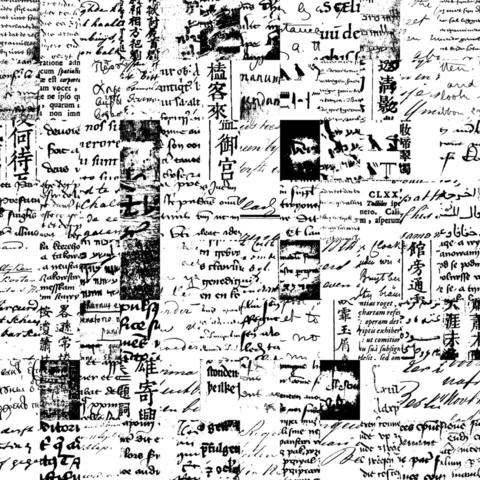}}
		\subcaption{Sauvola}	
	\end{subfigure} \hfill 
	\begin{subfigure}{.23\textwidth}	
		\fbox{\includegraphics[width=.99\linewidth]{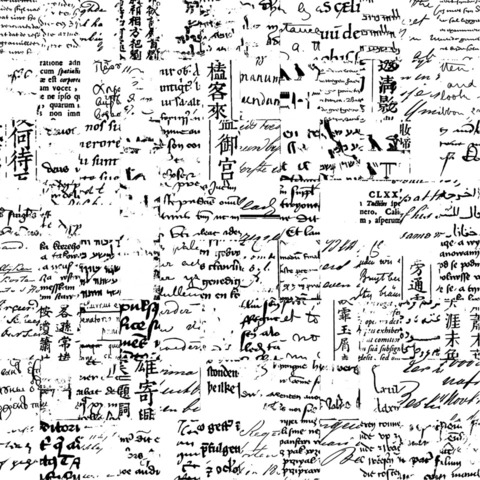}}
		\subcaption{BiNet}
	\end{subfigure}
	\begin{subfigure}{.23\textwidth}
		\fbox{\includegraphics[width=.99\linewidth]{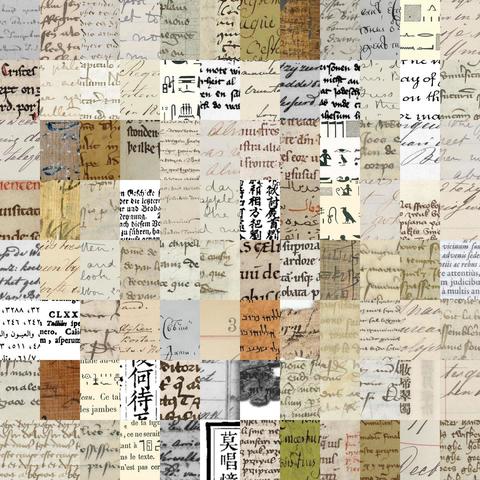}}
		\subcaption{Original color}
	\end{subfigure} \hfill
	\begin{subfigure}{.23\textwidth}
		\fbox{\includegraphics[width=.99\linewidth]{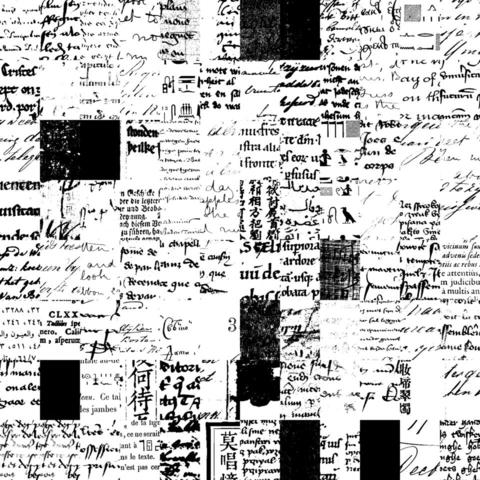}}
		\subcaption{Otsu}
	\end{subfigure} \hfill
	\begin{subfigure}{.23\textwidth}
		\fbox{\includegraphics[width=.99\linewidth]{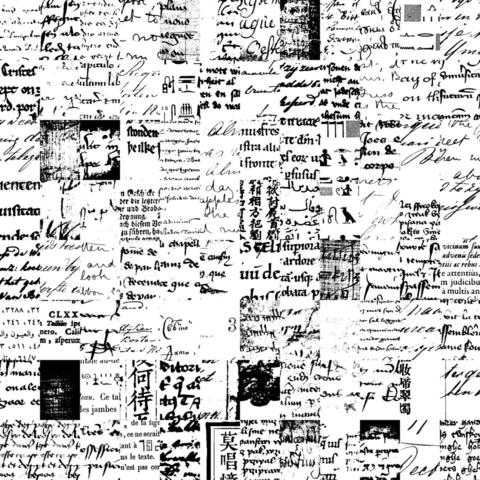}}
		\subcaption{Sauvola}	
	\end{subfigure} \hfill 
	\begin{subfigure}{.23\textwidth}	
		\fbox{\includegraphics[width=.99\linewidth]{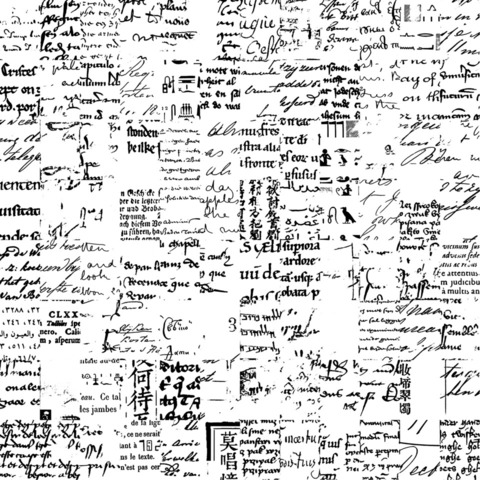}}
		\subcaption{BiNet}
	\end{subfigure}
	\begin{subfigure}{.23\textwidth}
		\fbox{\includegraphics[width=.99\linewidth]{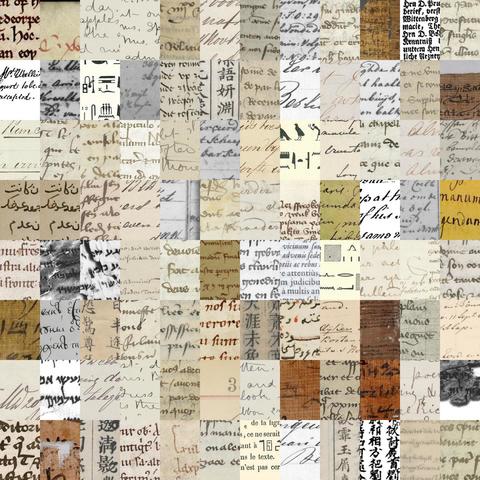}}
		\subcaption{Original color}
	\end{subfigure} \hfill
	\begin{subfigure}{.23\textwidth}
		\fbox{\includegraphics[width=.99\linewidth]{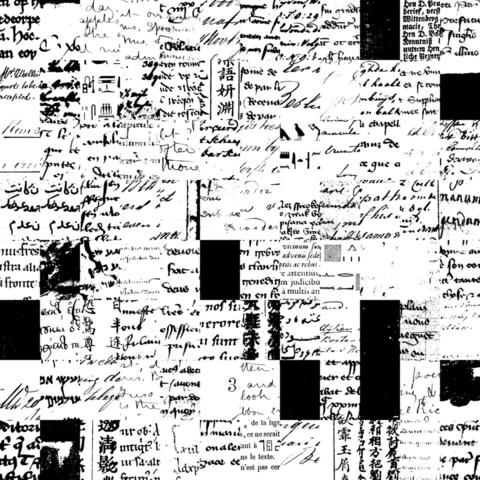}}
		\subcaption{Otsu}
	\end{subfigure} \hfill
	\begin{subfigure}{.23\textwidth}
		\fbox{\includegraphics[width=.99\linewidth]{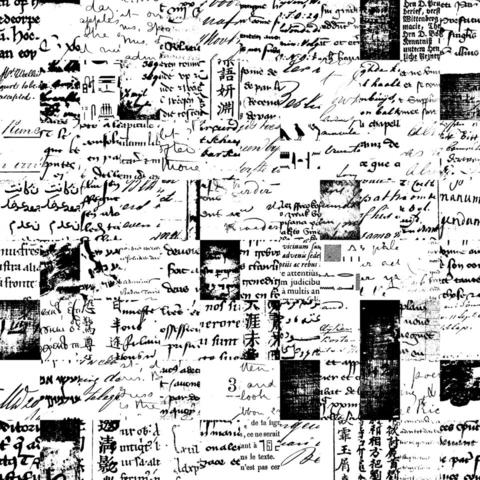}}
		\subcaption{Sauvola}	
	\end{subfigure} \hfill 
	\begin{subfigure}{.23\textwidth}	
		\fbox{\includegraphics[width=.99\linewidth]{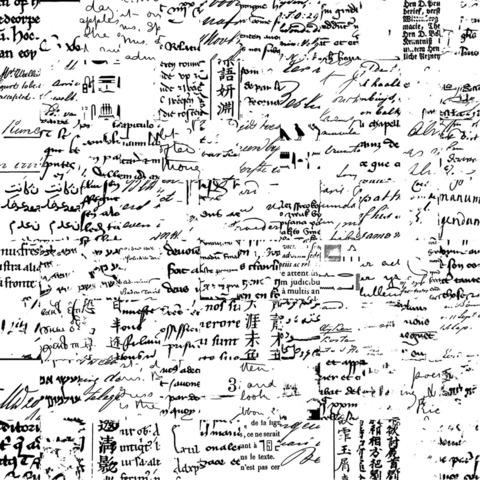}}
		\subcaption{BiNet}
	\end{subfigure}
	\begin{subfigure}{.23\textwidth}
		\fbox{\includegraphics[width=.99\linewidth]{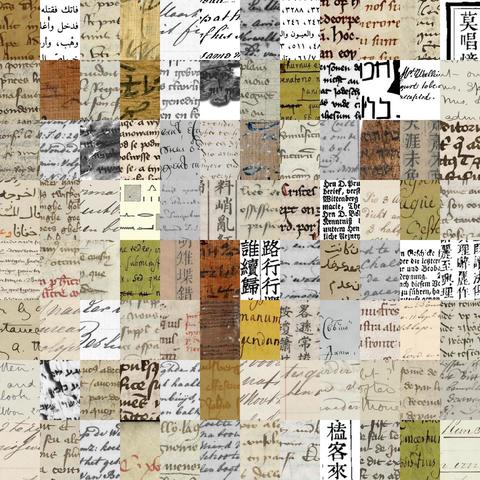}}
		\subcaption{Original color}
	\end{subfigure} \hfill
	\begin{subfigure}{.23\textwidth}
		\fbox{\includegraphics[width=.99\linewidth]{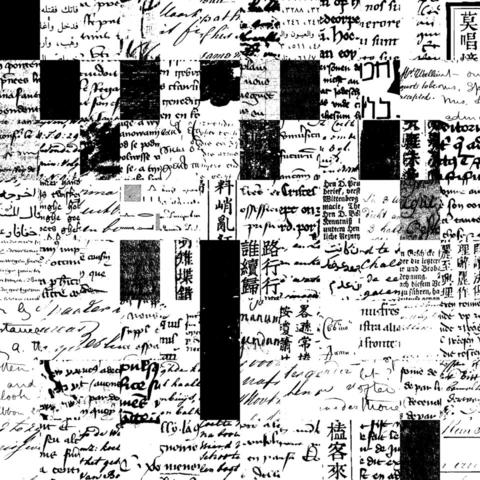}}
		\subcaption{Otsu}
	\end{subfigure} \hfill
	\begin{subfigure}{.23\textwidth}
		\fbox{\includegraphics[width=.99\linewidth]{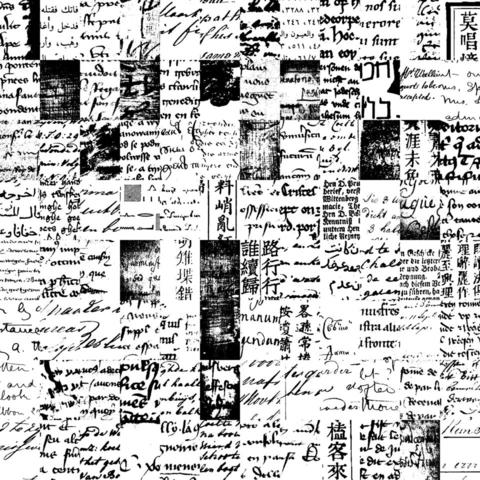}}
		\subcaption{Sauvola}	
	\end{subfigure} \hfill 
	\begin{subfigure}{.23\textwidth}	
		\fbox{\includegraphics[width=.99\linewidth]{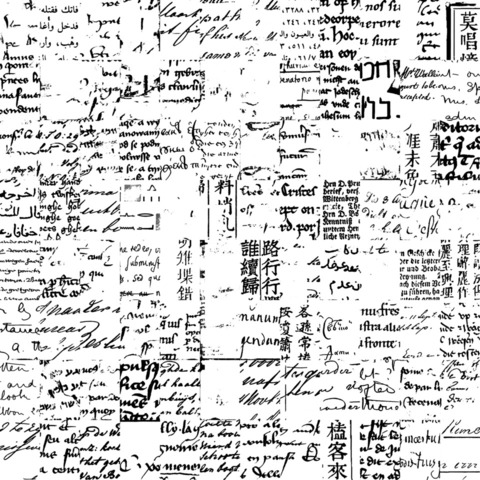}}
		\subcaption{BiNet}
	\end{subfigure}
	\caption{An illustration of BiNet output of four different test images (grid-images) created with various manuscript collections from Monk \cite{monknet}. The BiNet model used here is trained on DIBCO images. Results from Otsu (global) and Sauvola are presented as well.}
	\label{appen:fig:monkgrid}
\end{figure}

\begin{appendices} \label{appen:1}
This section contains additional information and results related to the article. Figure \ref{appen:fig:boxplot} illustrates the box-plots for evaluation metrics. The complete list of train and test images can be found in Table \ref{appen:train-test-DSS}, Table \ref{appen:train-plate}, and Table \ref{appen:train-test-DIBCO}. A brief configuration of the computing system is provided in Table \ref{appen:comp-config}. Figure \ref{appen:fig:plateimages} contains the binarization results of three test images (full-plates) of the DSS collection using BiNet with transfer learning. The BiNet is additionally tested on several different manuscript collections which is illustrated using the binarization of grid-images from Monk in Figure \ref{appen:fig:monkgrid}. Finally, the binarization results of DIBCO/H-DIBCO datasets are presented at the end of this section. 

\newpage

\begin{table}[h!]
	\centering
	\caption{List of train and test images along with material types (for DSS collection of fragment images).}
	\label{appen:train-test-DSS}
	\begin{tabular}{llll}
		\hline \hline
		\multicolumn{3}{c}{\textbf{Train-images (columns 1-3)}}  	 & \multicolumn{1}{c}{\textbf{Test-images}} \\ \hline \hline
		P25-Fg001-R-C01-R01                             & P497-Fg009-R-C01-R01                            & P976-Fg001-R-C01-R01                            & P385-Fg006-R-C01-R01                           \\ \hline
		P25-Fg005-R-C01-R01                             & P497-Fg009-R-C02-R01                            & P976-Fg002-R-C01-R01                            & P385-Fg010-R-C01-R01                           \\ \hline
		P124-Fg004-R-C01-R01                            & P524-Fg002-R-C01-R01                            & P976-Fg002-R-C01-R02                            & P385-Fg011-R-C01-R01                           \\ \hline
		P306-Fg001-R-C01-R01                            & P580-Fg004-R-C01-R01                            & P976-Fg003-R-C01-R01                            & P638-Fg001-R-C02-R01                           \\ \hline
		P307-Fg001-R-C01-R01                            & P580-Fg004-R-C02-R01                            & P976-Fg003-R-C01-R02                            & P638-Fg001-R-C06-R01                           \\ \hline
		P307-Fg003-R-C01-R01                            & P607-1-Fg001-R-C01-R01                          & P976-Fg004-R-C01-R01                            & P834-Fg001-R-C01-R01                           \\ \hline
		P405-Fg001-R-C01-R01                            & P607-1-Fg001-R-C01-R02                          & P976-Fg004-R-C01-R02                            & P834-Fg002-R-C01-R01                           \\ \hline
		P405-Fg001-R-C01-R02                            & P607-1-Fg001-R-C02-R01                          & P1001-Fg001-R-C01-R01                           & P980-Fg001-R-C01-R01                           \\ \hline
		P409-Fg001-R-C01-R01                            & P607-1-Fg001-R-C02-R02                          & P1001-Fg002-R-C01-R01                           & P1080-Fg001-R-C01-R01                          \\ \hline
		P470-Fg001-R-C01-R01                            & P704-1-Fg002-R-C01-R01                          & P1001-Fg008-R-C01-R01                           & P1080-Fg006-R-C01-R01                          \\ \hline
		P470-Fg002-R-C01-R01                            & P704-1-Fg002-R-C02-R01                          & P1001-Fg010-R-C01-R01                           & P1082-Fg001-R-C01-R02                          \\ \hline
		P480-Fg006-R-C01-R01                            & P807-Fg019-R-C01-R01                            & P1081A-Fg002-R-C01-R01                          &                                                \\ \hline
		P480-Fg007-R-C01-R01                            & P807-Fg019-R-C02-R01                            &                                                 &                                                \\ \hline
		P480-Fg007-R-C02-R01                            & P807-Fg019-R-C03-R01                            &                                                 &                                                \\ \hline \hline
		\multicolumn{4}{c}{\textbf{Fragment material types}}                                                                                                                                                 \\ \hline
		\textbf{}                                       & \multicolumn{1}{c}{\textbf{Train-images}}       & \multicolumn{1}{c}{\textbf{Test-images}}        & \multicolumn{1}{c}{\textbf{Total}}             \\ \hline
		\textbf{Parchment}                              & \multicolumn{1}{c}{38}                          & \multicolumn{1}{c}{9}                           & \multicolumn{1}{c}{47}                         \\ \hline
		\textbf{Papyrus}                                & \multicolumn{1}{c}{2}                           & \multicolumn{1}{c}{2}                           & \multicolumn{1}{c}{4}                          \\ \hline
		\textbf{Total}                                  & \multicolumn{1}{c}{40}                          & \multicolumn{1}{c}{11}                          & \multicolumn{1}{c}{51}                         \\ \hline \hline
	\end{tabular}
\end{table}%
\begin{table}[h!]
	\centering
	\caption{List of train (for transfer-learning) and test images (for DSS collection of full-plate images).}
	\label{appen:train-plate}
	\begin{tabular}{llllllc}
		\hline \hline
		\multicolumn{6}{c}{\textbf{Train-images (columns 1-6)}}     & \textbf{Test-images} \\ \hline \hline
		25      & 307     & 470     & 580       & 807     & 1081A   & 1                    \\ \hline
		124     & 405     & 480     & 607-1     & 976     &         & 386                  \\ \hline
		306     & 409     & 524     & 704-1     & 1001    &         & 997                  \\ \hline \hline
		\multicolumn{7}{l}{\textbf{Total number of images:} \hfill Train: 16, \hfil Test: 3}    \\ \hline \hline
	\end{tabular}
\end{table}%
\begin{table}[h!]
	\centering
	\caption{List of train and test images (from DIBCO/H-DIBCO \cite{gatos2009icdar, pratikakis2010h, pratikakis2012icfhr, pratikakis2013icdar, ntirogiannis2014icfhr2014, pratikakis2016icfhr2016, pratikakis2017icdar2017, pratikakis2018icfhr2018}).}
	\label{appen:train-test-DIBCO}
	\begin{tabular}{lll}
		\hline \hline
		\multicolumn{2}{c}{\textbf{Train-images (columns 1-2)}} & \multicolumn{1}{c}{\textbf{Test-images}} \\ \hline \hline
		DIBCO 2009                & DIBCO 2012                  & H-DIBCO 2016                             \\ \hline
		DIBCO 2010                & DIBCO 2013                  & DIBCO 2017                               \\ \hline
		DIBCO 2011                & H-DIBCO 2014                & H-DIBCO 2018                             \\ \hline \hline
		\multicolumn{3}{l}{\textbf{Total number of images:} \hfill Train: 76, \hfil Test: 40}              \\ \hline \hline
	\end{tabular}
\end{table}%
\begin{table}[h!]
	\centering
	\caption{Brief configuration of the work-station.}
	\label{appen:comp-config}
	\begin{tabular}{lll}
		\hline \hline
		\multicolumn{1}{c}{\textbf{CPU}}             & \multicolumn{1}{c}{\textbf{Memory}} & \multicolumn{1}{c}{\textbf{Display card}} \\ \hline \hline
		Intel(R) Core(TM) i5-4590 @ 3.30GHz 		 & description: System memory          & GP106 {[}GeForce GTX 1060 6GB{]} 		   \\ \hline
		size: 3583MHz                                & size: 7730MiB                       & vendor: NVIDIA Corporation                \\ \hline
		capacity: 3700MHz; width: 64 bits            &                                     & width: 64 bits; clock: 33MHz              \\ \hline \hline
	\end{tabular}
\end{table}%
\begin{table}[h!]
	\centering
	\caption{Time needed to binarize one of the test images (fragment 1, plate 1082)}
	\label{appen:time}
	\begin{tabular}{ccc}
		\hline \hline
		\textbf{Grayscale image} & \textbf{RGB-color image} & \textbf{Pseudo-color image} \\ \hline
		0 min 28.905 sec & 0 min 34.991 sec            & 1 min 5.987 sec \\ \hline \hline
	\end{tabular}
\end{table}%

\newpage

\begin{table}[h!]
	\centering
	\begin{tabular}{|c|c|c|}
		\hline
		\includegraphics[width=.3\textwidth]{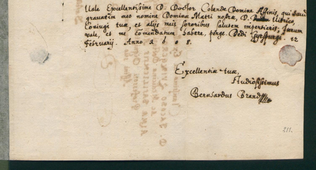} & \includegraphics[width=.3\textwidth]{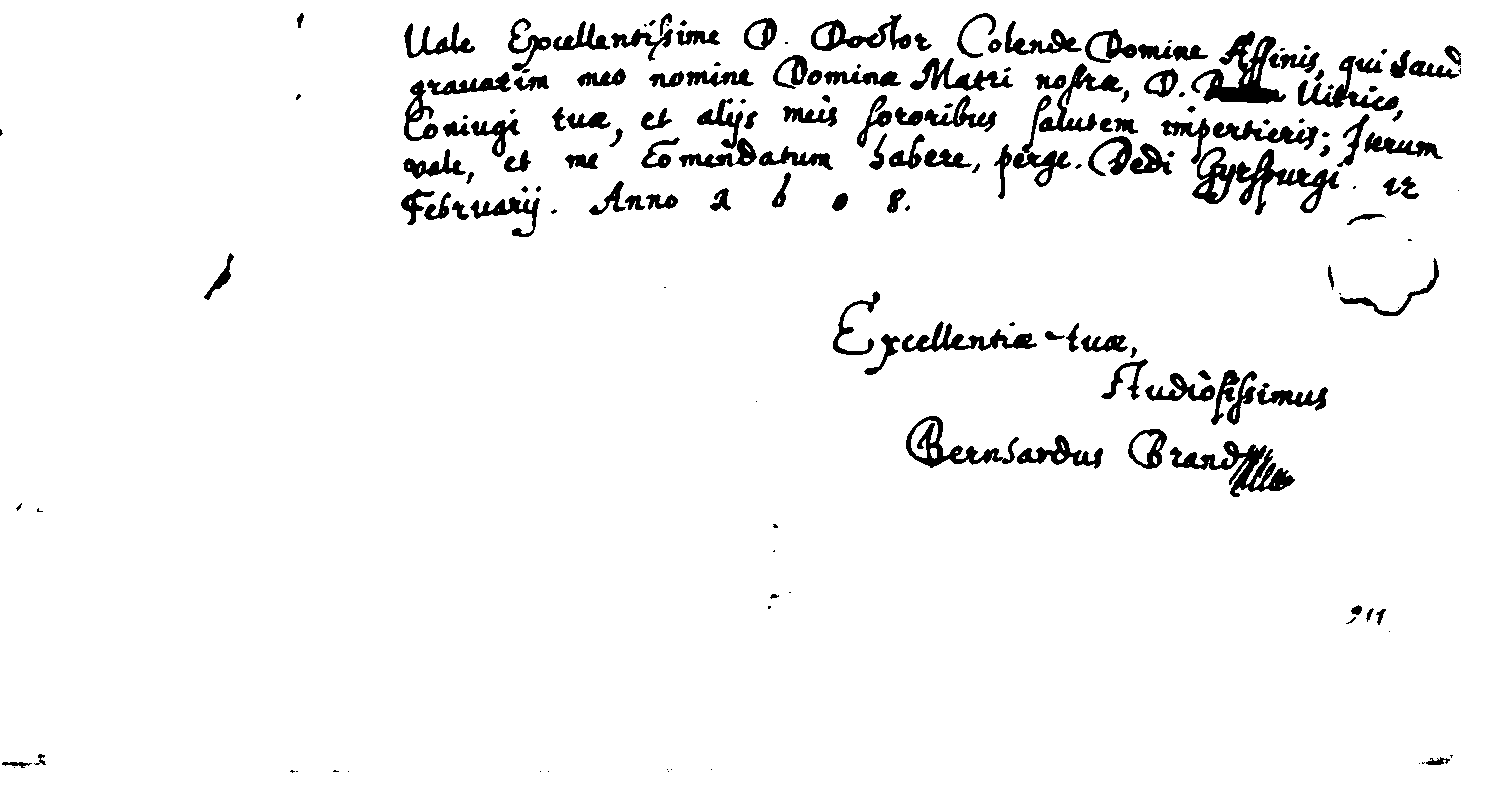} & \includegraphics[width=.3\textwidth]{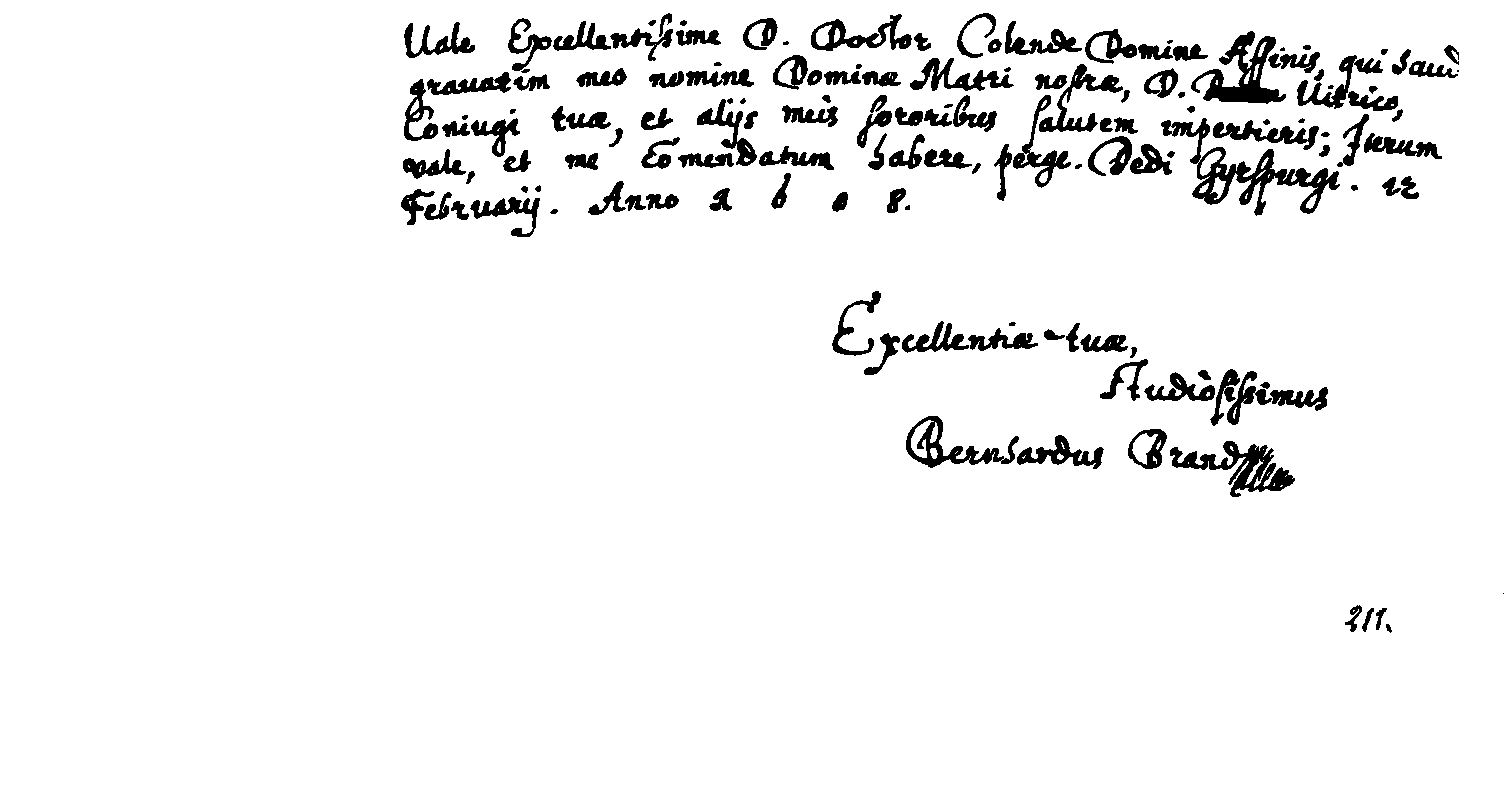} \\ \hline
		\includegraphics[width=.3\textwidth]{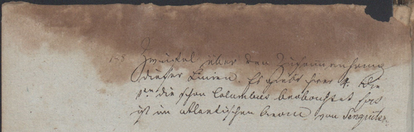} & \includegraphics[width=.3\textwidth]{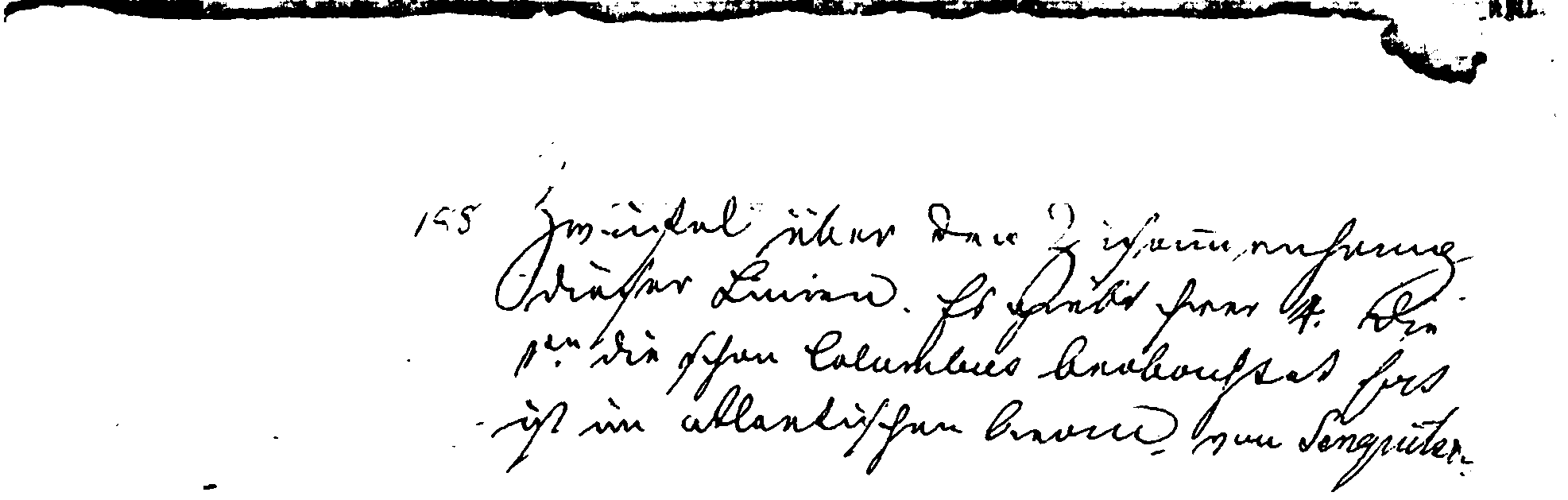} & \includegraphics[width=.3\textwidth]{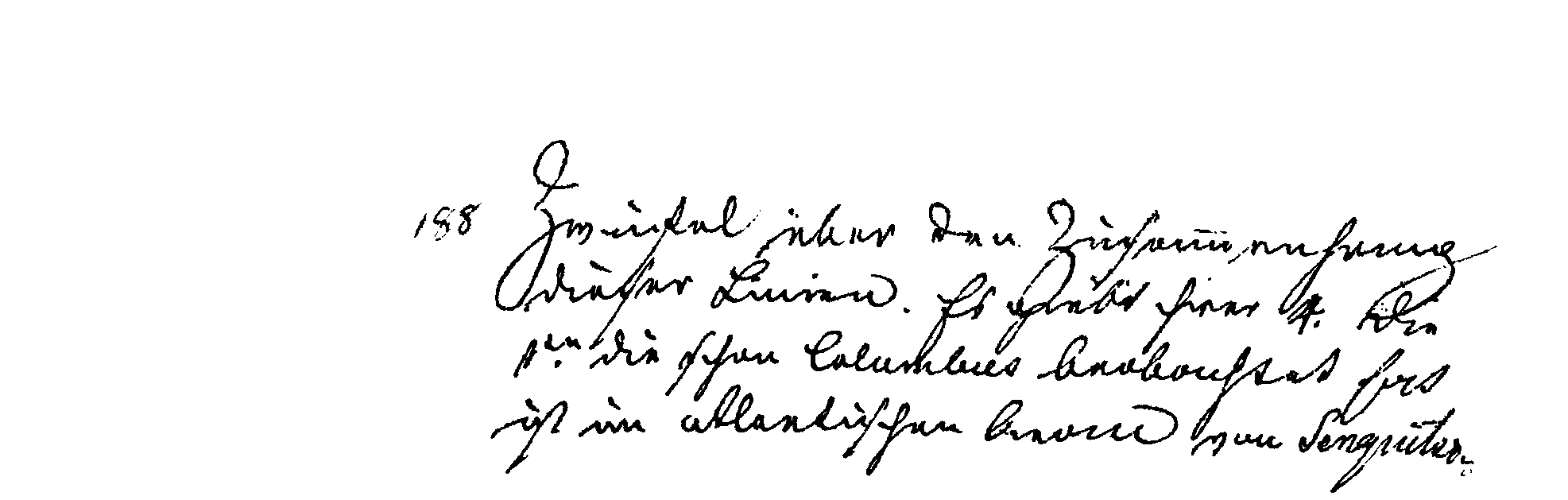} \\ \hline
		\includegraphics[width=.3\textwidth]{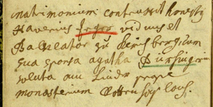} & \includegraphics[width=.3\textwidth]{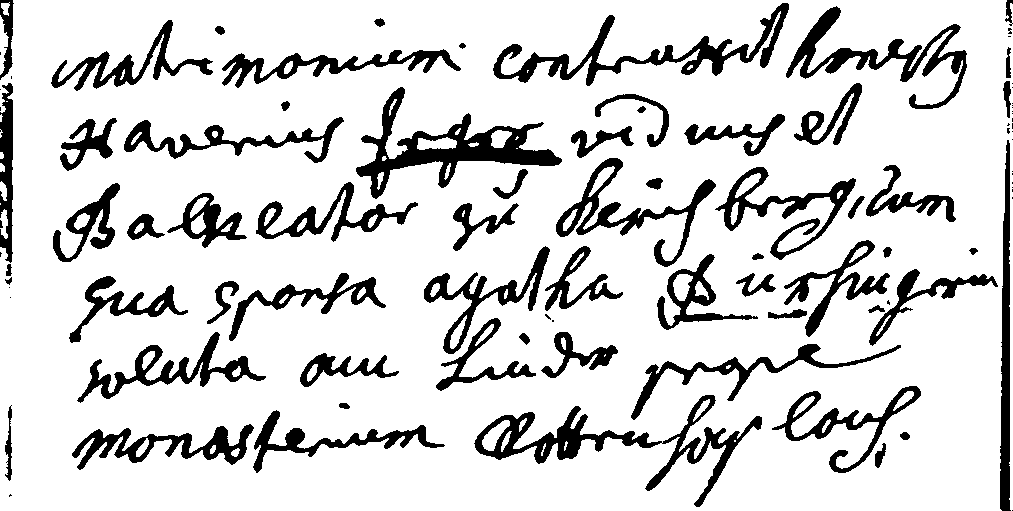} & \includegraphics[width=.3\textwidth]{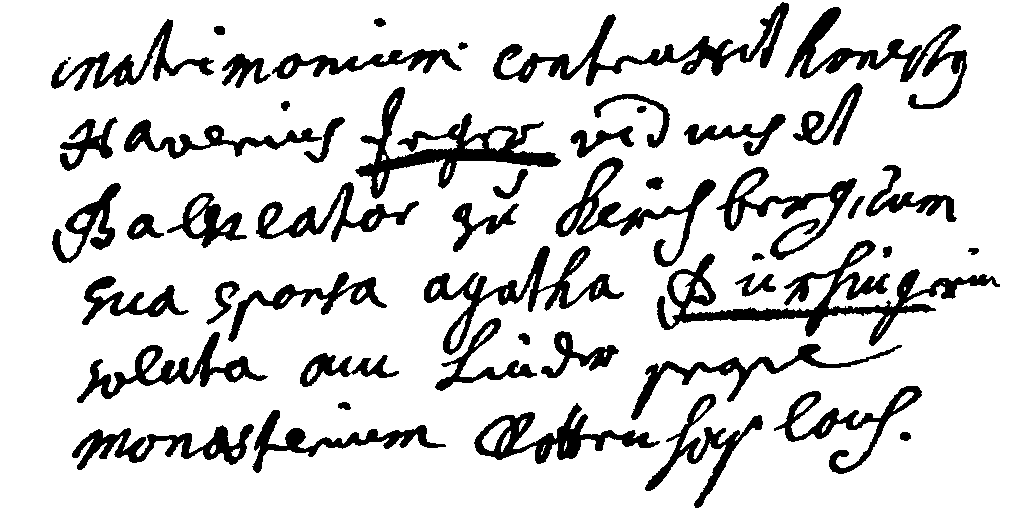} \\ \hline
		\includegraphics[width=.3\textwidth]{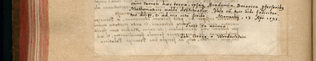} & \includegraphics[width=.3\textwidth]{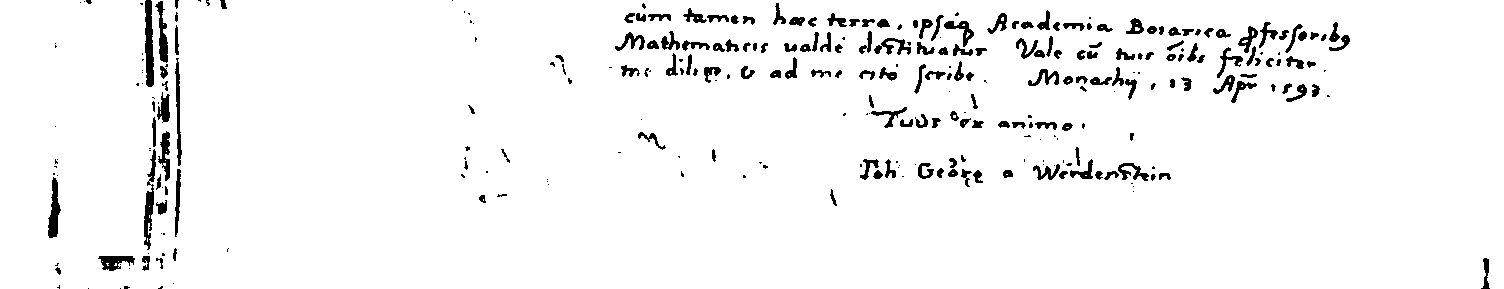} & \includegraphics[width=.3\textwidth]{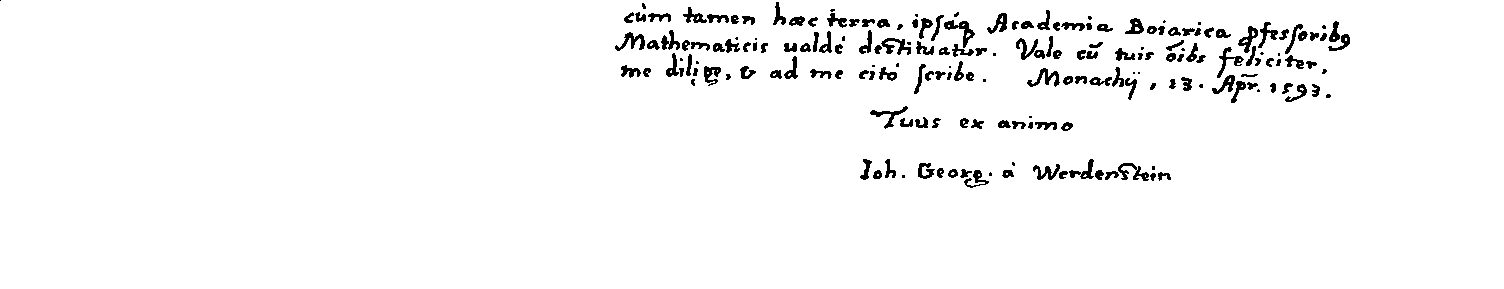} \\ \hline
		\includegraphics[width=.3\textwidth]{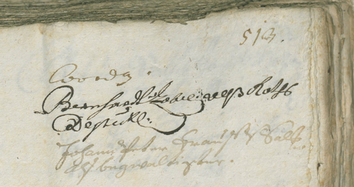} & \includegraphics[width=.3\textwidth]{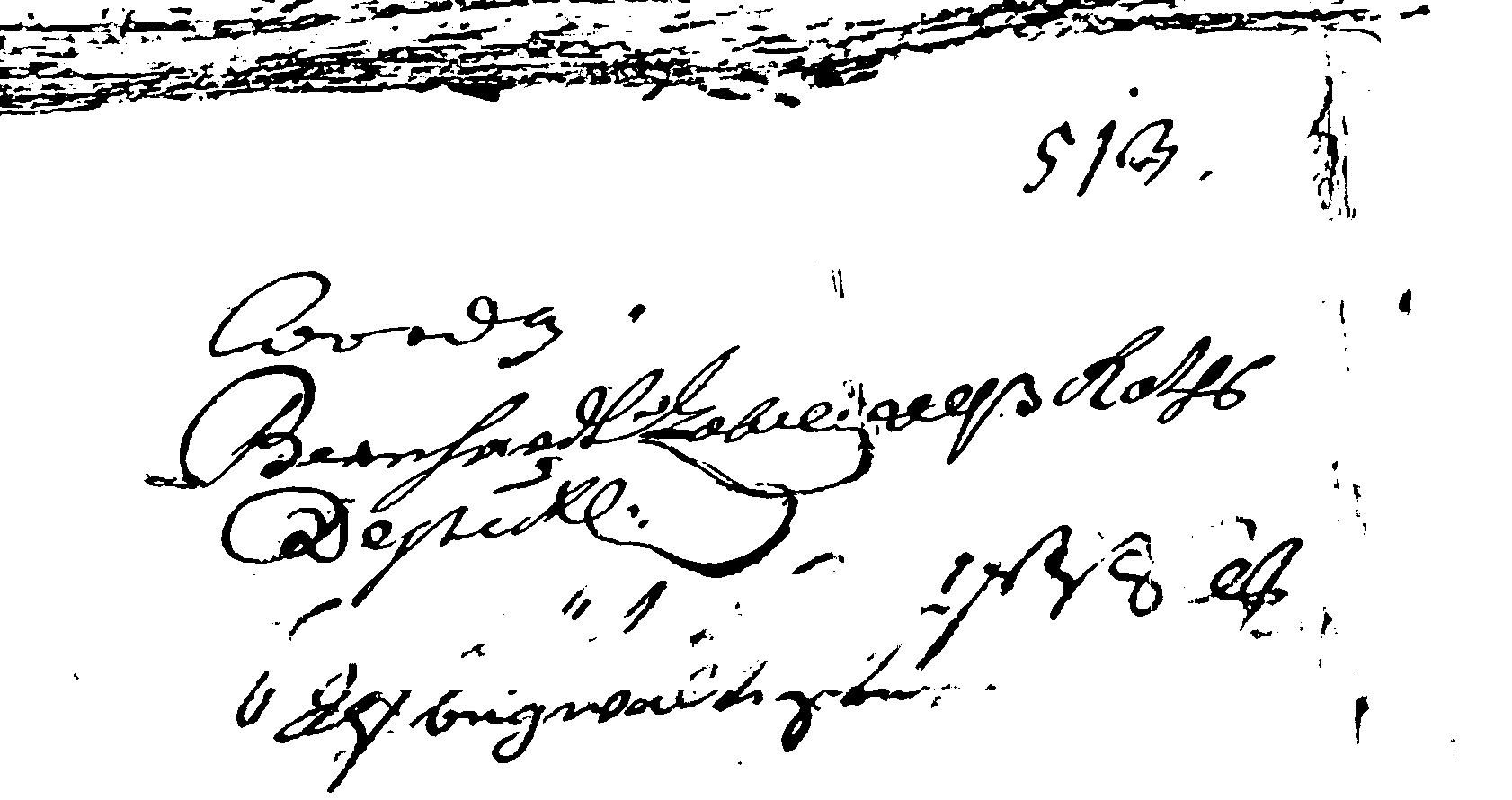} & \includegraphics[width=.3\textwidth]{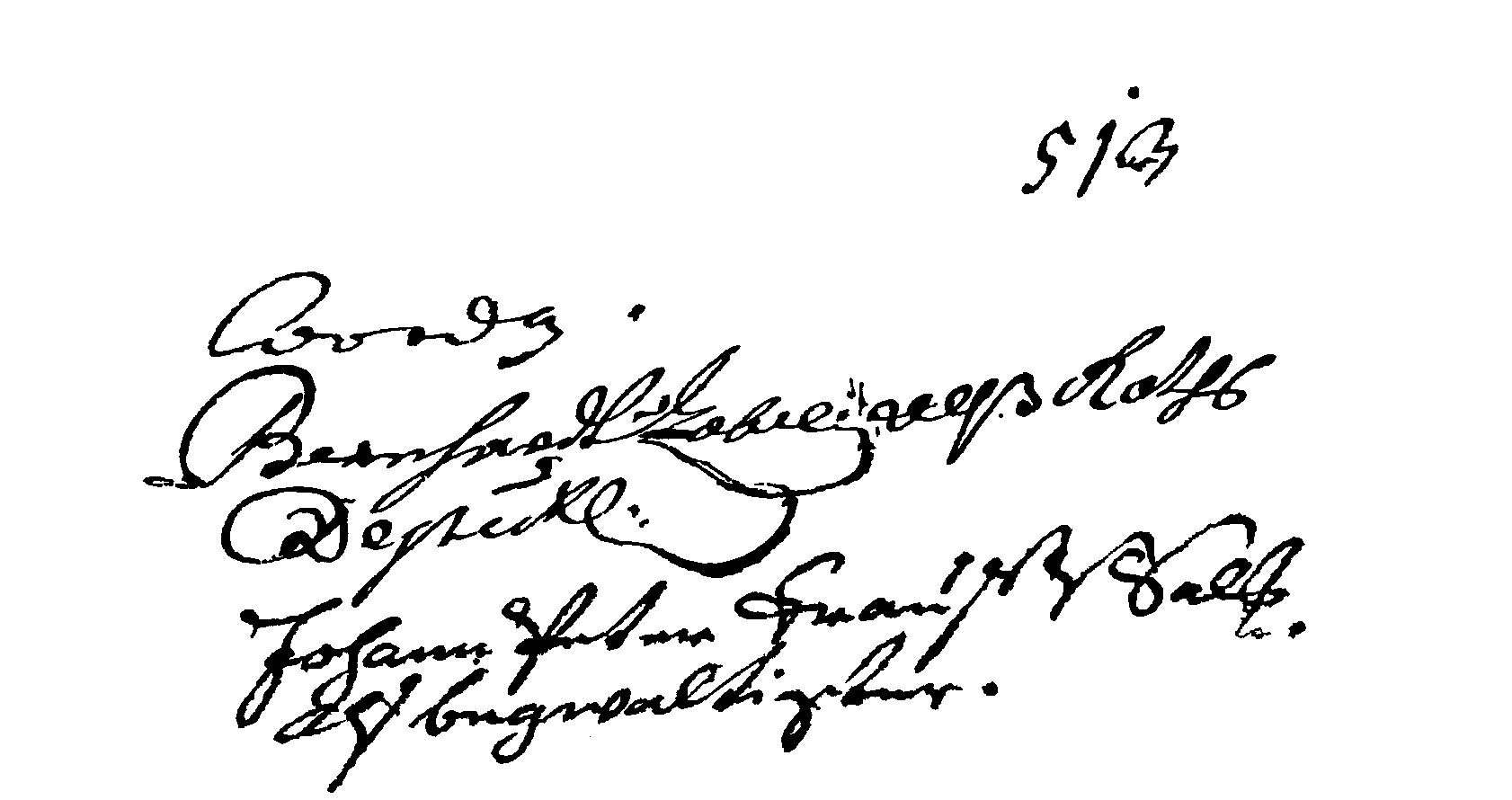} \\ \hline
		\includegraphics[width=.3\textwidth]{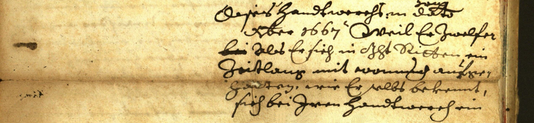} & \includegraphics[width=.3\textwidth]{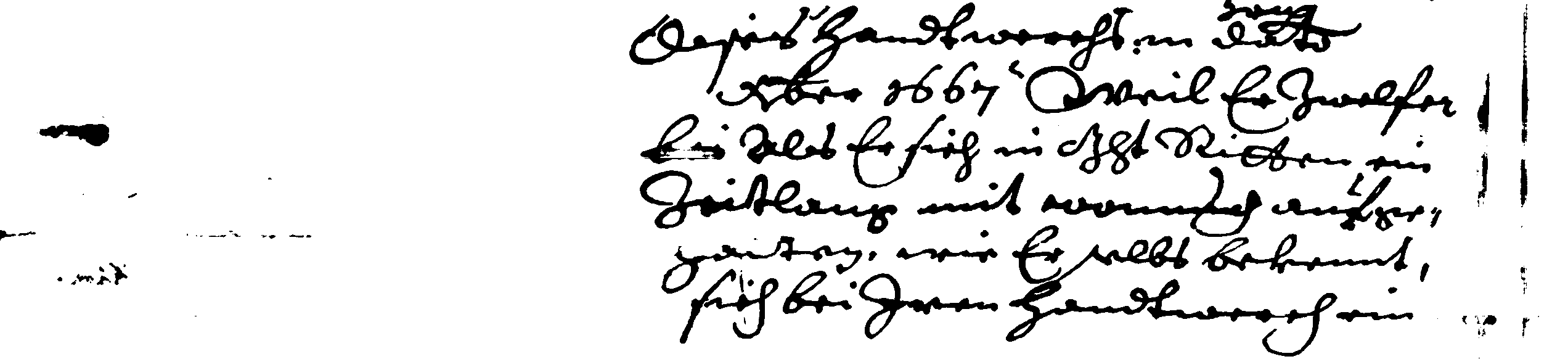} & \includegraphics[width=.3\textwidth]{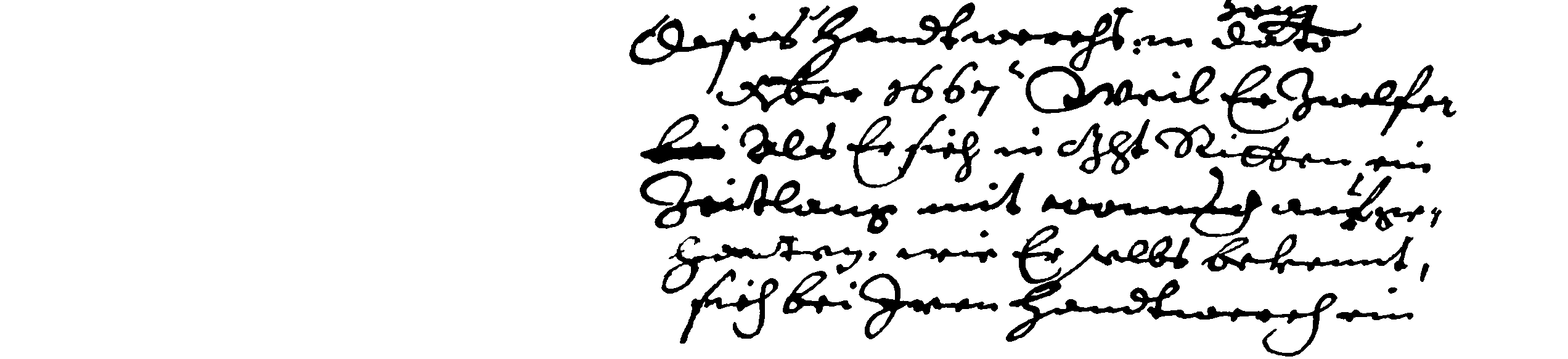} \\ \hline
		\includegraphics[width=.3\textwidth]{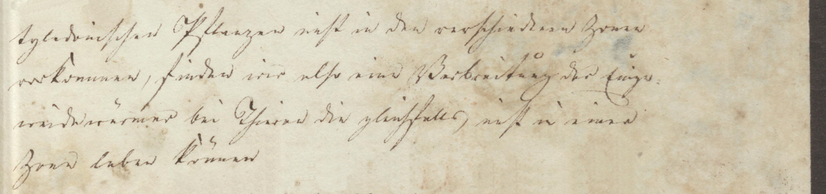} & \includegraphics[width=.3\textwidth]{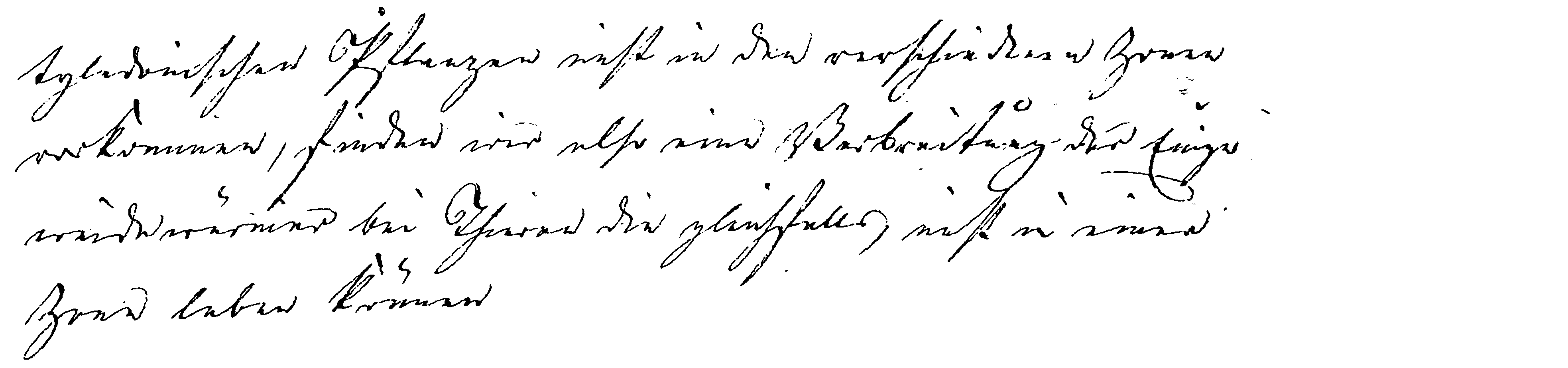} & \includegraphics[width=.3\textwidth]{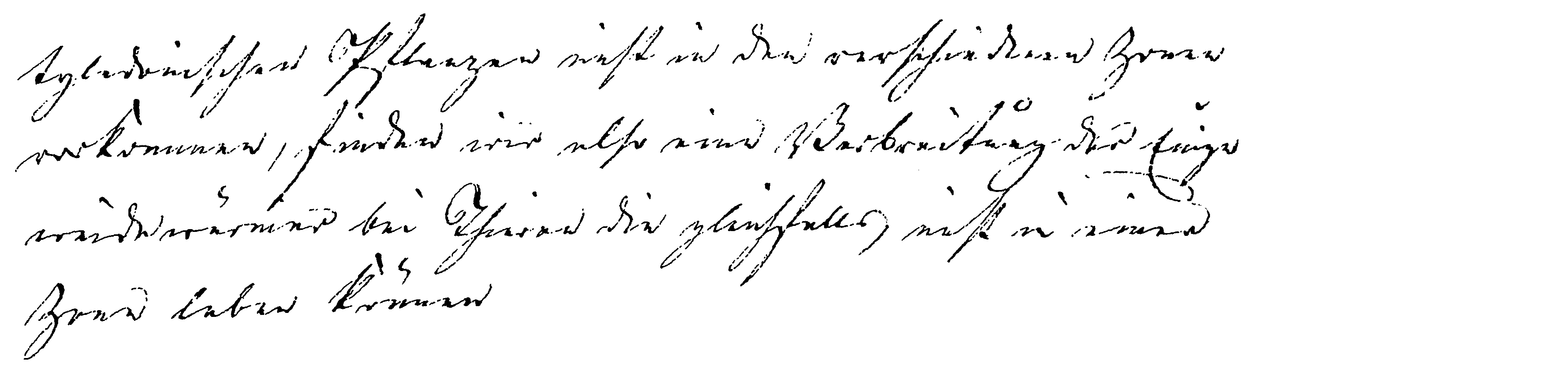} \\ \hline
		\includegraphics[width=.3\textwidth]{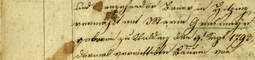} & \includegraphics[width=.3\textwidth]{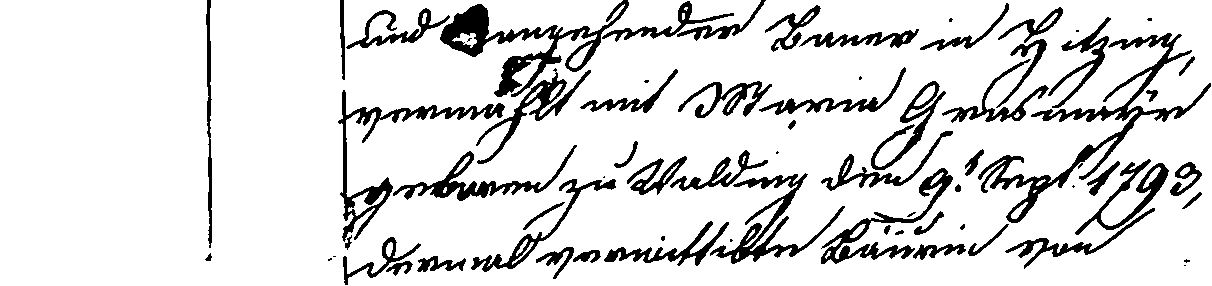} & \includegraphics[width=.3\textwidth]{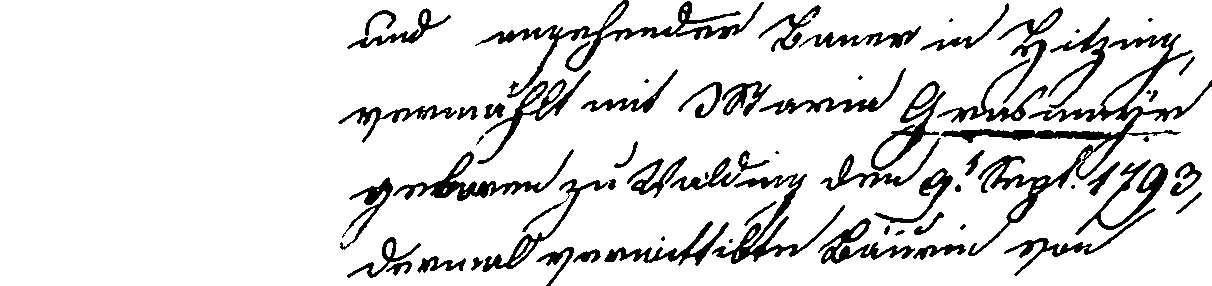} \\ \hline
		\includegraphics[width=.3\textwidth]{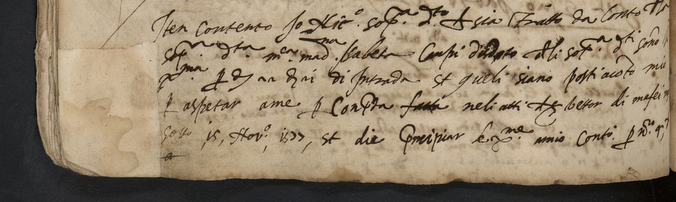} & \includegraphics[width=.3\textwidth]{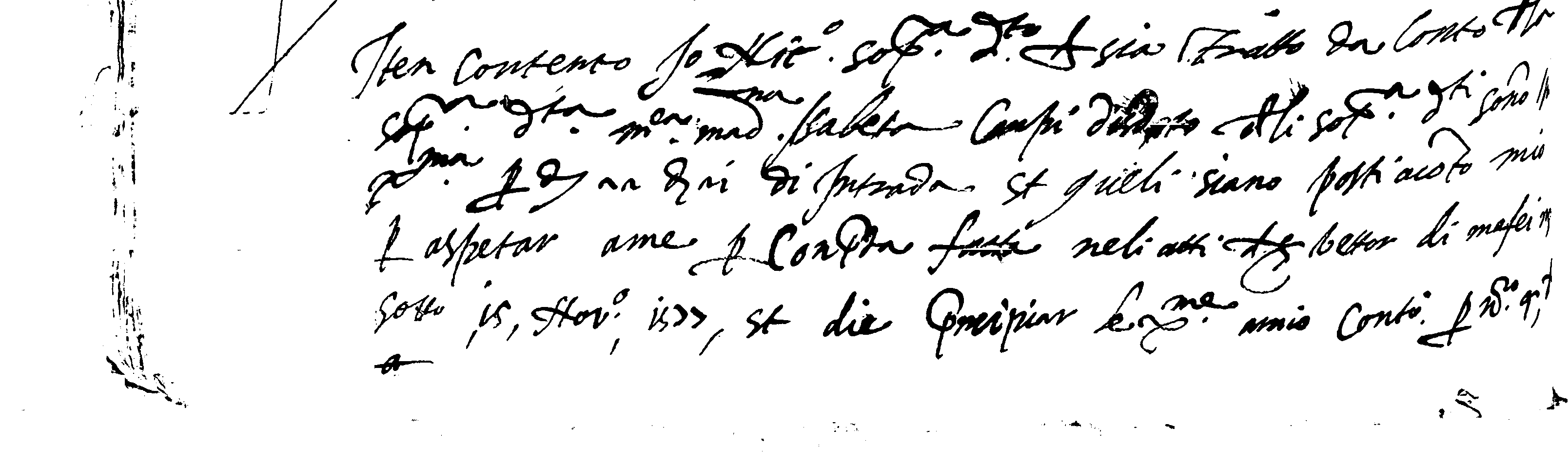} & \includegraphics[width=.3\textwidth]{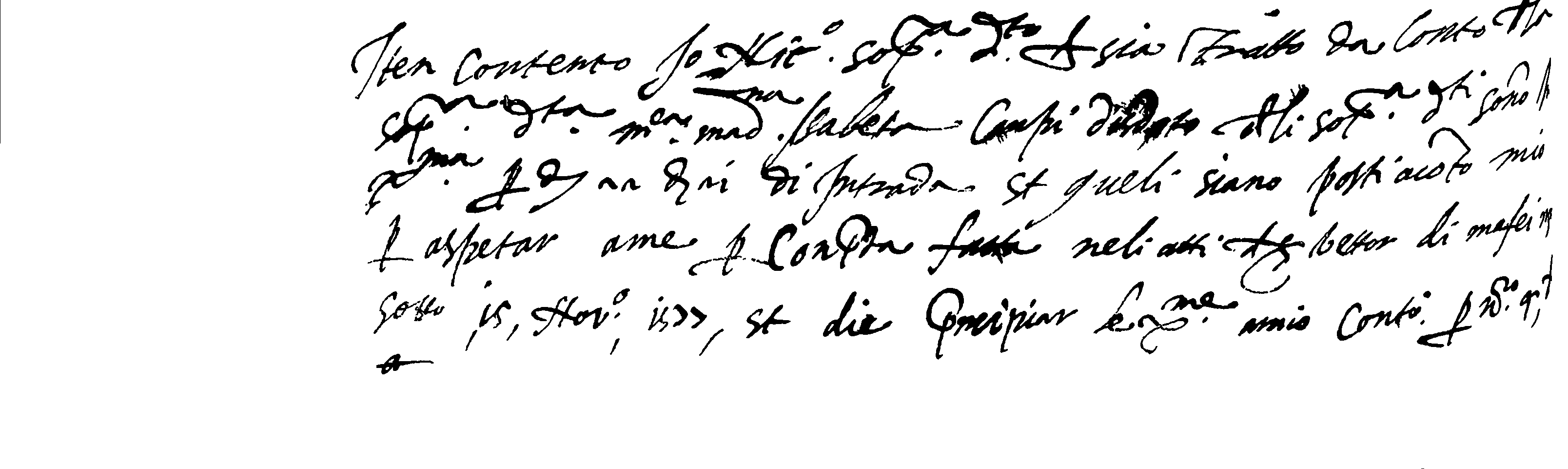} \\ \hline
		\includegraphics[width=.3\textwidth]{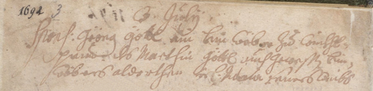} & \includegraphics[width=.3\textwidth]{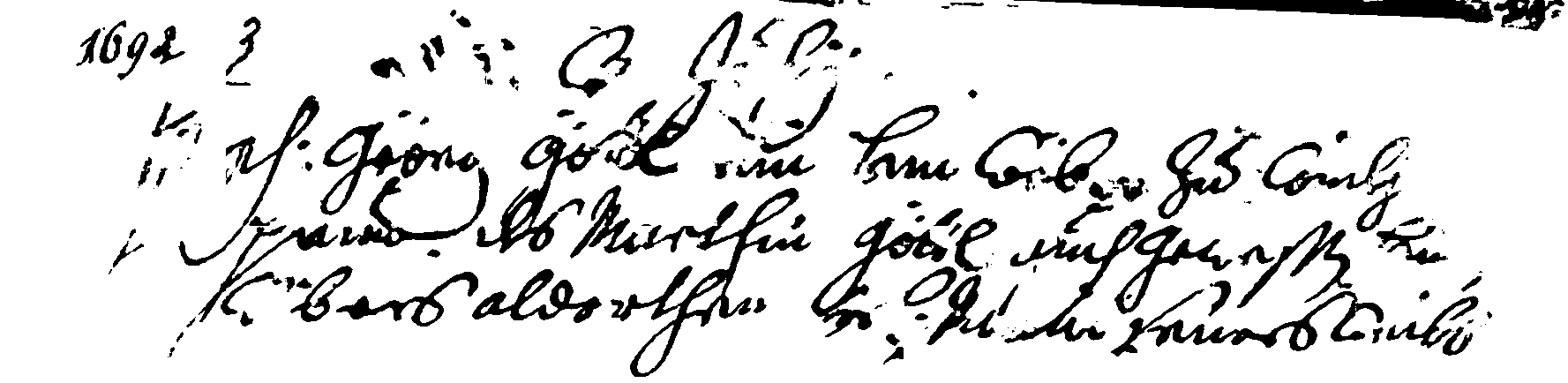} & \includegraphics[width=.3\textwidth]{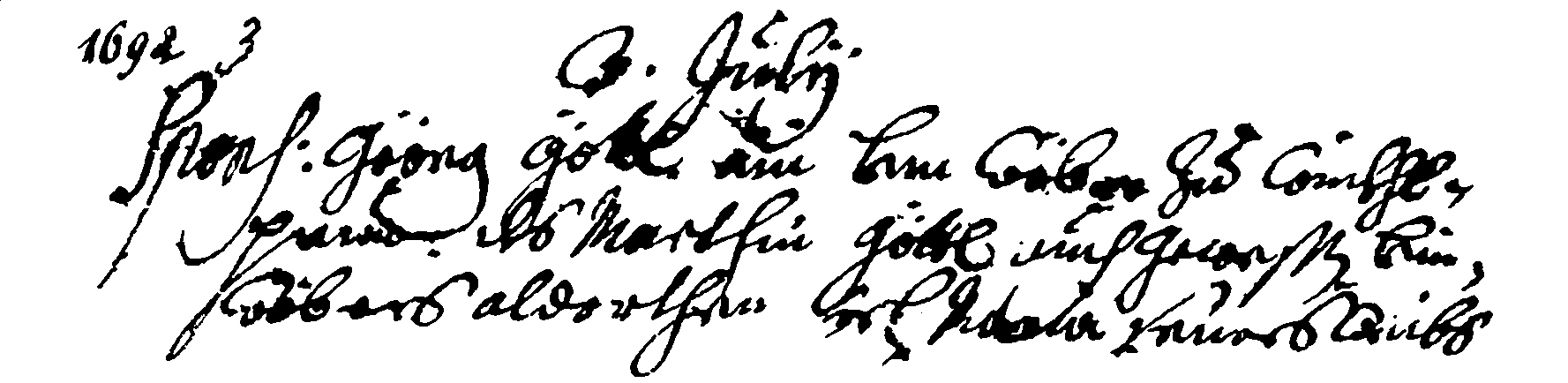} \\ \hline 
		Original image                                                               & BiNet output                                                                   & Ground-truth                                                                \\ \hline
	\end{tabular}
	\caption{Figures of the H-DIBCO 2018 testing dataset along with the binarization results from BiNet.}
	\label{appen:dibco2018}
\end{table}

\begin{table}[h!]
	\centering
	\begin{tabular}{|c|c|c|}
		\hline
		\includegraphics[width=.22\textwidth]{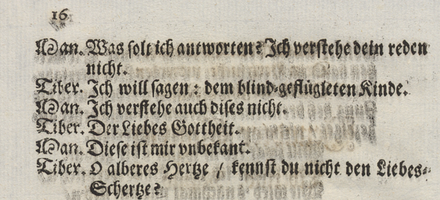} & \includegraphics[width=.22\textwidth]{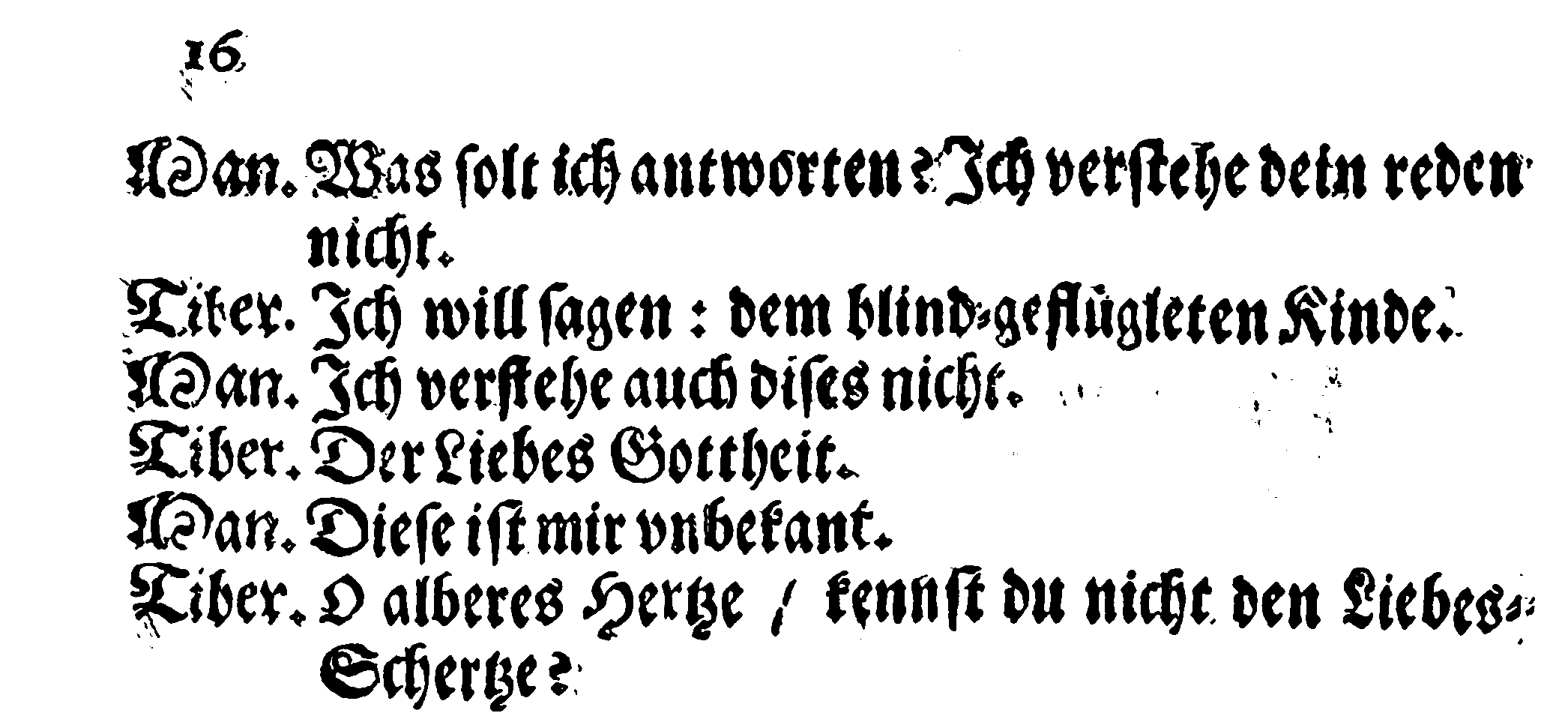} & \includegraphics[width=.22\textwidth]{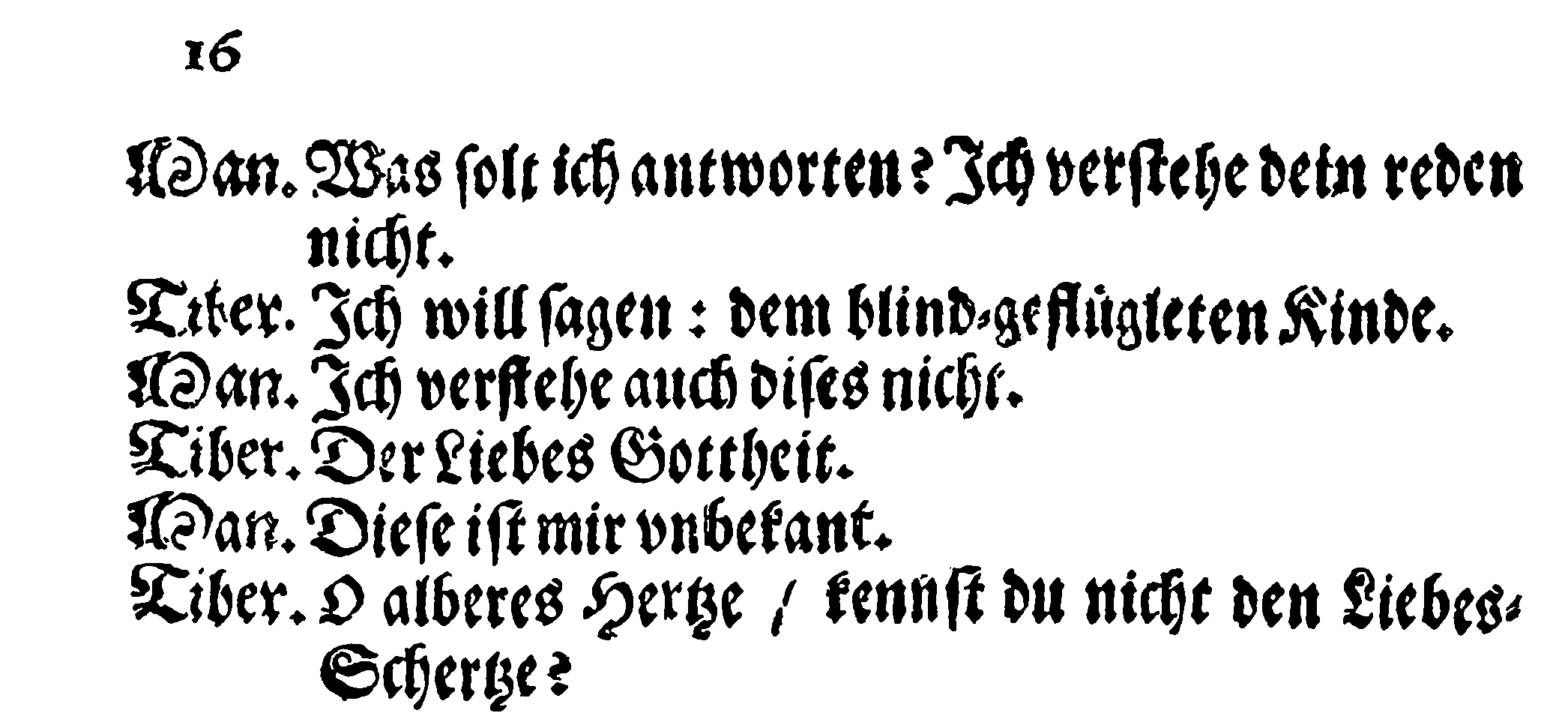} \\ \hline
		\includegraphics[width=.20\textwidth]{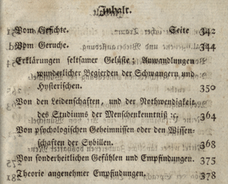} & \includegraphics[width=.20\textwidth]{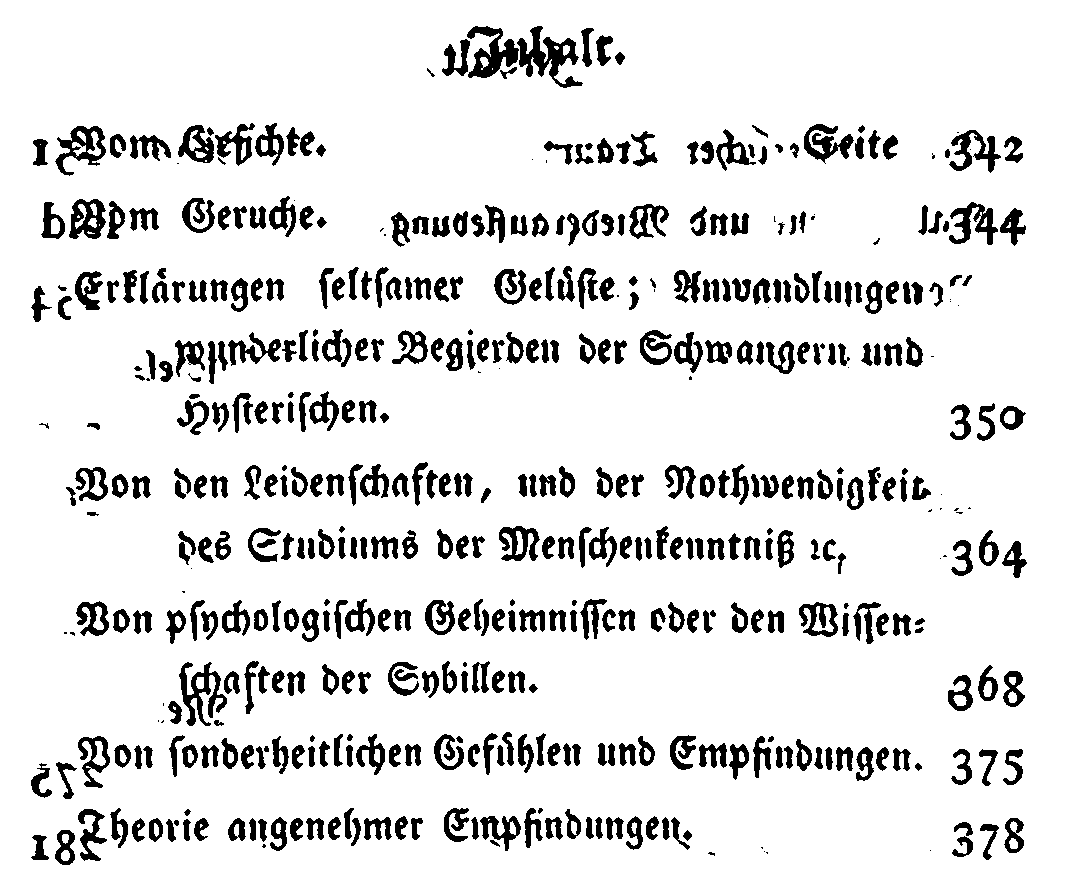} & \includegraphics[width=.20\textwidth]{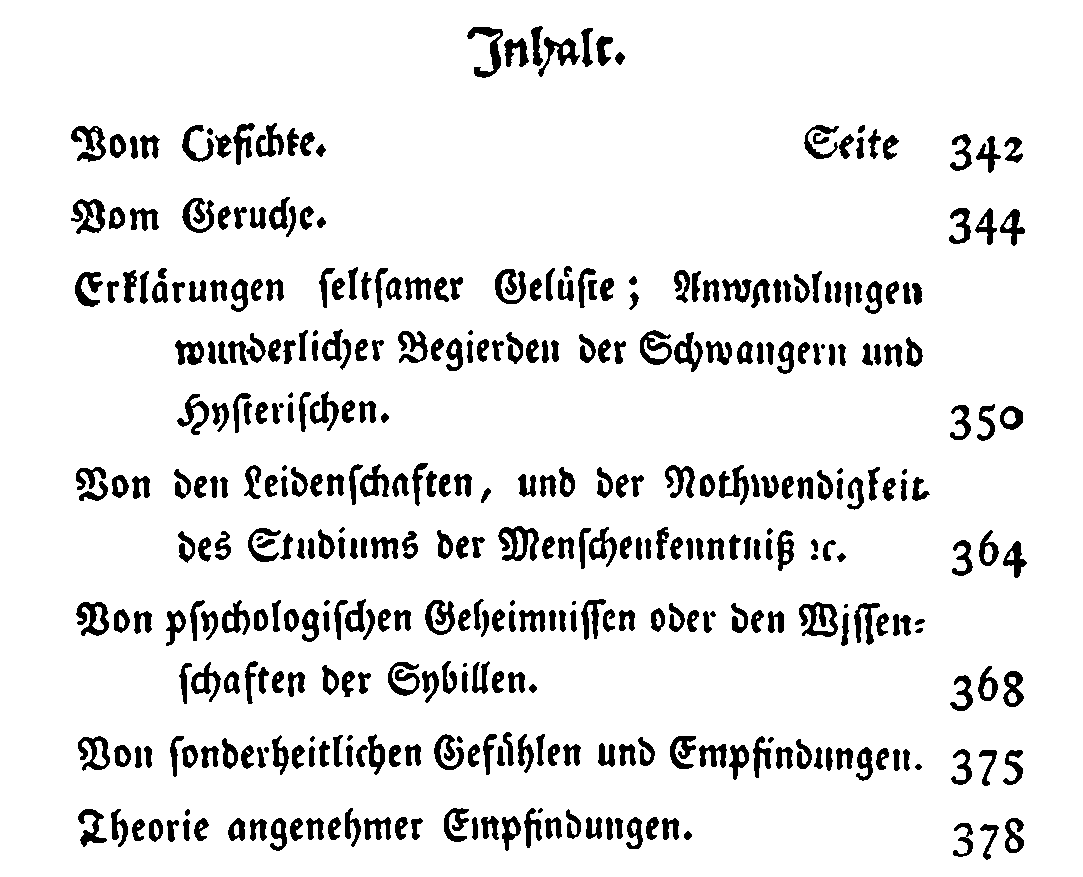} \\ \hline
		\includegraphics[width=.22\textwidth]{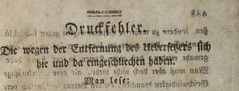} & \includegraphics[width=.22\textwidth]{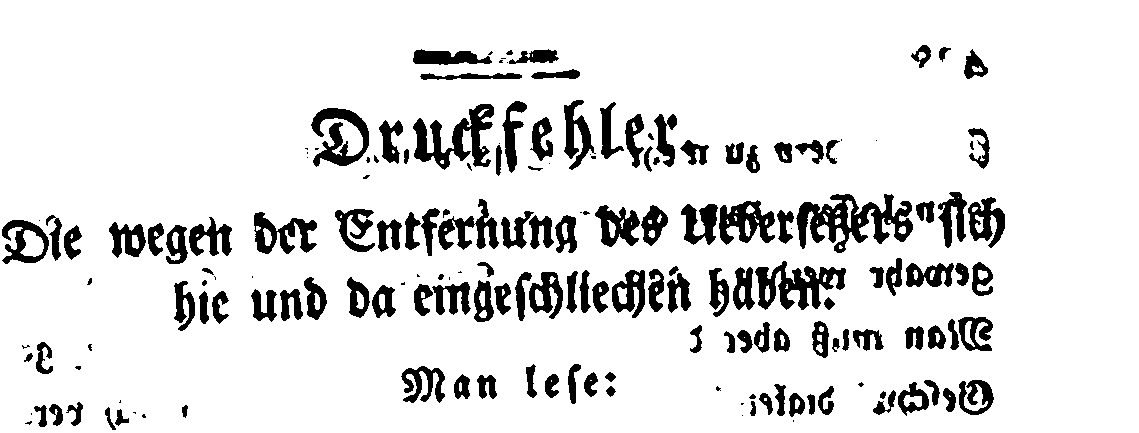} & \includegraphics[width=.22\textwidth]{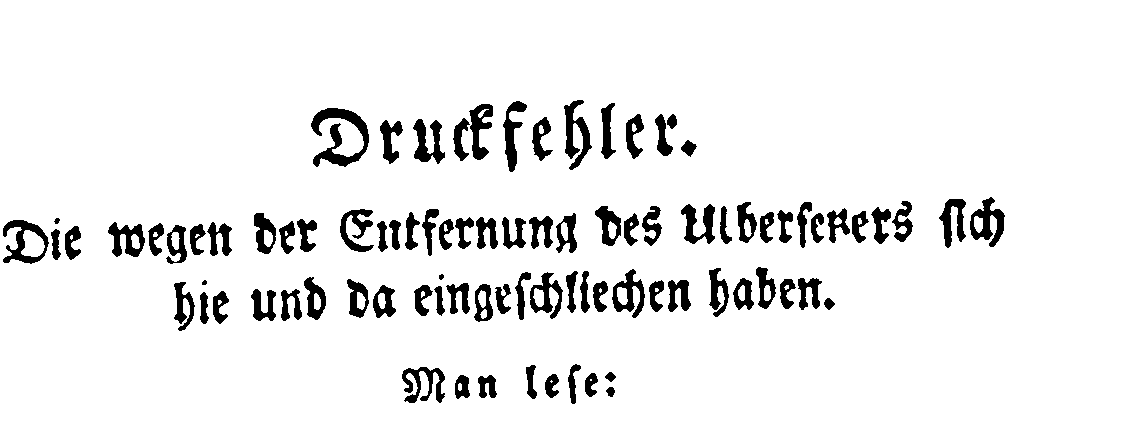} \\ \hline
		\includegraphics[width=.22\textwidth]{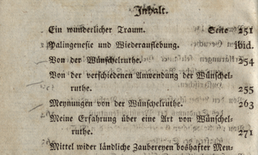} & \includegraphics[width=.22\textwidth]{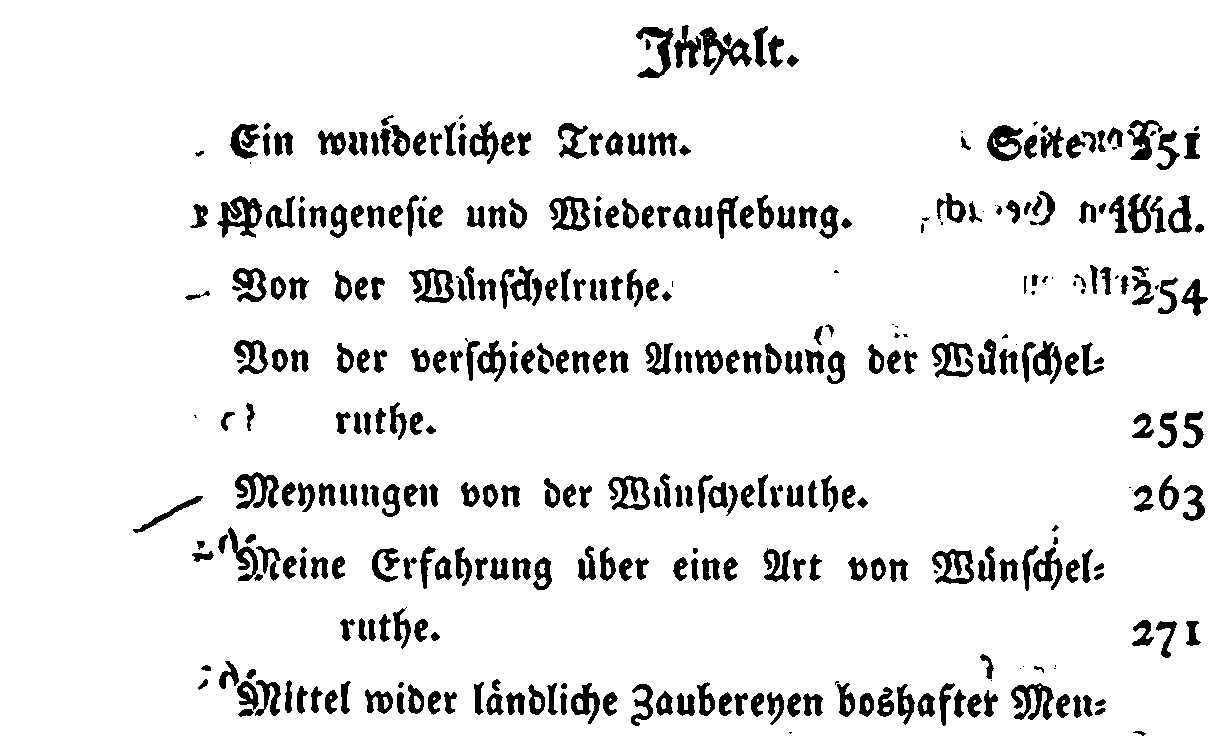} & \includegraphics[width=.22\textwidth]{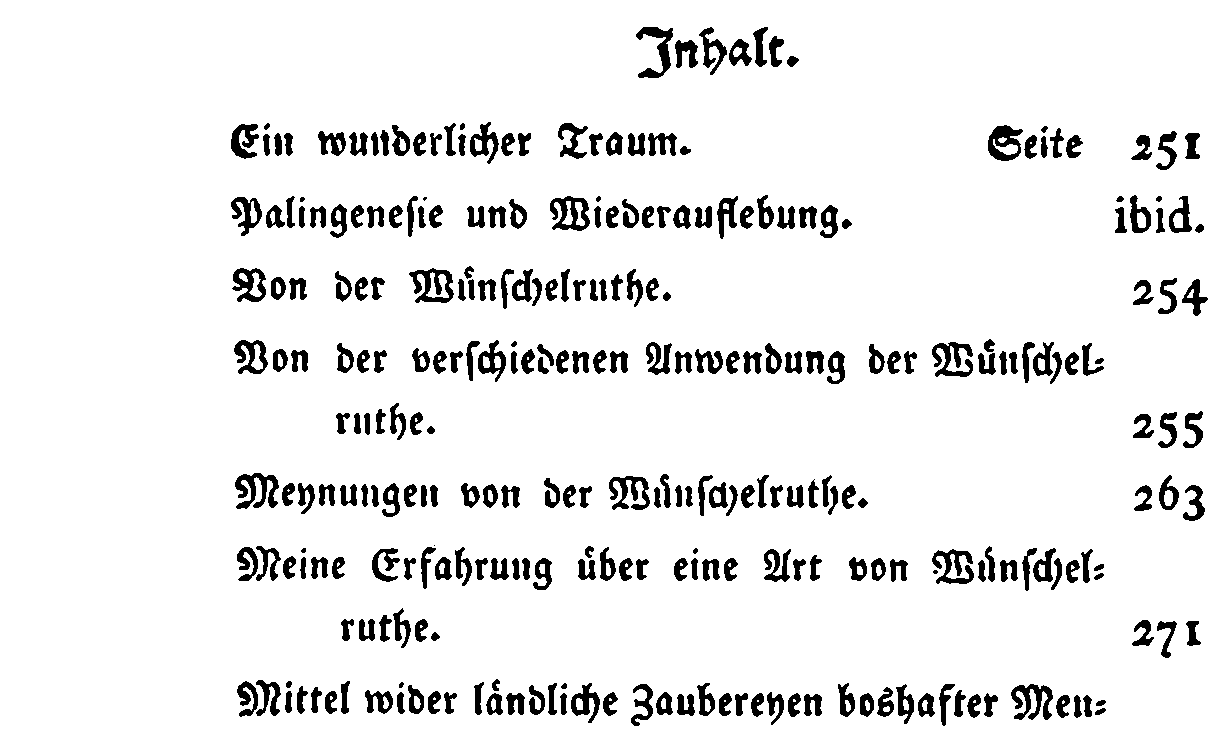} \\ \hline
		\includegraphics[width=.22\textwidth]{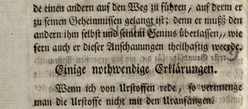} & \includegraphics[width=.22\textwidth]{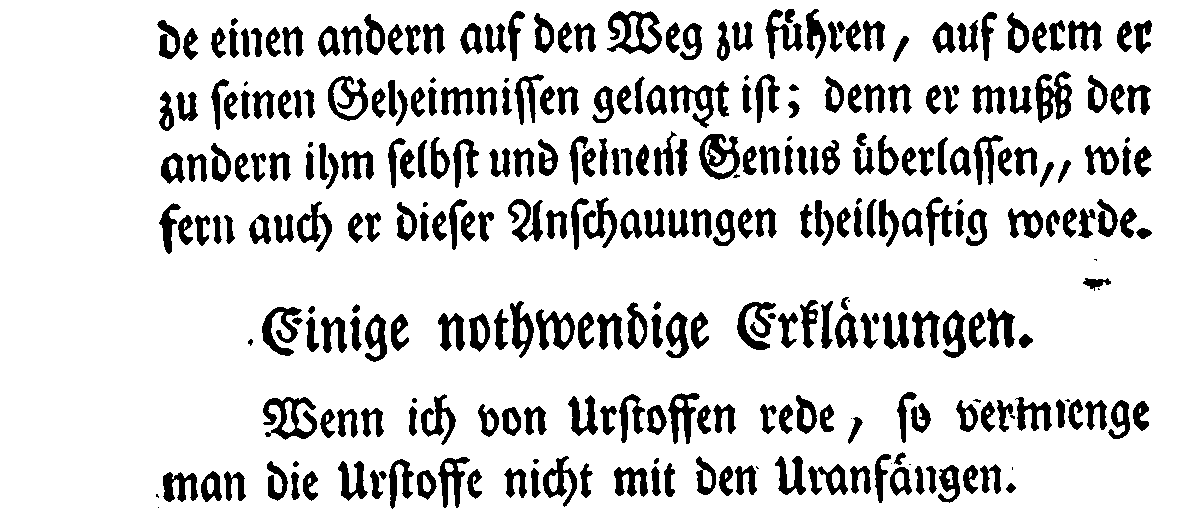} & \includegraphics[width=.22\textwidth]{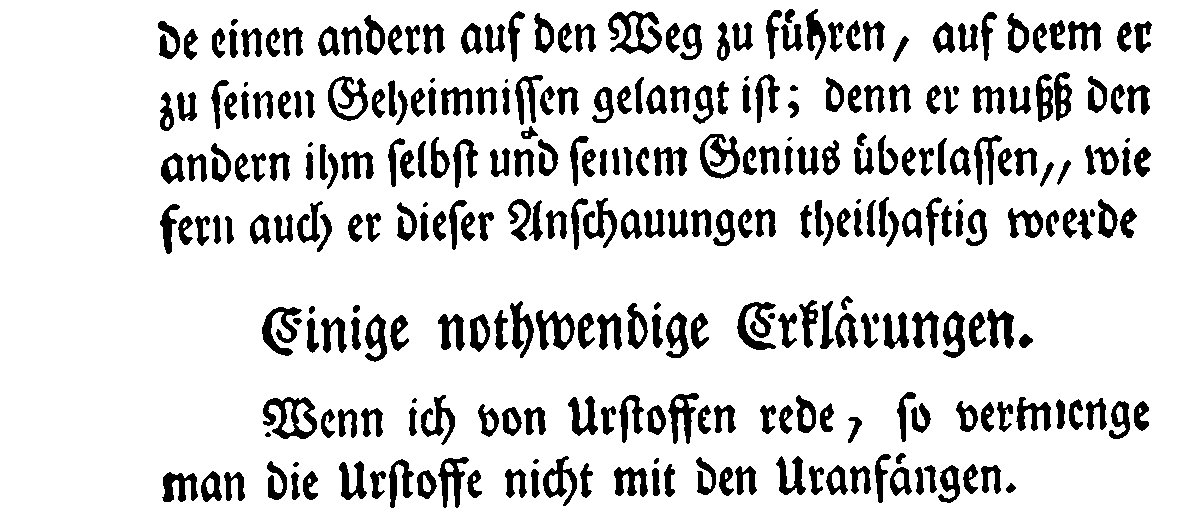} \\ \hline
		\includegraphics[width=.22\textwidth]{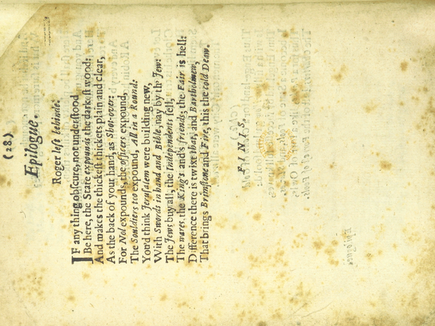} & \includegraphics[width=.22\textwidth]{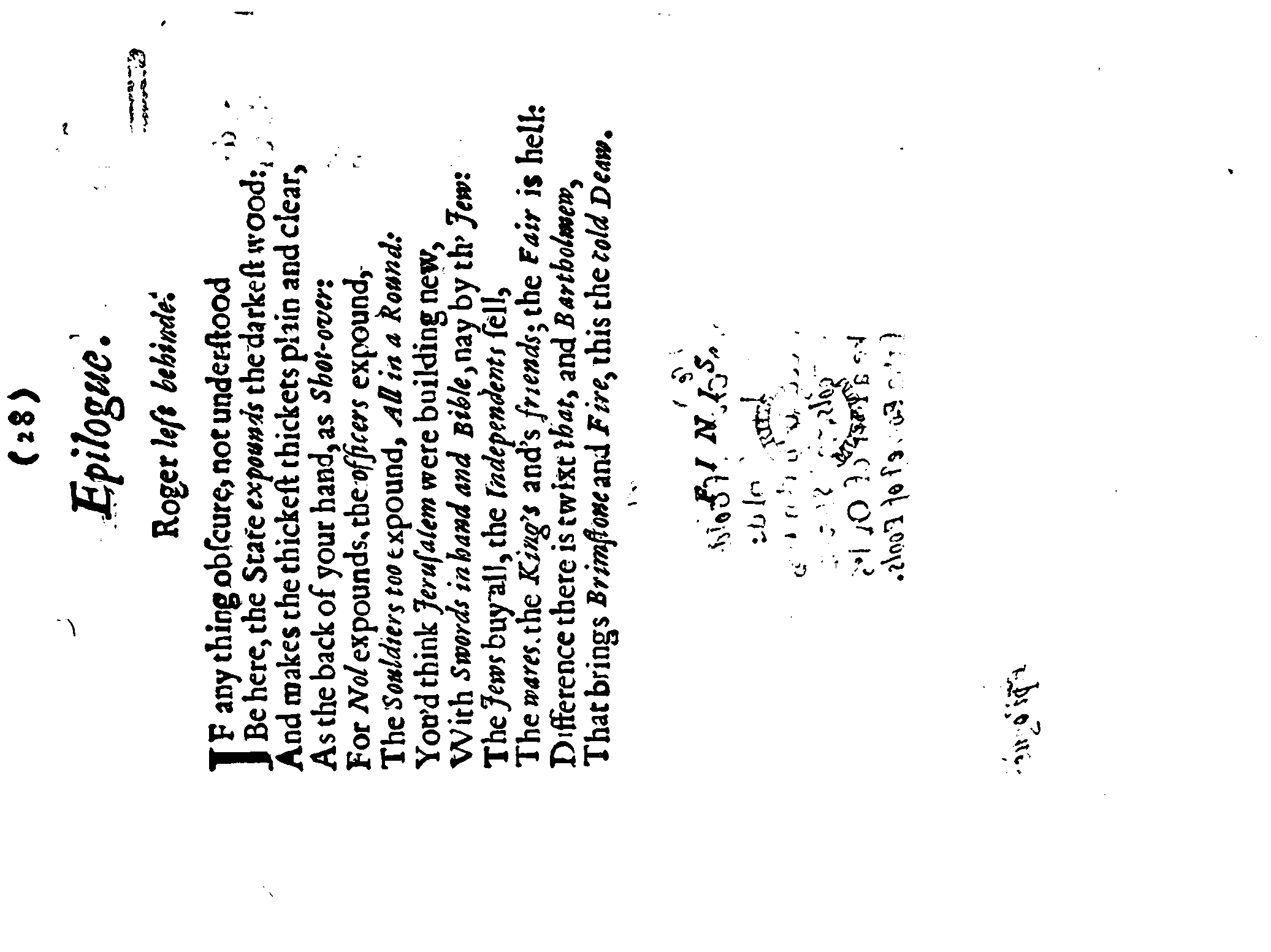} & \includegraphics[width=.22\textwidth]{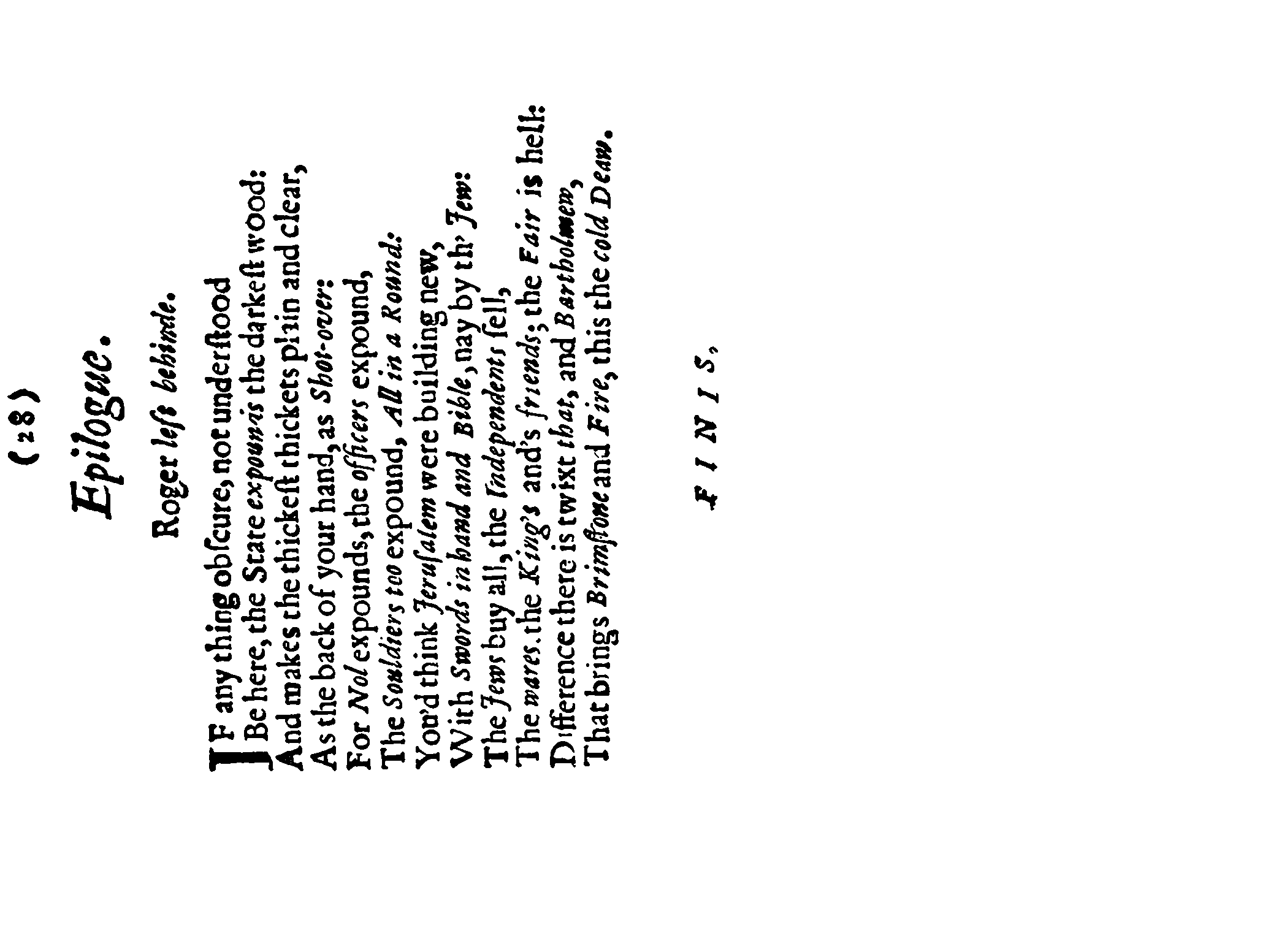} \\ \hline
		\includegraphics[width=.22\textwidth]{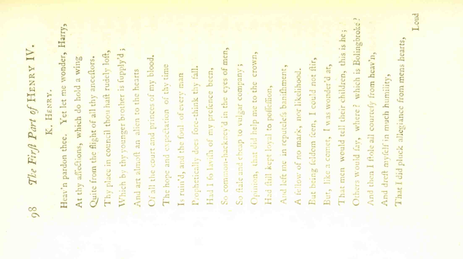} & \includegraphics[width=.22\textwidth]{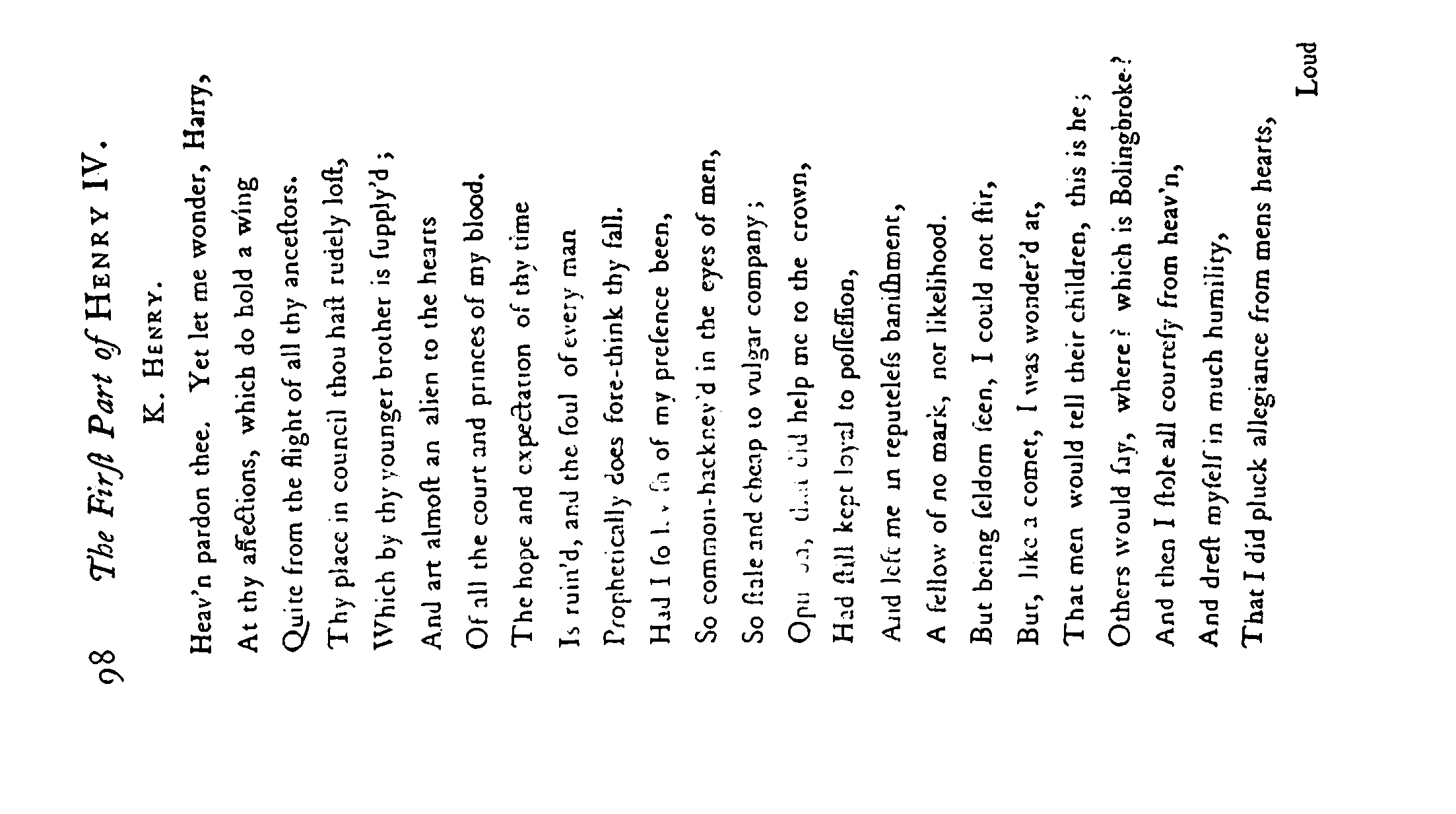} & \includegraphics[width=.22\textwidth]{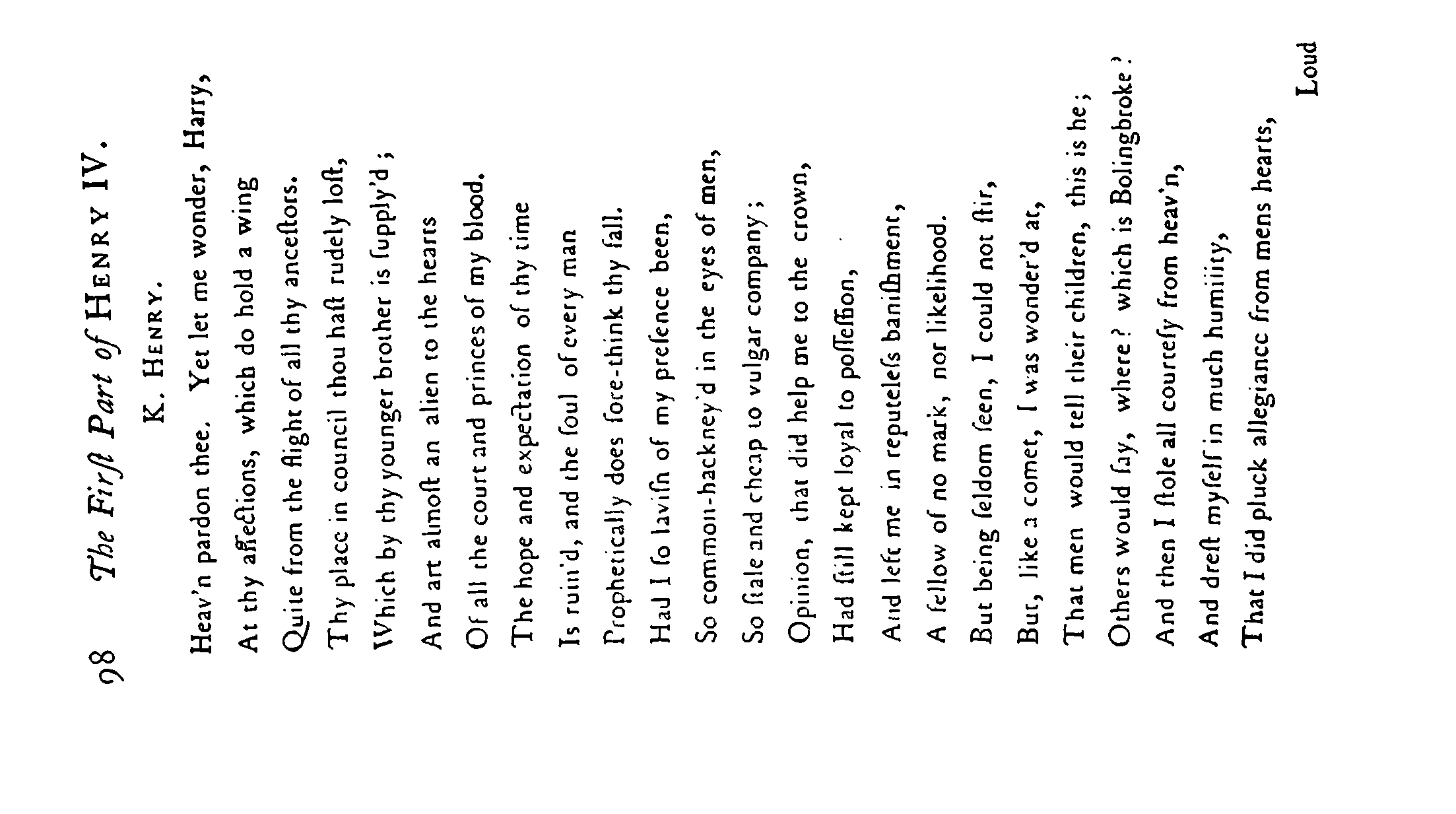} \\ \hline
		\includegraphics[width=.22\textwidth]{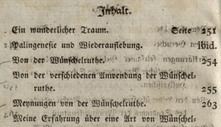} & \includegraphics[width=.22\textwidth]{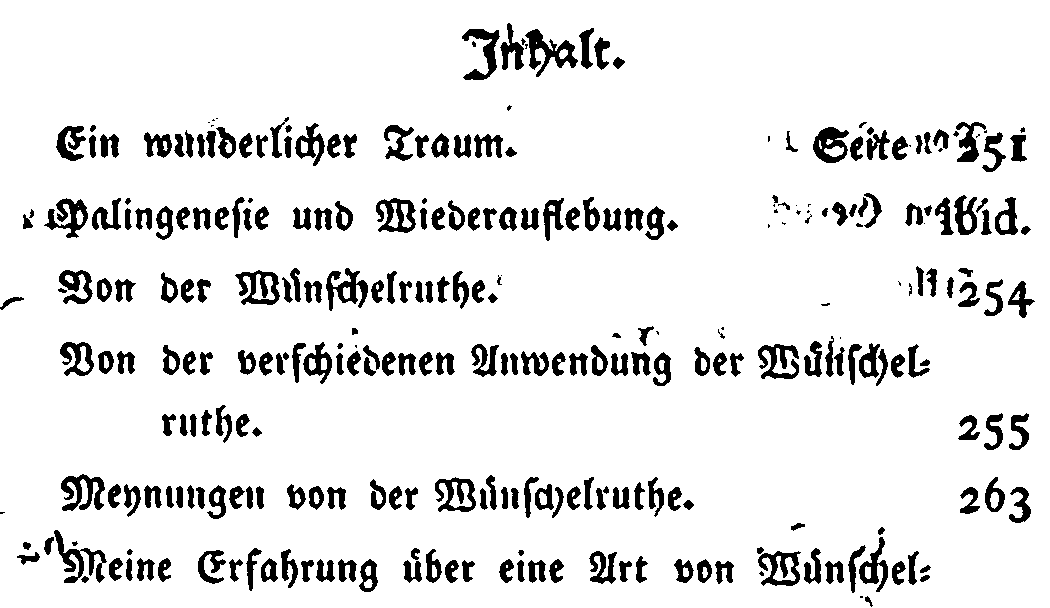} & \includegraphics[width=.22\textwidth]{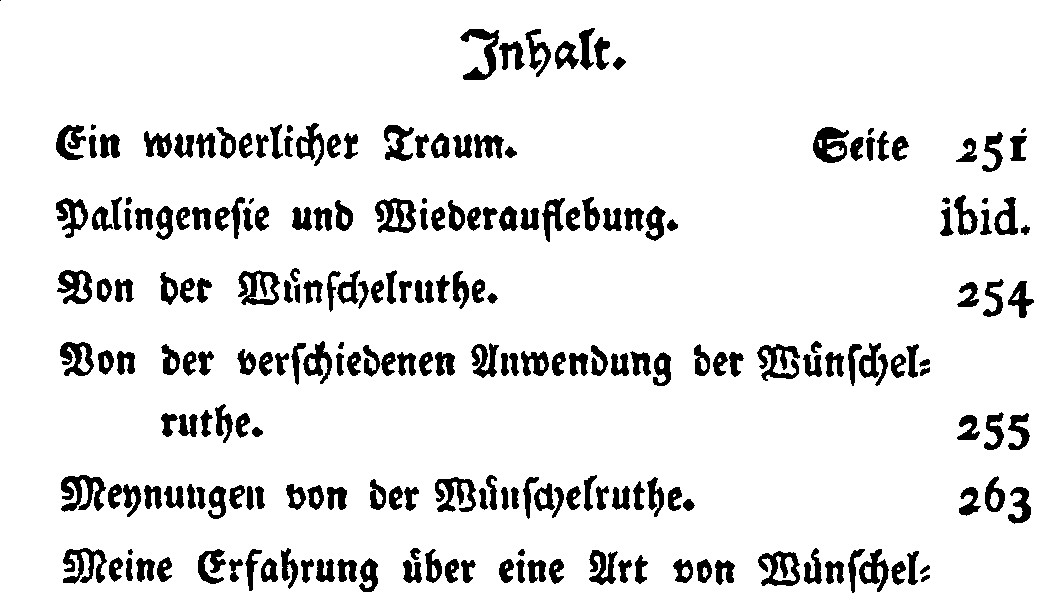} \\ \hline
		\includegraphics[width=.22\textwidth]{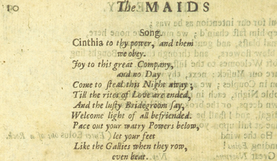} & \includegraphics[width=.22\textwidth]{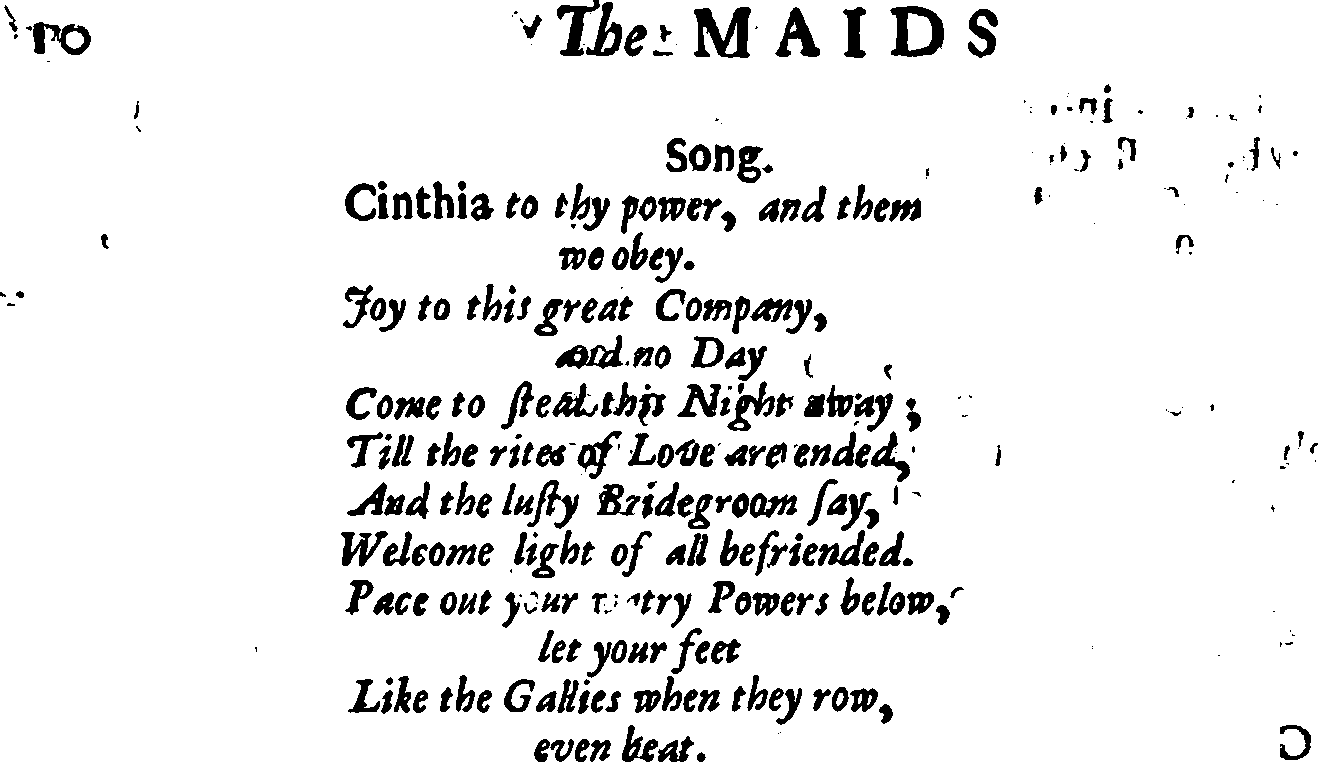} & \includegraphics[width=.22\textwidth]{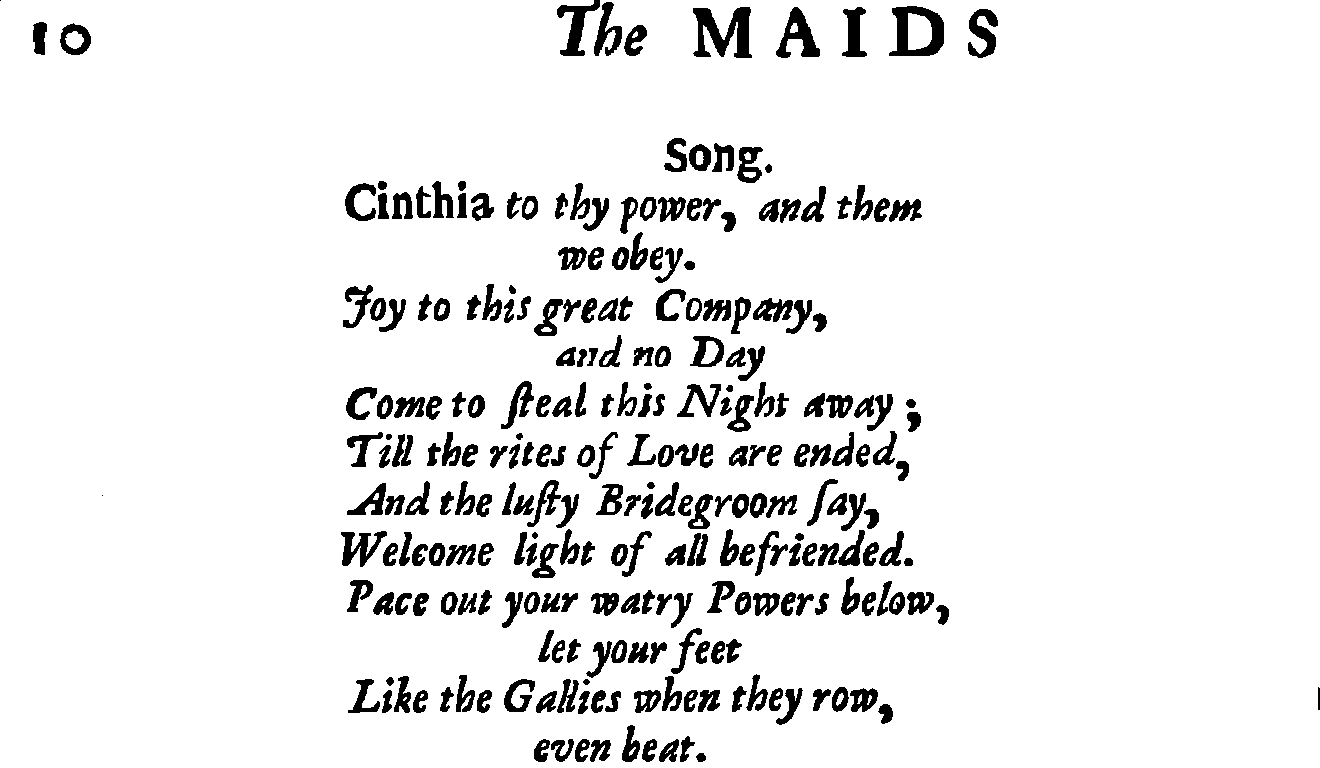} \\ \hline
		\includegraphics[width=.20\textwidth]{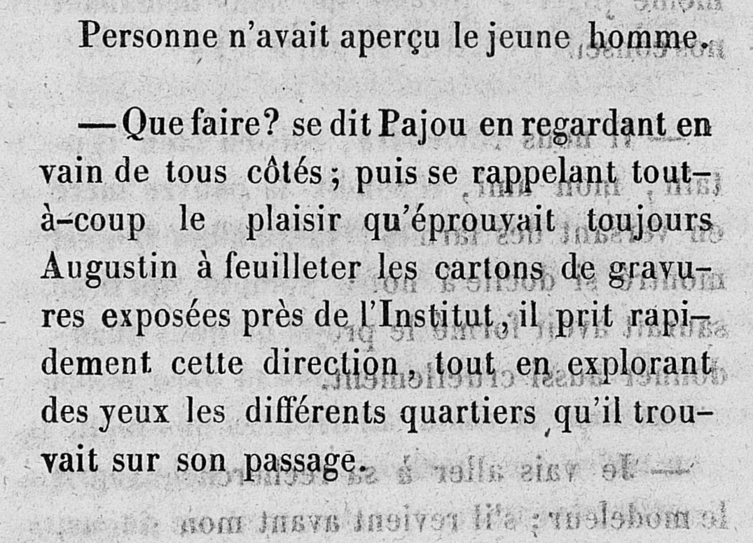} & \includegraphics[width=.20\textwidth]{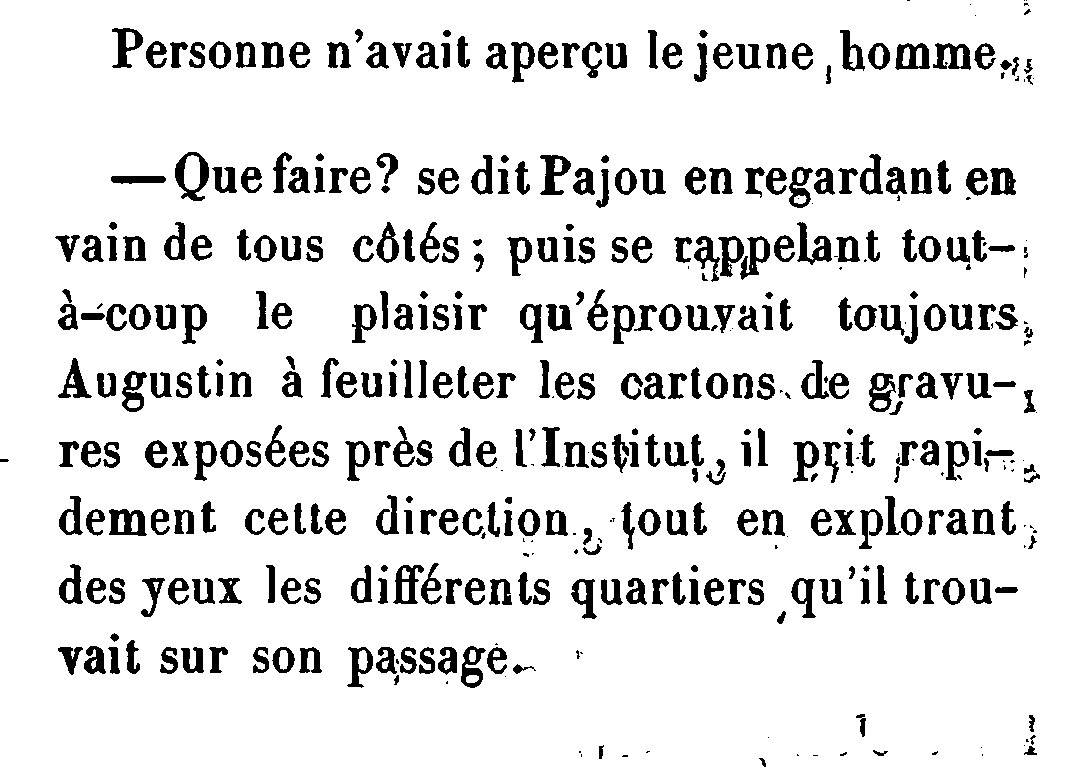} & \includegraphics[width=.20\textwidth]{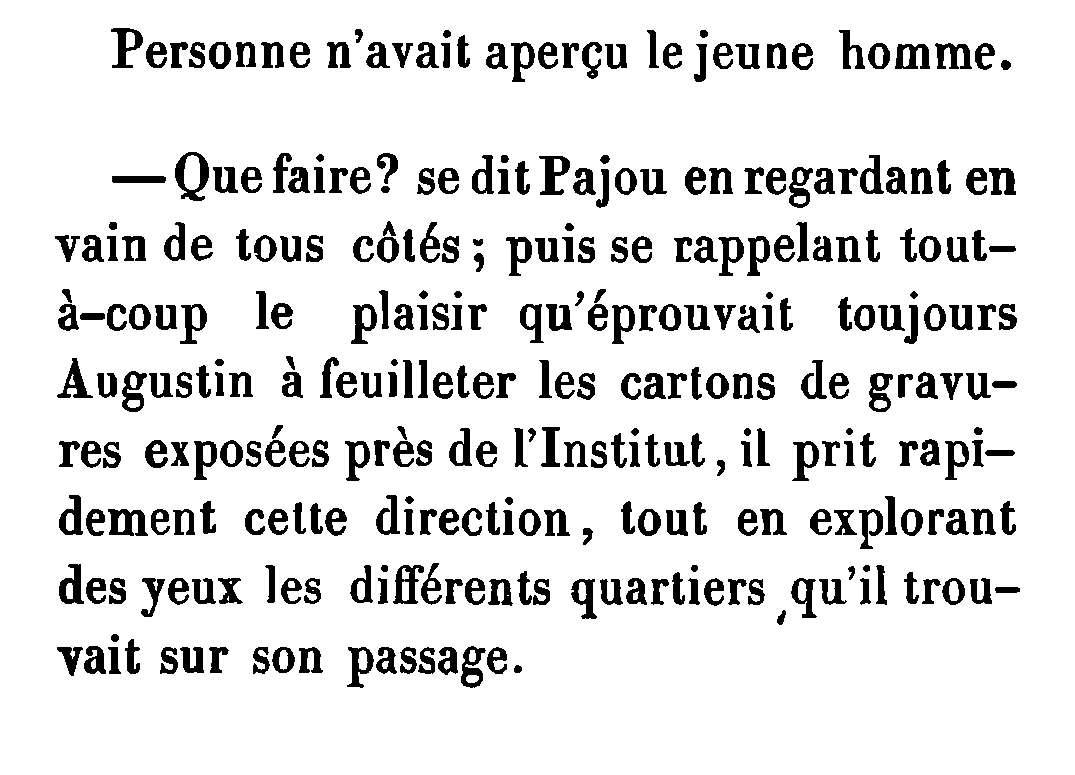} \\ \hline 
		Original image                                                               & BiNet output                                                                   & Ground-truth                                                                \\ \hline
	\end{tabular}
	\caption{Figures of the DIBCO 2017 testing dataset (the machine-printed ones) along with the binarization results from BiNet.}
	\label{appen:dibco2017}
\end{table}

\begin{table}[h!]
	\centering
	\begin{tabular}{|c|c|c|}
		\hline
		\includegraphics[width=.25\textwidth]{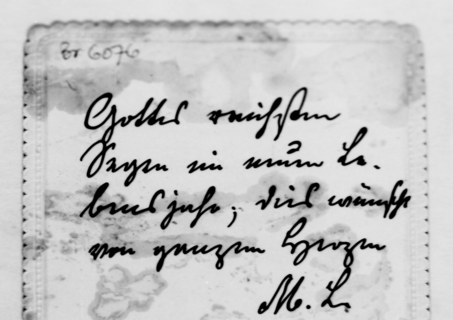} & \includegraphics[width=.25\textwidth]{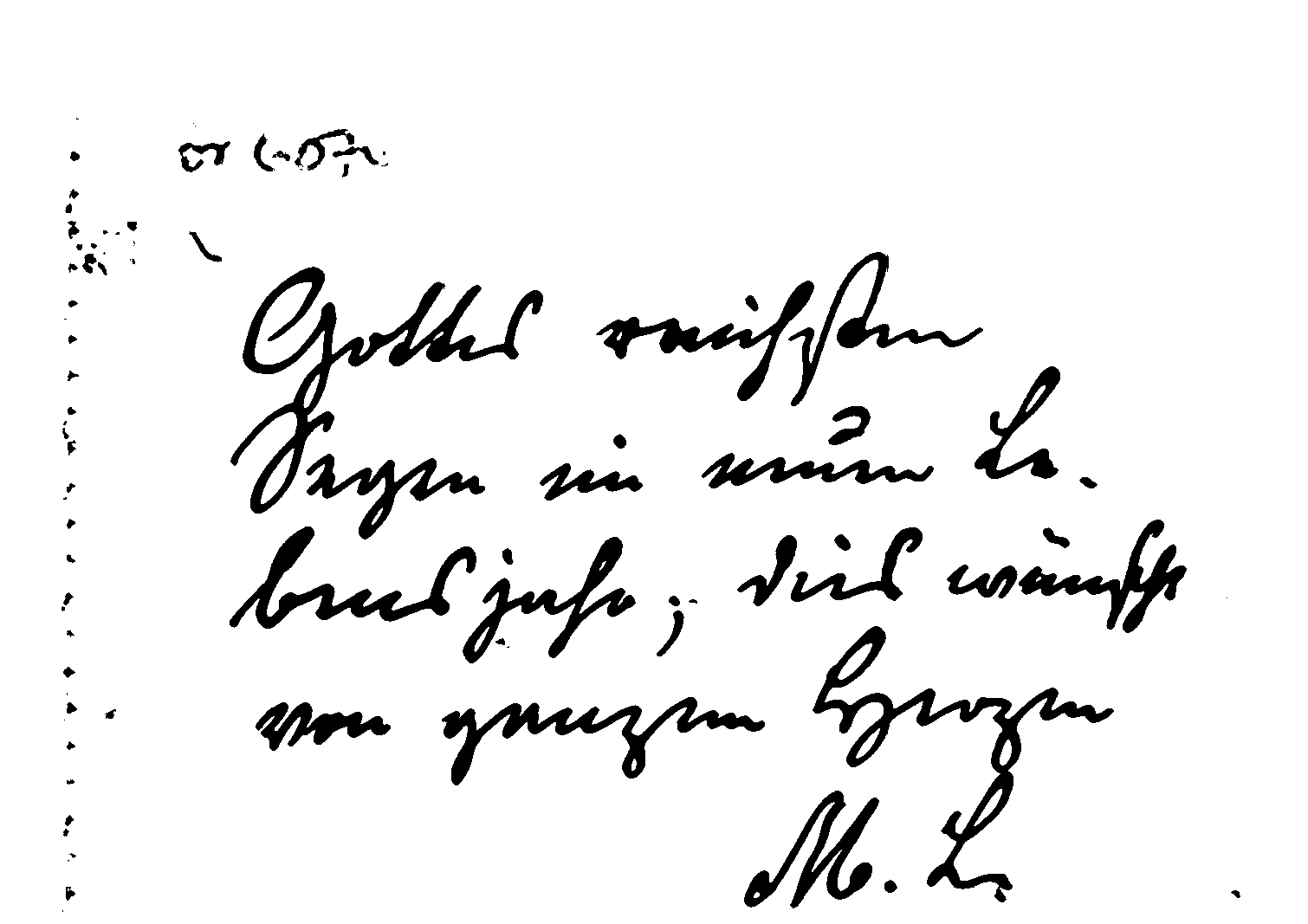} & \includegraphics[width=.25\textwidth]{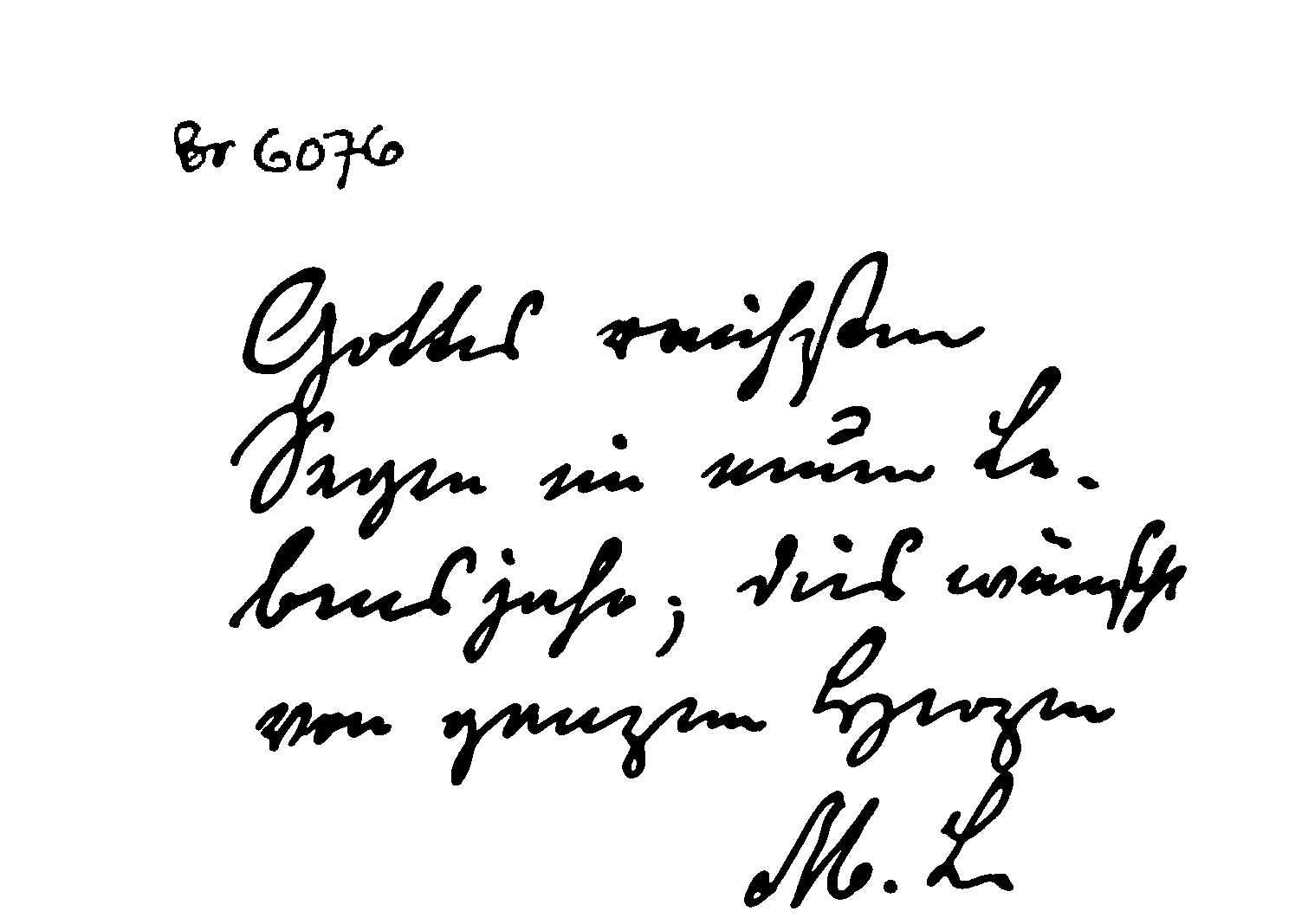} \\ \hline
		\includegraphics[width=.25\textwidth]{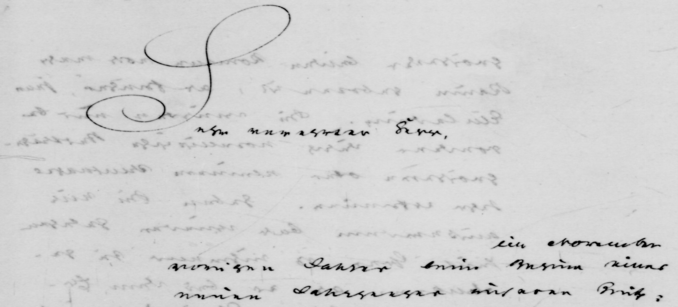} & \includegraphics[width=.25\textwidth]{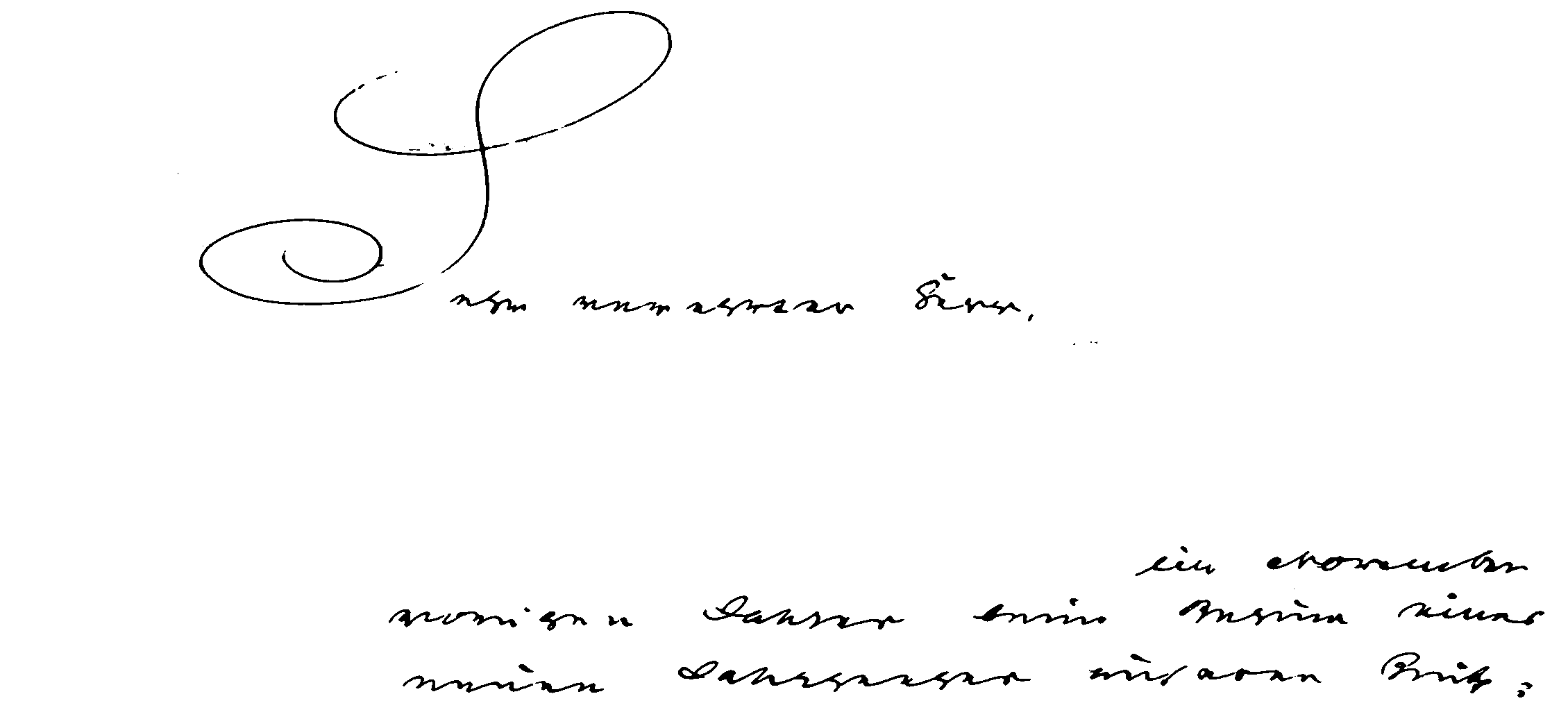} & \includegraphics[width=.25\textwidth]{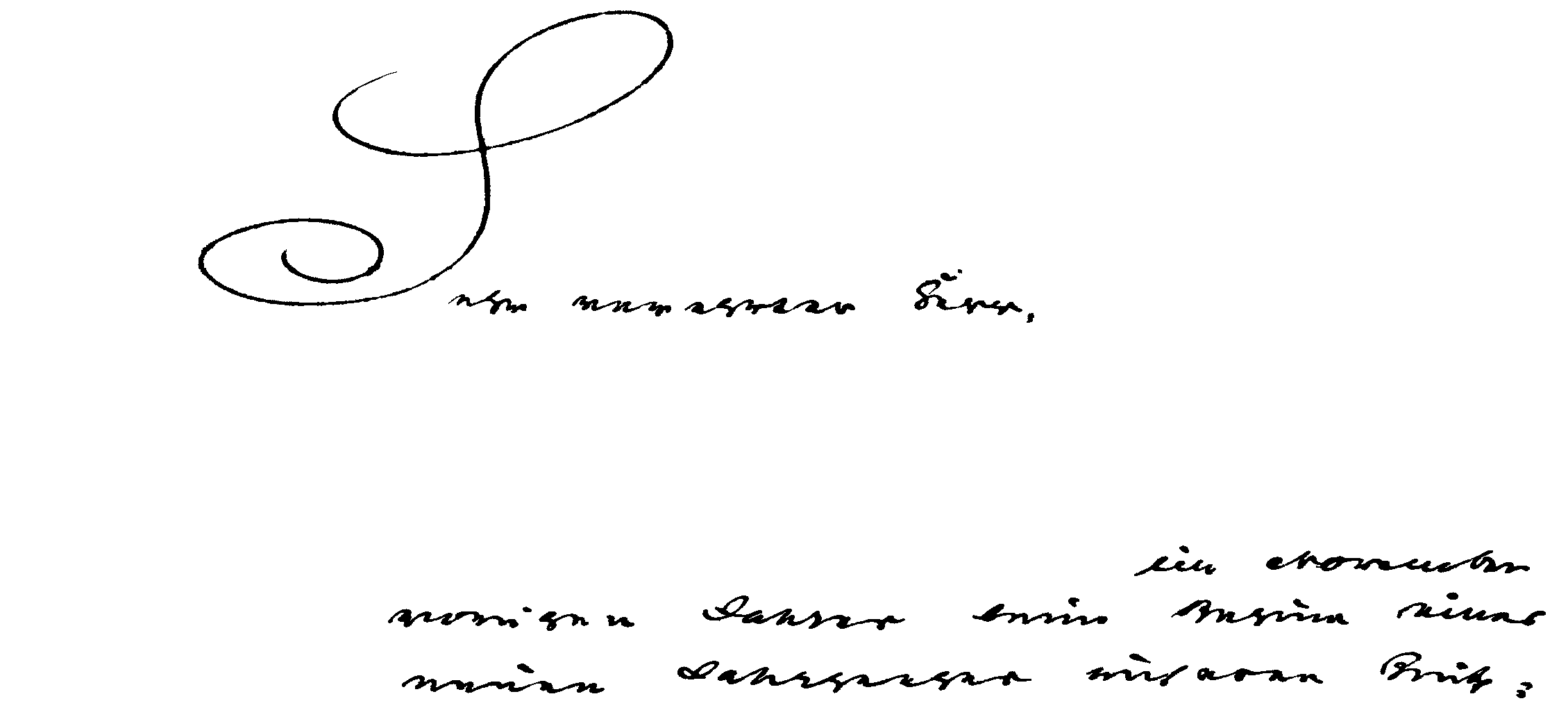} \\ \hline
		\includegraphics[width=.25\textwidth]{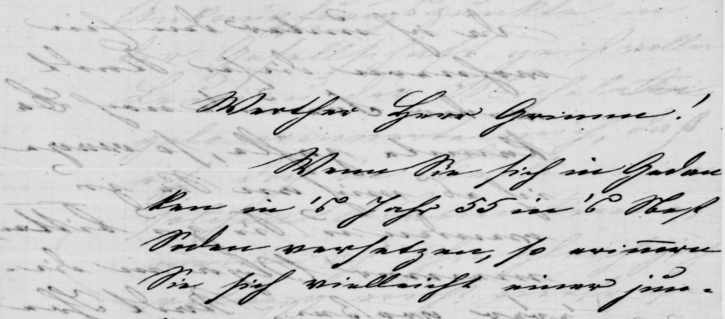} & \includegraphics[width=.25\textwidth]{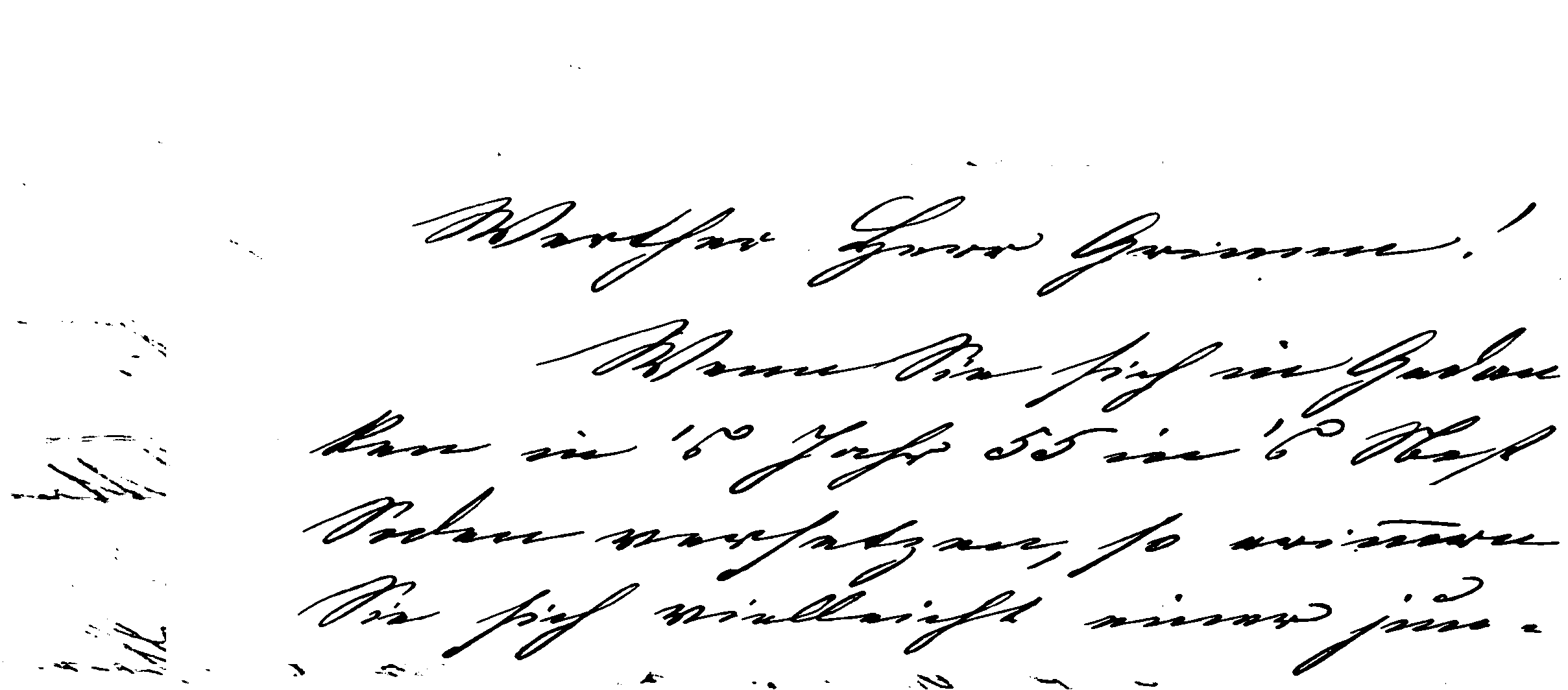} & \includegraphics[width=.25\textwidth]{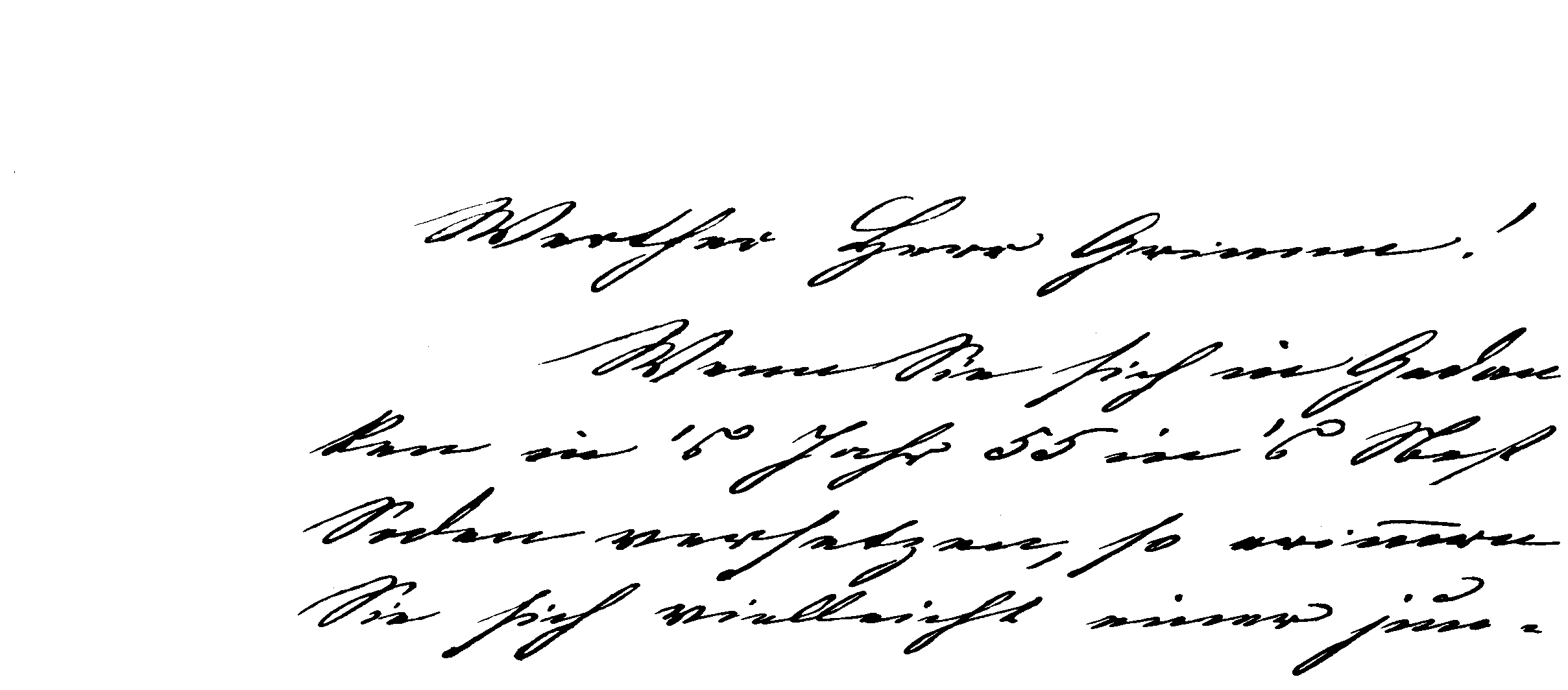} \\ \hline
		\includegraphics[width=.25\textwidth]{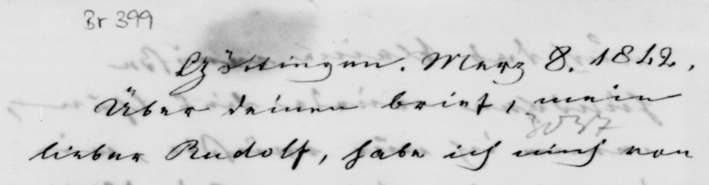} & \includegraphics[width=.25\textwidth]{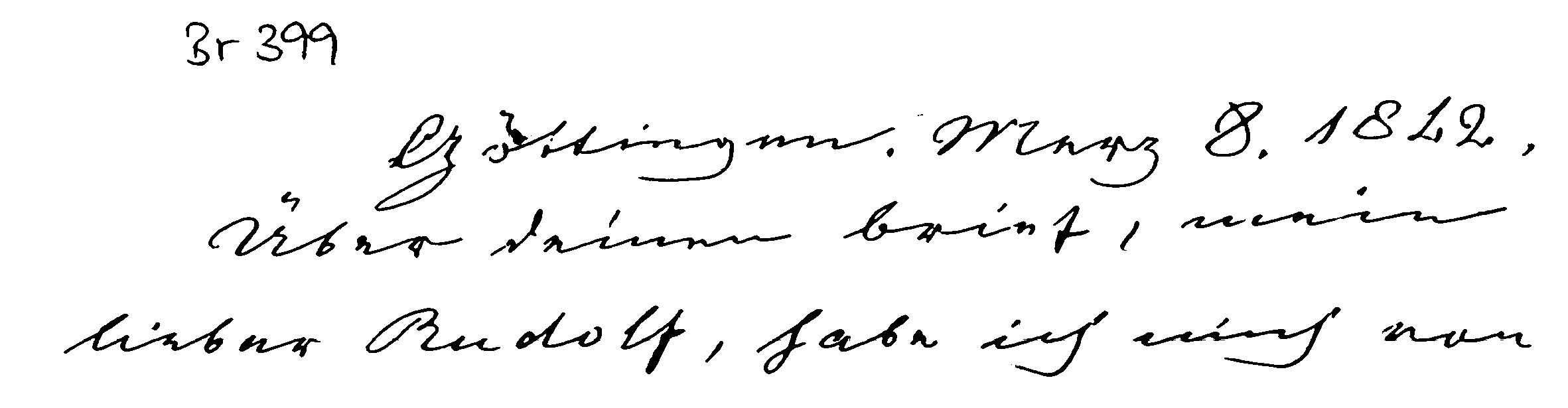} & \includegraphics[width=.25\textwidth]{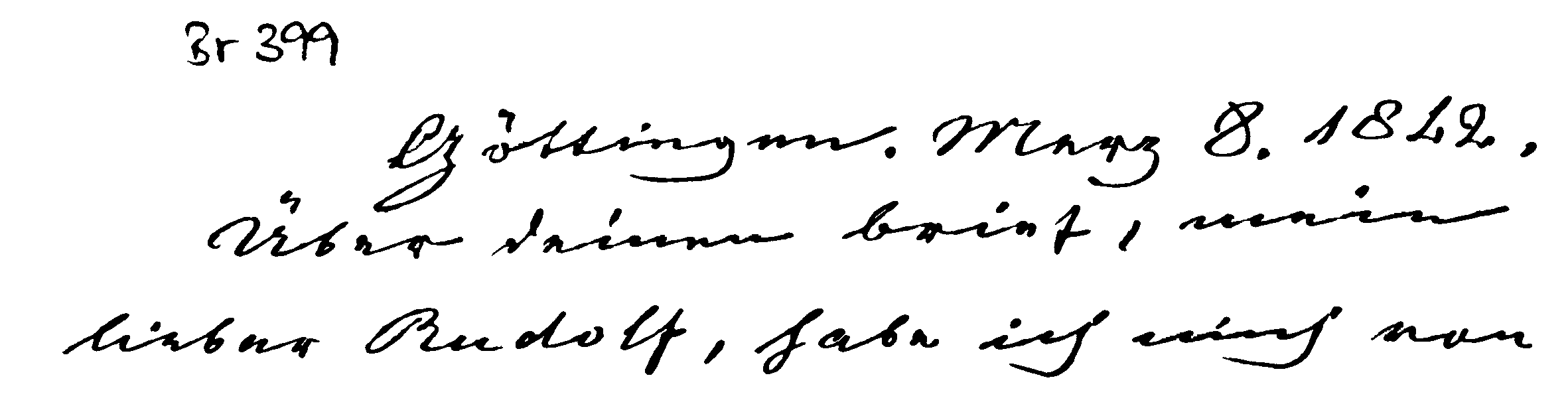} \\ \hline
		\includegraphics[width=.25\textwidth]{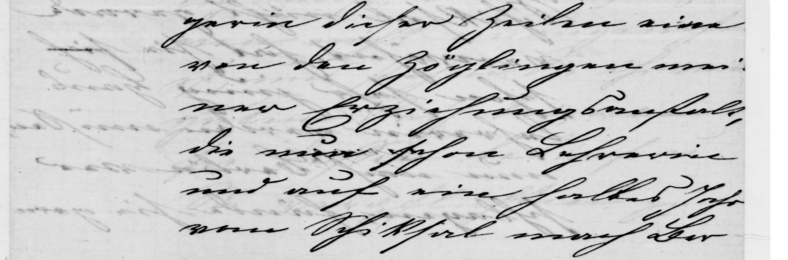} & \includegraphics[width=.25\textwidth]{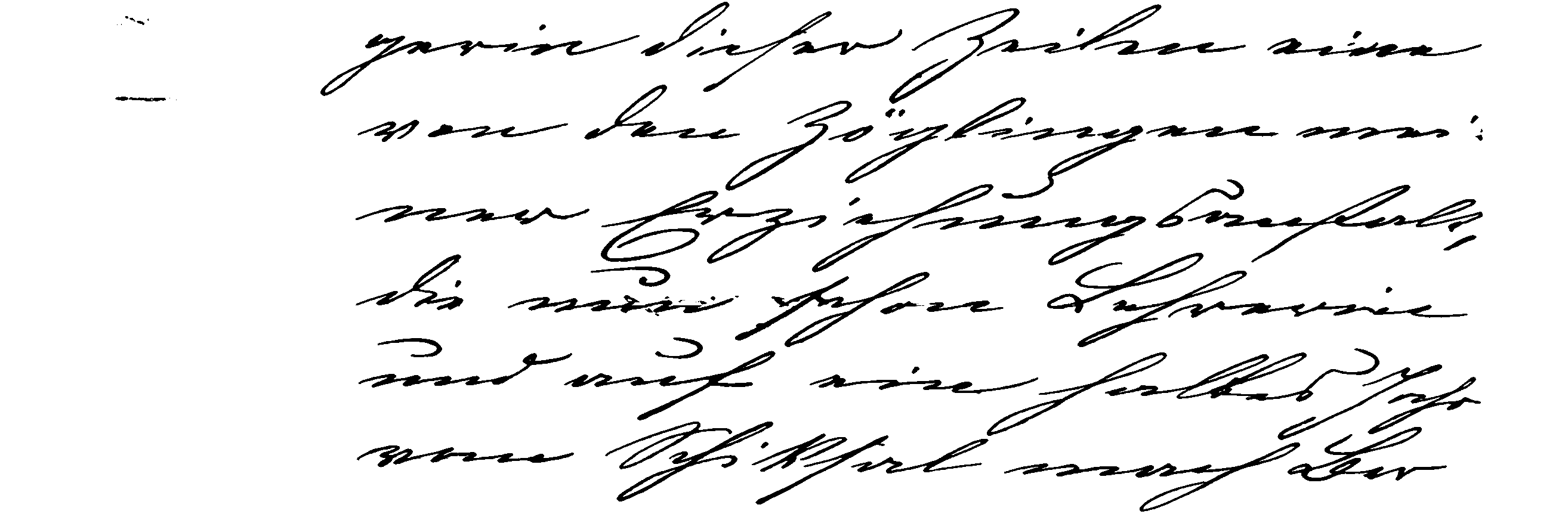} & \includegraphics[width=.25\textwidth]{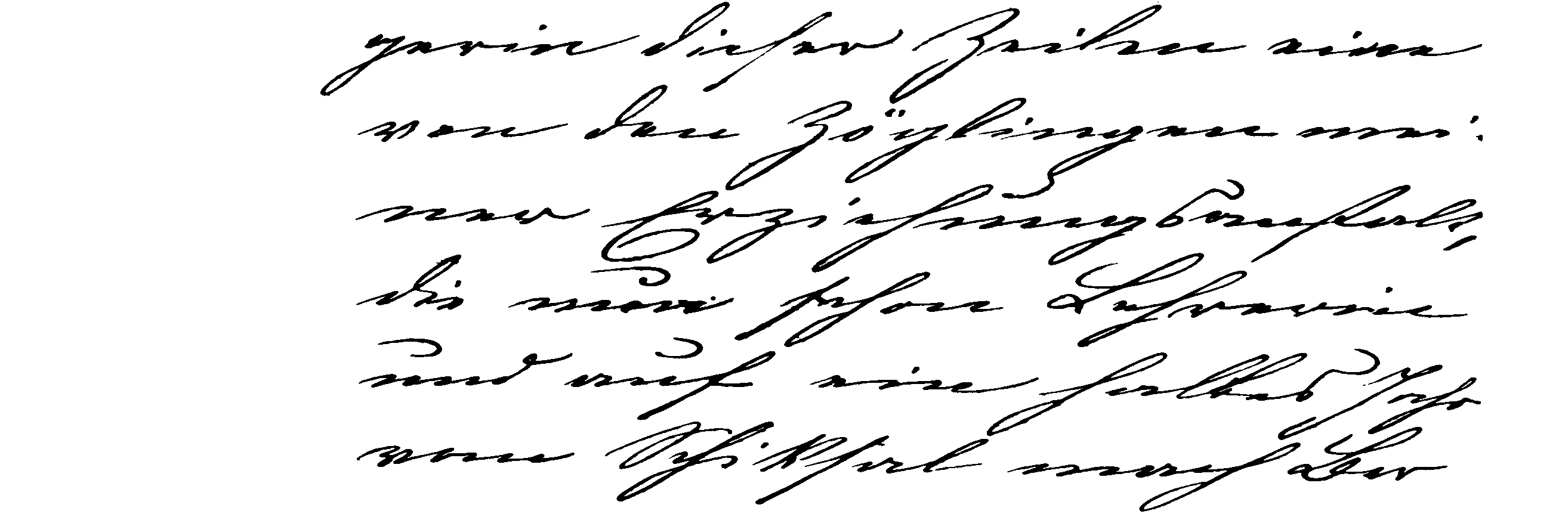} \\ \hline
		\includegraphics[width=.25\textwidth]{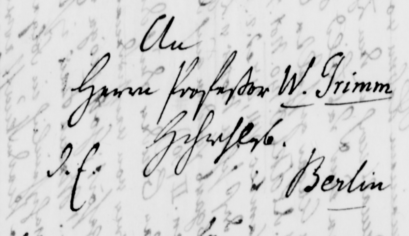} & \includegraphics[width=.25\textwidth]{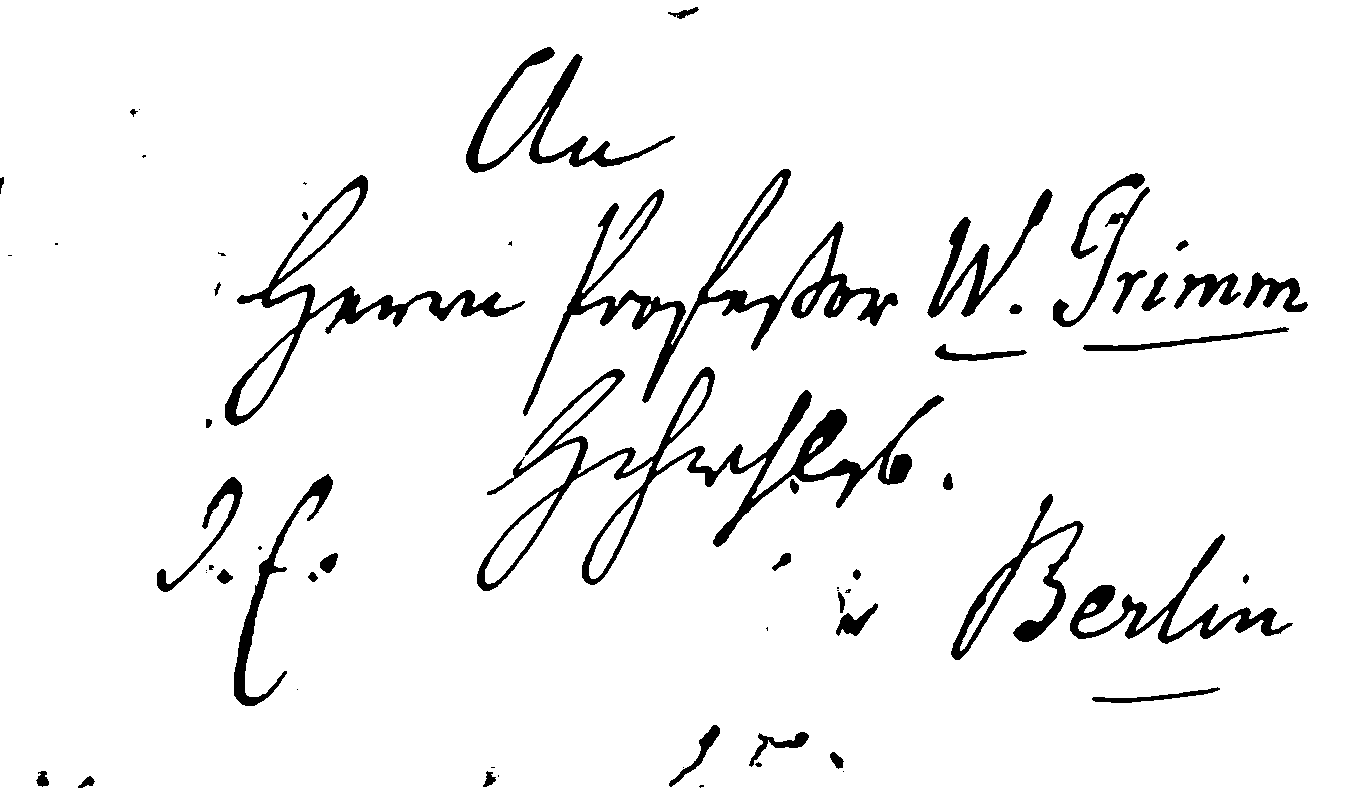} & \includegraphics[width=.25\textwidth]{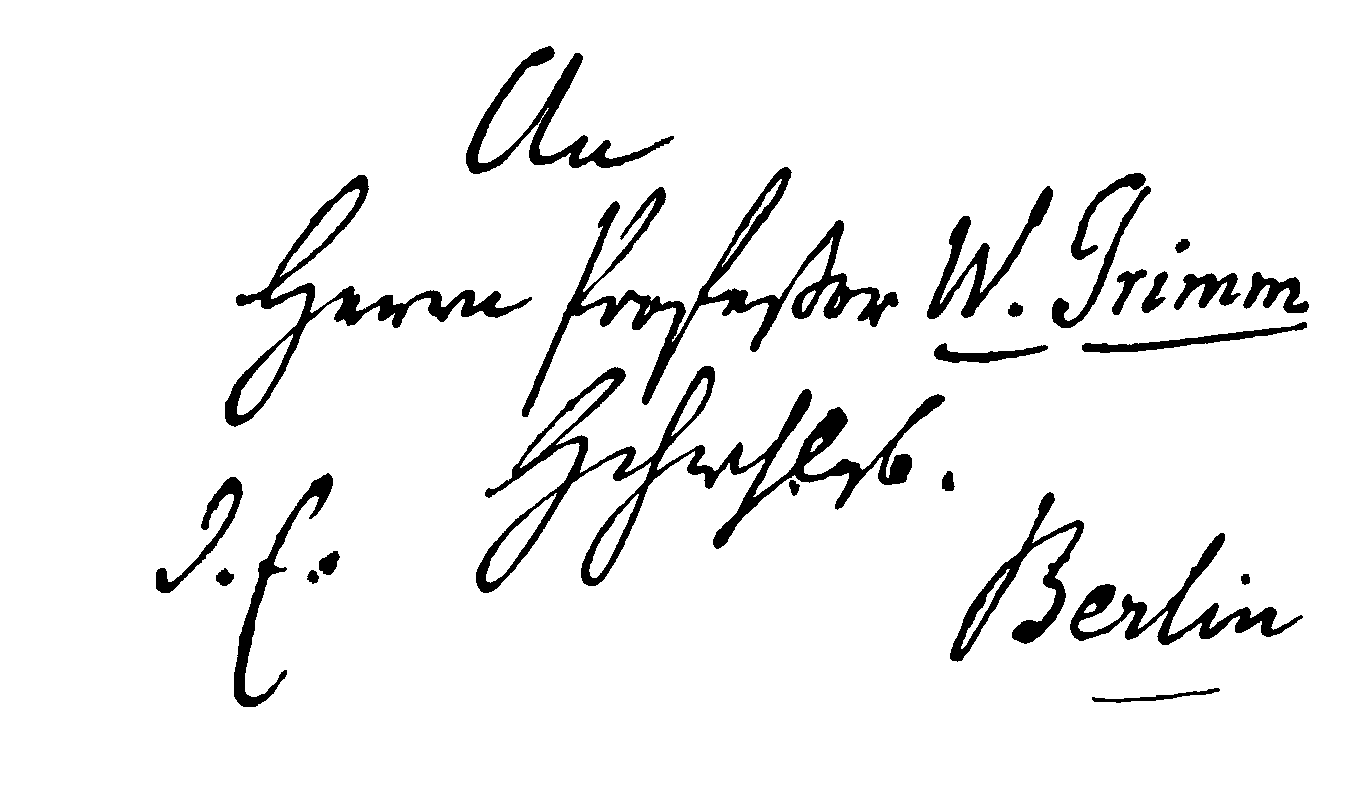} \\ \hline
		\includegraphics[width=.25\textwidth]{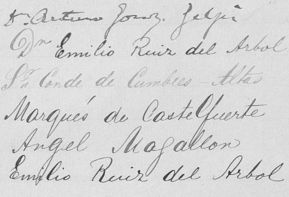} & \includegraphics[width=.25\textwidth]{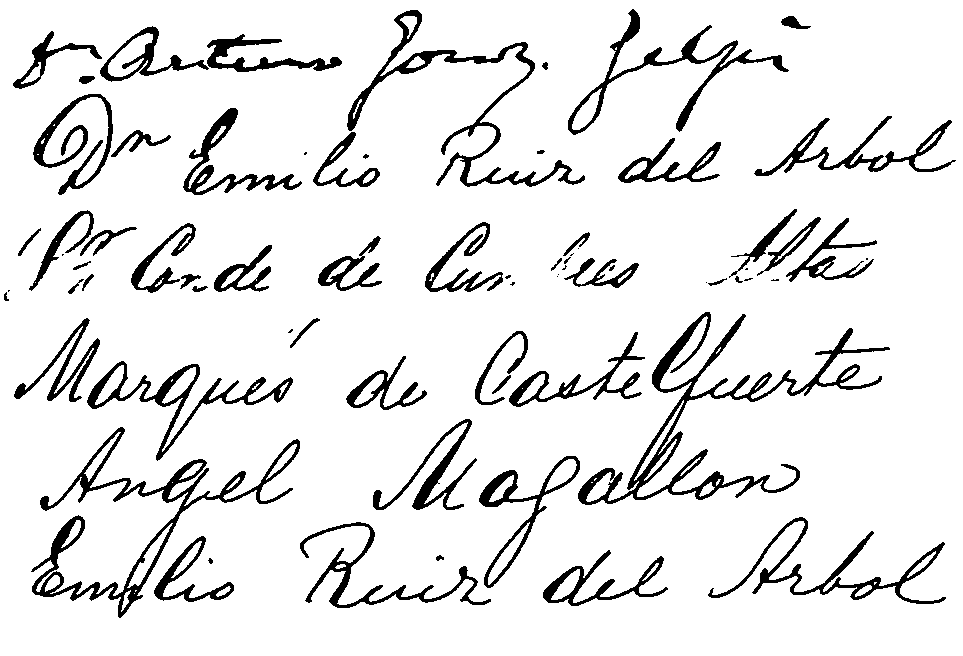} & \includegraphics[width=.25\textwidth]{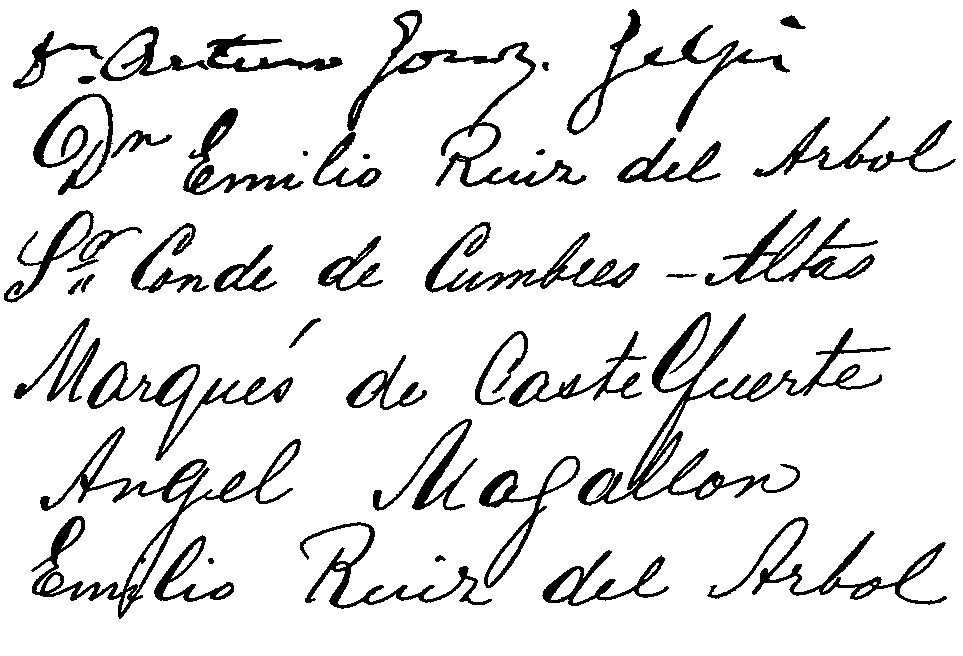} \\ \hline
		\includegraphics[width=.25\textwidth]{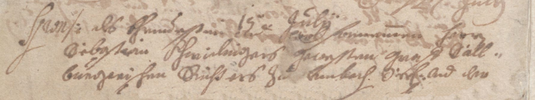} & \includegraphics[width=.25\textwidth]{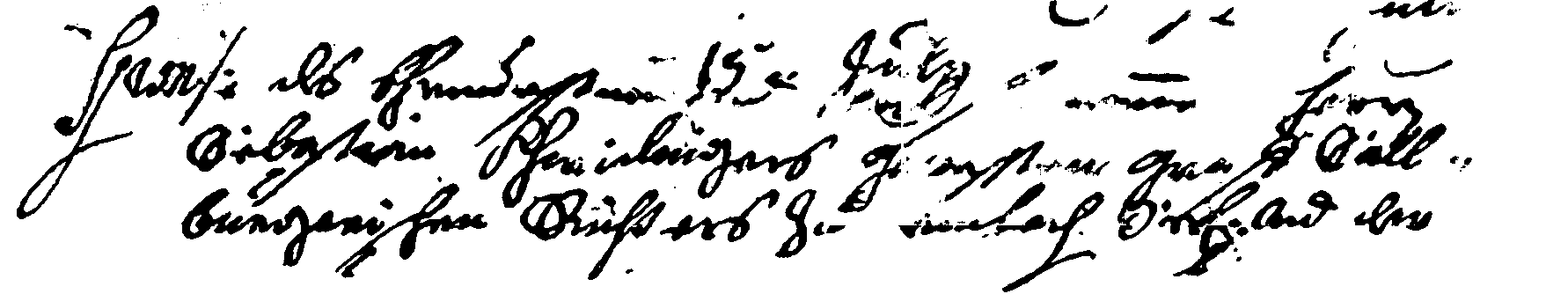} & \includegraphics[width=.25\textwidth]{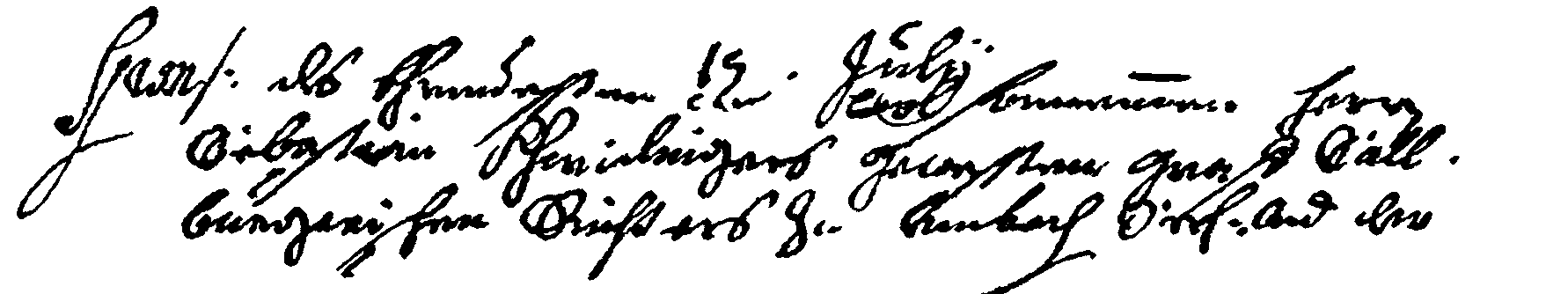} \\ \hline
		\includegraphics[width=.25\textwidth]{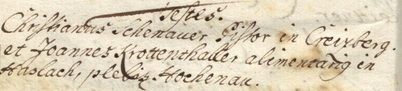} & \includegraphics[width=.25\textwidth]{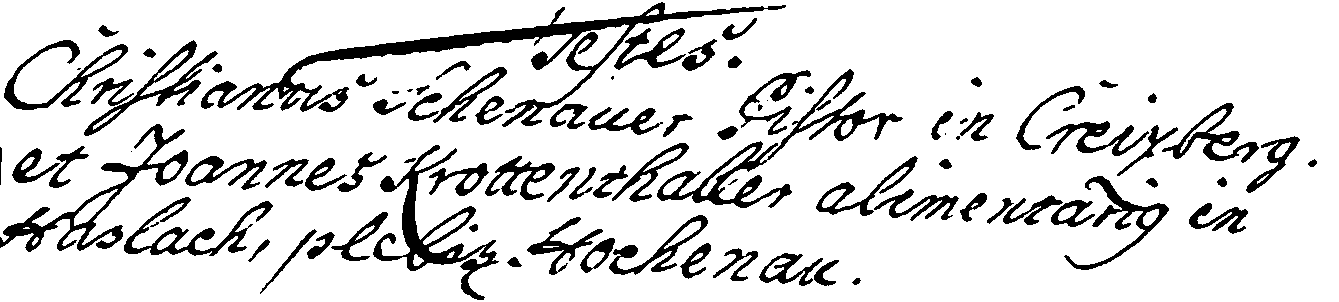} & \includegraphics[width=.25\textwidth]{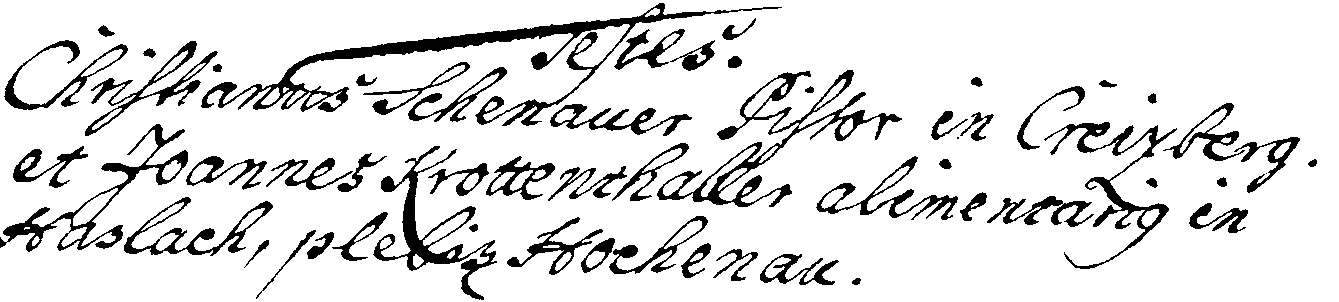} \\ \hline
		\includegraphics[width=.25\textwidth]{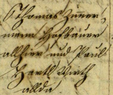} & \includegraphics[width=.25\textwidth]{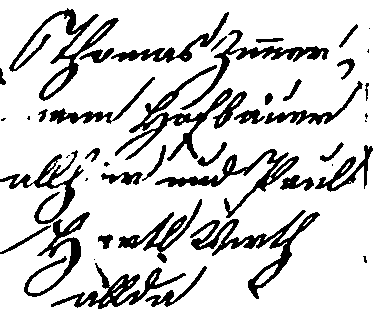} & \includegraphics[width=.25\textwidth]{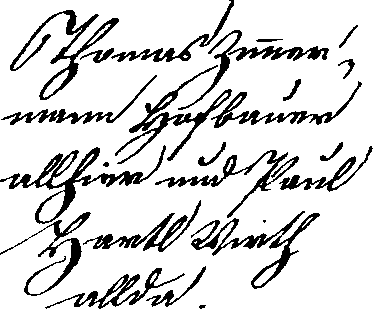} \\ \hline 
		Original image                                                               & BiNet output                                                                   & Ground-truth                                                                \\ \hline
	\end{tabular}
	\caption{Figures of the H-DIBCO 2016 testing dataset along with the binarization results from BiNet.}
	\label{appen:dibco2016}
\end{table}

\end{appendices}